\pdfoutput=1

\documentclass[11pt]{article}

\usepackage[]{ARR2023}

\usepackage{times}
\usepackage{latexsym}

\usepackage[T1]{fontenc}

\usepackage[utf8]{inputenc}

\usepackage{microtype}

\usepackage{inconsolata}

\usepackage{epsfig}
\usepackage{graphicx}
\usepackage{amsmath}
\usepackage{amssymb}
\usepackage{xspace}
\usepackage{amstext}
\usepackage{multirow}
\usepackage{tabularx}   %
\usepackage{booktabs}   
\usepackage{arydshln}
\usepackage{subcaption}
\usepackage{colortbl}
\usepackage{xcolor}
\usepackage{hyperref}
\usepackage{todonotes}
\usepackage{placeins}

\usepackage{microtype}

\newcommand{\stdt}[1] {\textcolor{gray}{\scriptsize{$\pm$#1}}}

\title{One does \textit{not} fit all! \\ On the Complementarity of Vision Encoders for Vision and Language Tasks}

\author{\bf Gregor Geigle\thanks{\,\, Gregor is now affiliated with W\"uNLP \& Computer Vision Lab, CAIDAS, University of W\"urzburg.}, Chen Cecilia Liu,  Jonas Pfeiffer\thanks{\,\, Jonas is now affiliated with Google DeepMind.}, Iryna Gurevych \\
Ubiquitous Knowledge Processing Lab (UKP Lab) \\
Department of Computer Science and Hessian Center for AI (hessian.AI) \\
Technical University of Darmstadt \\
{\url{www.ukp.tu-darmstadt.de}}
}

\begin{document}
\maketitle
\begin{abstract}

Current multimodal models, aimed at solving Vision and Language (V+L) tasks, predominantly repurpose  Vision Encoders (VE) as feature extractors. While many VEs---of different architectures, trained on different data and objectives---are publicly available, they are not designed for the downstream  V+L tasks. Nonetheless, most current work assumes that a \textit{single} pre-trained VE can serve as a general-purpose encoder. In this work, we focus on analysis and aim to understand whether the information stored within different VEs is complementary, i.e. if providing the model with features from multiple VEs can improve the performance on a target task, and how they are combined. We exhaustively experiment with three popular VEs on six downstream V+L tasks and analyze the attention and VE-dropout patterns. Our analyses suggest that diverse VEs complement each other, resulting in improved downstream V+L task performance, where the improvements are not due to simple ensemble effects (i.e. the performance does not always improve when increasing the number of encoders).  We demonstrate that future VEs, which are not \textit{repurposed}, but explicitly \textit{designed} for V+L tasks, have the potential of improving performance on the target V+L tasks.

\end{abstract}

\section{Introduction}

The  dominant strategy for solving Vision+Language (V+L) tasks involves using Transformer models \citep{vaswani_attention_2017} that jointly attend over the representations of the respective modalities~\citep[][\textit{inter alia}]{lu_vilbert_2019,su_vl-bert_2020,li_oscar_2020,chen_uniter_2020,huang_pixel-bert_2020}. While representation-learning of the text modality is comparatively straightforward using token embeddings,\footnote{But still far from solved especially in multilingual settings~\citep{rust_how_2021,clark_canine_2021,xue_byt5_2021}} image representations are more difficult to learn.  Given an image,  a common approach is to use pre-trained Vision Encoders (VE), where the VE's output features are passed as inputs, together with the text embeddings, into a Transformer model. The attention mechanism then learns a cross-modal representation space over the text and image features to solve the target V+L task. 

Consequently, the success of a multimodal model builds heavily on the features extracted from a VE and is thus highly dependent on the VE's architecture, training objectives (e.g. image classification, image encoding, object detection, etc.), and pre-training data. This dependency is further exacerbated for multimodal models that utilize VEs as static feature extractors (i.e. the weights of the VE are frozen), but also for models that are trained end-to-end, as the biases introduced by the architecture, objectives, and data of the VE remain.

Since many computer vision models can be \textit{repurposed} as VEs for V+L tasks, a few prior works have focused on identifying \textit{individual} VEs that perform the best on downstream tasks~\citep{jiang_defense_2020,shen_how_2021,eichenberg_magma_2021,zhang_vinvl_2021}. A common assumption is that a \textit{single} pre-trained VE can perform the best for a target task or even serve as a general-purpose encoder for a wide range of V+L tasks.
However, a natural question arises: to what extent is this assumption 
correct? Given that all VEs differ in architecture, objectives, and pre-training data, we hypothesize that the extracted features of multiple different VEs encode \textit{complementary} information.

In this work, we focus on answering: 1) Do different VEs encode complementary features? 2) \textit{How} are features from different VEs utilized by Transformers? We provide comprehensive analyses for multi-VE models and test whether combining VEs is beneficial over a single-VE setup under the viewpoint of feature complementarity. Similar to prior work that analyzed other components of V+L Transformers~\citep{bugliarello_multimodal_2020,hendricks_decoupling_2021}, we will not focus on improving the performance through ensembling like~\citet{yan_achieving_2021}. Rather, we utilize combinations of VEs as the setting for answering our research questions.  

We cover three popular classes of VEs in our experiments: 1) object detection models providing a feature representation of salient image parts containing objects (\textit{Region})~\citep{anderson_bottom-up_2018}, 2) CNN models computing a feature map of the image for grid features (\textit{Grid}), and 3) Vision Transformers (ViT) \citep{dosovitskiy_image_2020} computing contextualized patch features of the image (\textit{Patch}).
As the downstream domain and task type can be heavily impacted by the different VEs, we probe all combinations of the \textbf{three} VEs on \textbf{six} different V+L tasks, covering retrieval, Q\&A, and reasoning.

To investigate the VE complementarity and feature utilization, we analyze 1) the attention patterns across modalities and VEs, and 2) the dependency of specific VEs when performing VE-dropout during training and inference. While multi-VE seems to perform better than single-VE (which could partially attribute to the increased parameter count), we consistently observe performance gaps between different multi-VE configurations (e.g. a gap as large as 8.9 points for the same task) and no single winning combination for all task types.  Our attention patterns analysis across the different VEs reveals that the distinctive information encoded in the VEs is important for different tasks, and the model composes the representations by enriching a dominant VE with complementary information of the other VEs.  

To sum up, our results and analysis suggest that VEs trained on different objectives, architectures, and data can have a high impact on the model's V+L task performance. We cannot rely on simple ensemble effects to improve performance; selecting and repurposing off-the-shelf VEs is non-trivial, which emphasizes the necessity to design VEs explicitly for V+L tasks in the future.

\section{Related Work}
\label{sec:related}
\noindent\textbf{Multimodal Transformer Architectures.}
\label{sec:related:models}
Multimodal Transformer architectures can be divided into single-stream and dual-stream models \citep{bugliarello_multimodal_2020}.
The single-stream Transformer takes the concatenated visual and text tokens as input and processes them modality-agnostic, i.e. the self-attention jointly attends over the tokens of both modalities.
Dual-stream models use separate Transformers for each modality that are connected through a co-attention mechanism \citep{tan_lxmert_2019,lu_vilbert_2019}, concatenated in a single-stream model on top \citep{singh_flava_2021,kamath_mdetr_2021}, or the image model output is used asymmetrically for cross-attention in the text model \citep{li_align_2021,li_blip_2022}.

The Faster R-CNN \citep{ren_faster_2015} object detector has been the dominant choice for multimodal models as a \textit{Region} VE, where most methods propose to use it as a static feature extractor \citep{tan_lxmert_2019,lu_vilbert_2019,su_vl-bert_2020,chen_uniter_2020,gan_large-scale_2020,li_oscar_2020,zhang_vinvl_2021,cho_unifying_2021}, with the notable exception being \citet{su_vl-bert_2020} who backpropagate through the Faster R-CNN model.
Less popular VEs are 
\textit{Grid} \citep{huang_pixel-bert_2020,kamath_mdetr_2021,yan_grid-vlp_2021,shen_how_2021,eichenberg_magma_2021},
and \textit{Patch} \citep{kim_vilt_2021,wang_simvlm_2021,eichenberg_magma_2021}.
In contrast to \textit{Region} VEs, \textit{Grid} and \textit{Patch} VEs are commonly fine-tuned on the target V+L task, with the notable exception being \citet{yan_grid-vlp_2021}.
Following \citet{bugliarello_multimodal_2020,hendricks_decoupling_2021} we focus on single-stream models as they have been shown to perform on par with dual-stream models
 while being  easier to extend to multi-VE setups. %

\noindent\textbf{Comparing and Combining VEs.}
Recently, several works aim to compare different VEs for V+L tasks.
\citet{jiang_defense_2020} compare
\textit{Region} and \textit{Grid} for visual QA tasks,
showing that training data, objectives and other factors all affect the downstream task performance.
 \citet{shen_how_2021,eichenberg_magma_2021} compare different pre-trained \textit{Grid} and \textit{Patch} VEs building on CLIP \citep{radford_learning_2021}.
\citet{zhang_vinvl_2021} compare   different design choices for  \textit{Region} VEs with \textit{Grid} VEs trained on the same data.
\citet{dai-etal-2023-plausible} compares different VEs in influence object hallucinations in caption generation.
Closest to our work is the work by \citet{yan_achieving_2021}. While they also experiment with combining representations of \textit{Grid}-, \textit{Patch}-, and \textit{Region} VEs, they only focus on the Visual Question Answering \cite[VQA;][]{goyal_making_2017} dataset and only use the combination of all three VEs.
Our work provides a more in-depth evaluation of different multi-VE setups while experimenting with six diverse tasks, and shows that different combinations work best for each task.

\noindent\textbf{Analysis of Multimodal Transformers.}
\label{sec:rw:probing}
Our analysis methods draw inspiration from recent works that probe and analyze pre-trained multimodal Transformers for a better understanding of their different components  \citep{bugliarello_multimodal_2020,cao_behind_2020,li_what_2020,frank_vision-and-language_2021,hendricks_decoupling_2021}.
\citet{cao_behind_2020} propose a range of different probing tasks to understand the inner workings of  multimodal models.
\citet{li_what_2020} analyze how accurate the attention heads of pre-trained models can perform visual grounding.
\citet{frank_vision-and-language_2021} mask parts of the text and image input and measure how the prediction performance changes for the respective other modality to test how symmetric the learned cross-modal connection is.
 \citet{bugliarello_multimodal_2020,hendricks_decoupling_2021} evaluate and disentangle  which components of multimodal pre-training proposed in different works are important for their success. 
While previous work has only focused on models with a \textit{Region} VE, we also experiment with \textit{Grid} and \textit{Patch} VEs.

In summary, our work is the first in-depth study of multimodal Transformers that use multiple VEs.

\section{Multimodal Multi-VE Transformers}
\begin{figure}[]
    \centering
    \includegraphics[width=\columnwidth]{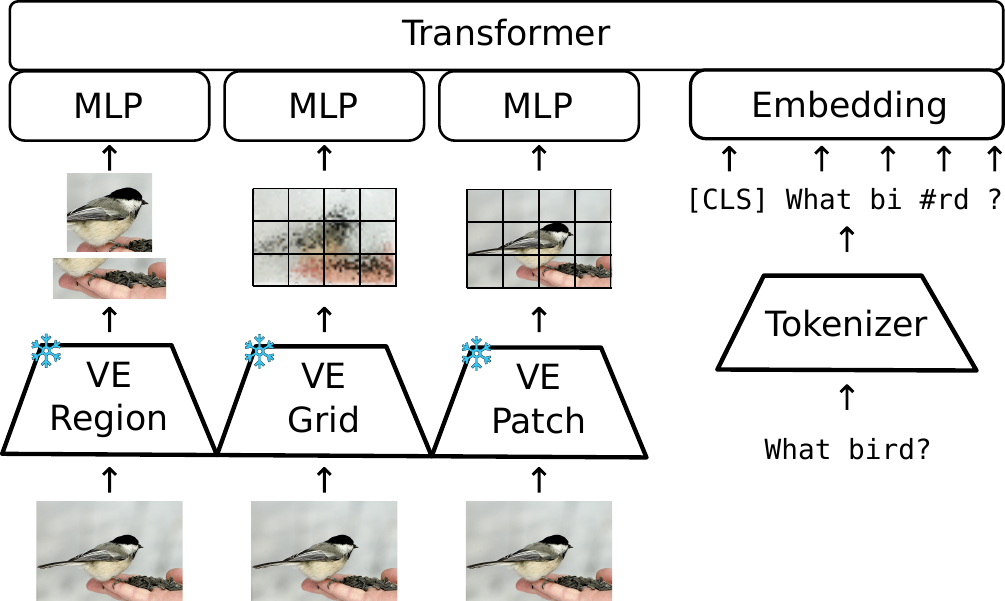}
    \caption{Our Multi-VE Architecture: Each VE produces a list of visual tokens, which are passed through MLPs and concatenated with the text embeddings. The Transformer is modality-agnostic and attends over all tokens. %
    We freeze the VEs during training and only optimize the MLPs, embeddings, and the Transformer.
    }
\label{fig:architecture}
\end{figure}

Recently, cross-modal attention is the dominant strategy to learn multimodal representations with V+L Transformers. In this work, we follow \citet{bugliarello_multimodal_2020} and focus on the single-stream architecture, which shares the attention components across all modalities, i.e. the concatenated visual and text tokens are processed modality-agnostic. This architecture achieves state-of-the-art results \citep{bugliarello_multimodal_2020} while being easily extendable to multiple VEs, by concatenating all vision tokens.\footnote{We concatenate all VE for analysis in \S\ref{sec:ana}. In practice, concatenation increases the sequence length and incurs a high computational cost. More efficient methods like resampling~\citep{alayrac_flamingo_2022} can be explored in future work.} Figure~\ref{fig:architecture} illustrates our architecture. 

\vspace{0.2em}
\noindent\textbf{Multimodal input representations.}
The raw data for a V+L task consists of either discrete tokens/characters (text-modality) or high-resolution pixel values (image-modality). To extract dense representations of the respective modalities we follow the standard pre-processing strategies:  The \textit{text} modality is tokenized using word-piece tokenization \cite{devlin_bert_2019} and mapped to their corresponding dense embedding representations. At the input to the first Transformer layer, positional embeddings are added to the respective token embeddings. For the \textit{vision} modality, pre-trained VEs are utilized which encode the raw pixel values of the respective image into dense high-dimensional feature vectors. These VEs can either encode designated sections (e.g. \textit{Region}), or an entire image (e.g. \textit{Grid} and \textit{Patch}). The extracted feature vectors are then passed through a multi-layer perceptron (MLP),
and subsequently into the Transformer. This procedure can be repeated for any number of VEs of interest. In other words, the image features (from multiple VEs) and the text embeddings are concatenated and jointly passed through a shared Transformer model which learns to attend over the multimodal representations.

\vspace{0.2em}
\noindent\textbf{V+L task training.} 
We place a classification head on the output of the [CLS] token (following \citet{devlin_bert_2019}) and fine-tune the model with cross-entropy loss  on the training data of the target task.

\begin{table*}[ht!]
    \centering
    \footnotesize
    \begin{tabularx}{\linewidth}{l X r X r r}
    \toprule
    \bf VE & \bf Model & \bf \# Train & \bf Tasks & \bf \# V. Tok. & \bf Dim. \\
    \midrule
    Region & VinVL \citep{zhang_vinvl_2021} & 2.5M$^1$ & bounding box prediction, object \& attribute classification  & 36 & 2054\\
    Grid & CLIP RN50x4 \citep{radford_learning_2021} & 400M$^2$ & image-text contrastive loss & 36 & 2560 \\
    Patch & CLIP ViT/B-32 \citep{radford_learning_2021} & 400M$^2$ & image-text contrastive loss & 49 & 768\\
    \bottomrule
    \end{tabularx}
    \caption{The three VE models used in our experiments with the number of pre-training images, training objectives, the number of visual tokens (V.Tok.), and the output feature dimension. \textbf{Train Datasets}: $\mathbf{^1}$: Combination of multiple object detection datasets (see \citep{zhang_vinvl_2021}).
   $\mathbf{^2}$: Web-crawled \& cleaned image-caption pairs (proprietary).
    }
    \label{tab:ve:info}
\end{table*}
\section{Experiments}

We evaluate the impact of three different VEs on six downstream V+L tasks to assess  the complementarity of different image representations. Here, we experiment with all possible combinations of the three VEs (i.e. \textit{single VE}, \textit{2-VE}, and \textit{3-VE} setups).\footnote{While we report the results of a single VE, we do not aim to show that one VE outperforms others, as this would require a more controlled experimental setup, e.g. training dataset and training objectives amongst other factors \citep{jiang_defense_2020,zhang_vinvl_2021}, which is outside the scope of this work.}
To fairly compare the information stored in the respective VEs, we only fine-tune the multimodal models on the target V+L task in order to circumvent potentially beneficial information leaking into the multimodal model from auxiliary tasks. 
We therefore initialize all models with \textit{BERT weights} \citep{devlin_bert_2019} (base-size). 
We note, however,  that gains can be achieved when pre-training the multimodal model on auxiliary data prior to fine-tuning on the target V+L task \cite[][\textit{inter alia}]{tan_lxmert_2019, lu_vilbert_2019,chen_uniter_2020}.

\subsection{Vision Encoders}
We follow the standard approach and repurpose three pre-trained vision models as VEs. In a best-effort attempt for a fair setup, we use the current best publicly available models of similar sizes. Each VE has a designated, randomly initialized 2-layer perceptron (MLP) that maps the representations to the input of the Transformer and is trained on the target V+L task along with the multimodal Transformer weights.
We keep the VE weights \textit{frozen} during training.
For a full summary of the VEs including pre-training data, the number of extracted tokens as well as dimensions, see Table~\ref{tab:ve:info}.

\vspace{0.2em}
\noindent\textbf{\textit{Region} VE.} We utilize Faster R-CNN \citep{ren_faster_2015}, an object detection model that outputs a list of bounding boxes and feature vectors for Regions of Interest---salient parts of the image that likely contain an object. Here we select the pre-trained VinVL object detector  \citep{zhang_vinvl_2021},\footnote{Not to be confused with their Transformer.} which outperforms previous object detectors on V+L tasks.
We follow \citet{li_oscar_2020,zhang_vinvl_2021} and concatenate each extracted feature vector with the corresponding normalized box coordinates and width / height. We  extract the top-36 regions from the VinVL object detector. %

\vspace{0.2em}
\noindent\textbf{\textit{Grid} VE.} \textit{Grid} VEs linearize the grid feature map of a CNN (before final pooling or classification layers) to a list of visual tokens.
Each visual token corresponds to a specific part of the image with image features on different scales (through different pooling operations and convolution sizes throughout the CNN) encoded in it.
We use adaptive max pooling\footnote{\href{https://pytorch.org/docs/1.11/generated/torch.nn.AdaptiveMaxPool2d}{torch.nn.AdaptiveMaxPool2d}} on the feature map to reduce the number of tokens to 36 per image.
We use the CLIP CNN (RN50x4) \citep{radford_learning_2021}  as initialization, given it's recent success on V+L tasks \citep{shen_how_2021,eichenberg_magma_2021,alayrac_flamingo_2022}.

\vspace{0.2em}
 \noindent\textbf{\textit{Patch} VE.} \textit{Patch} VEs use the contextualized output representations of a Vision Transformer (ViT) \citep{dosovitskiy_image_2020} as visual tokens.
The ViT splits an image into uniform patches, which are used as input tokens.
Different from a CNN, the ViT tokens are fixed in size throughout the model but they have a global receptive field through the ViT's attention mechanism.
We exclude the ViT's special classification token from the Transformer input.
We also utilize the CLIP-based ViT models (ViT/B-32)  \citep{radford_learning_2021} for our \textit{Patch}-VE.
We extract all 49 tokens for the CLIP ViT due to their smaller feature dimension size.

\begin{table*}[!ht]
    \centering
    \footnotesize
    \def\arraystretch{0.97}
    \resizebox{0.98\linewidth}{!}{
    \begin{tabular}{l rrrrrr}
    \toprule
    & \multicolumn{2}{c}{\bf Retrieval} & \multicolumn{2}{c}{\bf Question Answering} & \multicolumn{2}{c}{\bf Reasoning}\\
    \cmidrule(lr){2-3} \cmidrule(lr){4-5} \cmidrule(lr){6-7}
    & \bf Flickr30k & \bf MSCOCO & \bf GQA & \bf VQA & \bf SNLI-VE & \bf Hateful M. \\ %
    \bf Vision Encoders & \bf R@1 & \bf R@1 & \bf Acc. & \bf Acc. & \bf Acc. & \bf AUROC  \\ %
    \cmidrule(lr){1-1} \cmidrule(lr){2-3} \cmidrule(lr){4-5} \cmidrule(lr){6-7} %
Region & 57.46 \, \stdt{2.74} & 50.79 \stdt{3.28} & \textbf{55.32} \stdt{0.33} & \textbf{65.73} \stdt{0.54} & 76.57 \stdt{0.10} & 74.83 \stdt{0.73} \\ %
Grid & \textbf{66.93} \, \stdt{3.59} & \textbf{58.30} \stdt{3.56} & 51.51 \stdt{0.17} & 62.99 \stdt{1.25} & \textbf{77.32} \stdt{0.11} & \textbf{79.03} \stdt{0.27} \\ %
Patch & 54.99 \, \stdt{6.00} & 46.30 \stdt{2.07} & 51.56 \stdt{0.44} & 62.96 \stdt{0.71} & 76.32 \stdt{0.09} & 75.78 \stdt{1.57} \\ %
    \cmidrule(lr){1-1} \cmidrule(lr){2-3} \cmidrule(lr){4-5} \cmidrule(lr){6-7} %
Region+Grid & 63.43 \, \stdt{5.85} & 54.87 \stdt{6.88} & 55.08 \stdt{0.44} & 66.30 \stdt{1.52} & 77.66 \stdt{0.11} & 78.68 \stdt{1.82} \\ %
Region+Patch & 58.60 \, \stdt{4.44} & \textbf{\underline{58.73}} \stdt{4.02} & \textbf{\underline{55.58}} \stdt{0.09} & \textbf{\underline{67.05}} \stdt{0.42} & 76.60 \stdt{0.19} & 75.87 \stdt{0.63} \\ %
Grid+Patch & \textbf{\underline{67.53}} \, \stdt{2.07} & 56.44 \stdt{3.80} & 51.55 \stdt{0.16} & 62.64 \stdt{0.30} & 77.39 \stdt{0.35} & \textbf{\underline{79.88}} \stdt{0.95} \\ %
    \cmidrule(lr){1-1} \cmidrule(lr){2-3} \cmidrule(lr){4-5} \cmidrule(lr){6-7} %
Region+Grid+Patch & 62.30  \, \stdt{2.04} & 58.33 \stdt{2.51} & 54.39 \stdt{0.59} & 66.82 \stdt{1.57} & \textbf{\underline{77.87}} \stdt{0.24} & 78.81 \stdt{0.38} \\ %
\midrule
\multicolumn{7}{l}{With VE-Dropout Training \S\ref{sec:ana:dropouttrain}} \\
\midrule
Region+Grid & 55.11 \stdt{13.44} & 54.28 \stdt{4.97} & \textbf{54.91} \stdt{0.26} & 64.72 \stdt{3.93} & 77.07 \stdt{0.21} & 75.75 \stdt{0.93} \\ %
Region+Patch & 51.53 \, \stdt{7.75} & 52.07 \stdt{4.44} & 54.46 \stdt{0.70} & \textbf{65.07} \stdt{0.63} & 76.42 \stdt{0.15} & 73.57 \stdt{0.69} \\ %
Grid+Patch & \textbf{63.13} \, \stdt{4.16} & \textbf{55.56} \stdt{3.92} & 51.80 \stdt{0.25} & 60.62 \stdt{1.38} & \textbf{77.41} \stdt{0.16} & \textbf{77.30} \stdt{0.43} \\ %
    \bottomrule
    \end{tabular}
    }
    \caption{Mean and standard deviation over three seeds. Metrics: for retrieval the average recall at 1 between image-text and text-image retrieval, for Hateful Memes AUROC, and accuracy otherwise.
    Best single- and multi-VE setup is \textbf{bolded} and overall best score is \underline{underlined}.
    We also report the results for VE-Dropout Training (see \S\ref{sec:ana:dropouttrain}).}
    \label{tbl:res:scores}
\end{table*}

\subsection{Tasks}
We experiment with a  set of six  V+L tasks:   \textit{Image-text retrieval} (\textbf{Flickr30k} \citep{young_image_2014} and \textbf{MSCOCO} \citep{lin_microsoft_2014}),  \textit{visual question answering} (\textbf{GQA} \citep{hudson_gqa_2019} and \textbf{VQA2.0} \citep{goyal_making_2017}), 
    \textit{visual entailment} (\textbf{SNLI-VE} \citep{xie_visual_2019}) and \textit{memes classification} (\textbf{Hateful Memes} \citep{kiela_hateful_2020}).
For all experiments we report the mean and standard deviations over three random seeds and present training details and hyperparameters in  Appendix \ref{sec:appendix:hyper}.
We train all models with a single Nvidia V100 GPU, training a single model (all tasks, three seeds) takes approximately 10 GPU days.

\subsection{Results \& Discussion}
\label{sec:results_discussion}
We report the results on the six tasks with all possible combinations of VEs in Table~\ref{tbl:res:scores}.
 
\vspace{0.2em}
\noindent\textbf{No ``one encoder to rule them all".} 
When comparing the results of the single-VE models, there is no clear single winning VE that outperforms all other VEs across all tasks. While for QA tasks \textit{Region} VE models perform best, for the other tasks  \textit{Grid} VE outperforms the others.
We hypothesize that the object-centric regions are useful for QA tasks, which focus on specific elements, while the uniform grid encoding might be useful for retrieval and other tasks that look at the entire image. 
Isolating \textit{why} certain VEs are useful for specific tasks, and the role of training objectives, data, and architecture, requires a controlled setup of training VEs from scratch \citep{jiang_defense_2020}, which we leave to future work.
Interestingly, the \textit{Patch} VE never achieves the best performance, which aligns with  previous findings by \citet{shen_how_2021,eichenberg_magma_2021}.
These single-VE results demonstrate that each VE encodes different types of information that impacts the downstream performance.

\vspace{0.2em}
\noindent\textbf{VEs can complement each other.}
When combining the representations from different VEs, we witness improvements across all V+L tasks.
Interestingly, MSCOCO  benefits greatly from combining the two weakest VEs (i.e. \textit{Region} and \textit{Patch}), surpassing their corresponding single VE results by 7.94 and 12.43 points respectively, achieving the best performance on this task. Although the \textit{Patch} VE never achieves the best performance in single-VE setups, it provides complementary information in combination with the best performing VE, achieving the best overall performance for many tasks.\footnote{Grid+Patch performs better than Grid for Flickr30k, Region+Patch performs better than Region for GQA, etc.} 
However, we see that simply using more encoders does \textit{not} guarantee improvements as is evident by the 3-encoder model.
While the 3-encoder model consistently achieves near-best results, it is rarely the best model (only on 1 out of 6 tasks). This result shows that simply using more encoders does \textit{\textbf{not}} guarantee improvement (i.e. model performance is not monotonically improving with more VEs.). Hence, it is unlikely that the improvements are due to an ensemble effect. 

\vspace{0.2em}
\noindent\textbf{One does not fit all.}
In summary, we see that neither one VE alone nor a fixed combination of VEs gives the best results for the entire breadth of V+L tasks.
This shows the current limitations of repurposed vision encoders and highlights the need for encoders designed specifically for V+L tasks.

\section{Analysis}
\label{sec:ana}
To better understand how the representations are combined in different multi-VE setups, we analyze the \textit{flow of attention}, \textit{phrase-to-image grounding}, and the robustness to \textit{dropping VEs} at test time.
We overload `cross-modality' to include both VE-text but also VE-VE interactions for simplicity.
We present the analysis for the best performing model combinations in what follows but provide a full list of results in 
Appendix~\ref{sec:appendix:plot}.
\subsection{CLS Attention Flow}
\label{sec:CLS_Attetion}
\begin{figure*}[!ht]
\centering
    \begin{subfigure}{.3\linewidth}
    \centering
        \includegraphics[width=.6\linewidth]{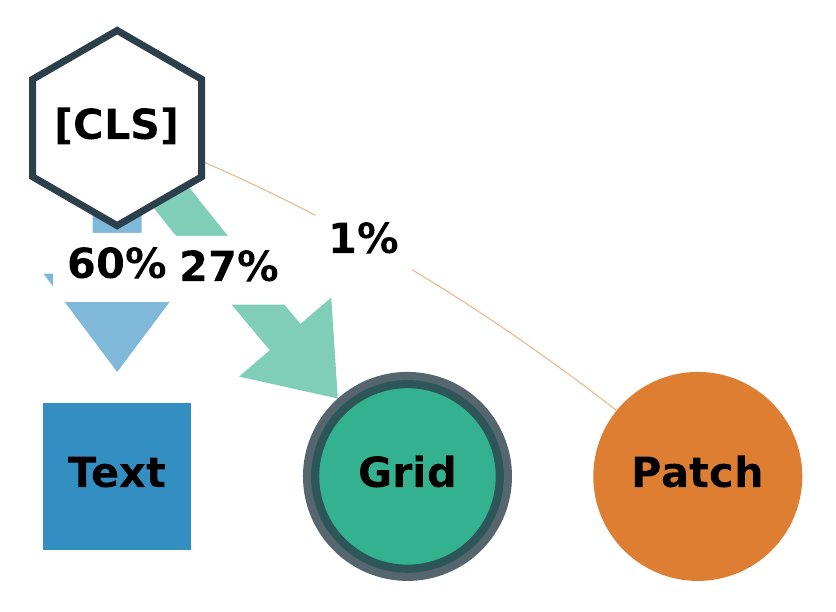}
        \caption{CLS Attention for Flickr30k}
        \label{fig:ana:clsattention:flickr}
    \end{subfigure}
    \begin{subfigure}{.3\linewidth}
    \centering
        \includegraphics[width=.6\linewidth]{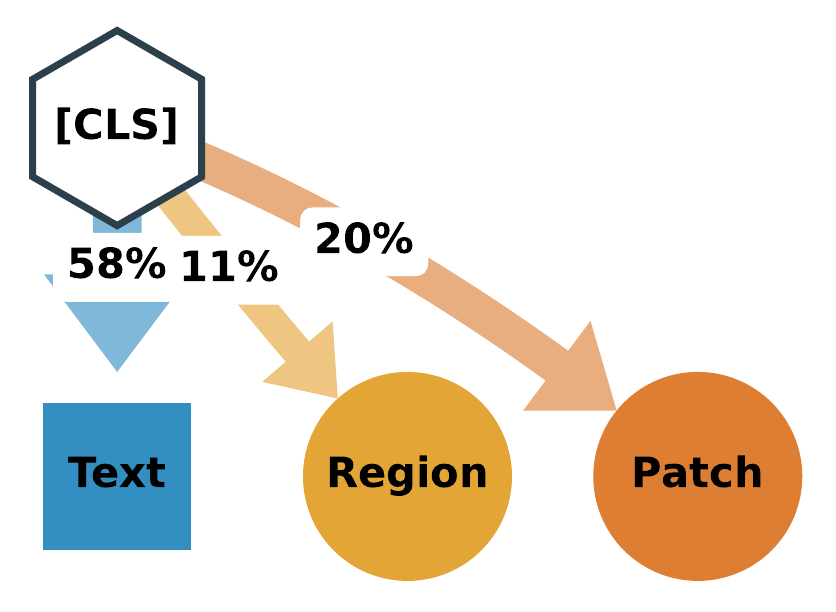}
        \caption{CLS Attention for MSCOCO}
        \label{fig:ana:clsattention:mscoco}
    \end{subfigure}
    \begin{subfigure}{.3\linewidth}
    \centering
        \includegraphics[width=.6\linewidth]{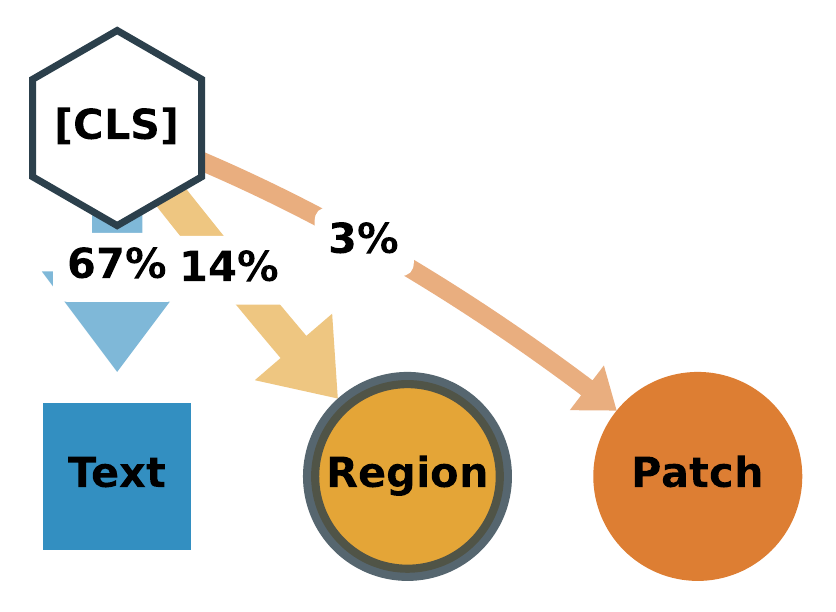}
        \caption{CLS Attention for GQA}
        \label{fig:ana:clsattention:gqa}
    \end{subfigure}
    \\
    
    \begin{subfigure}{.3\linewidth}
    \centering
        \includegraphics[width=.6\linewidth]{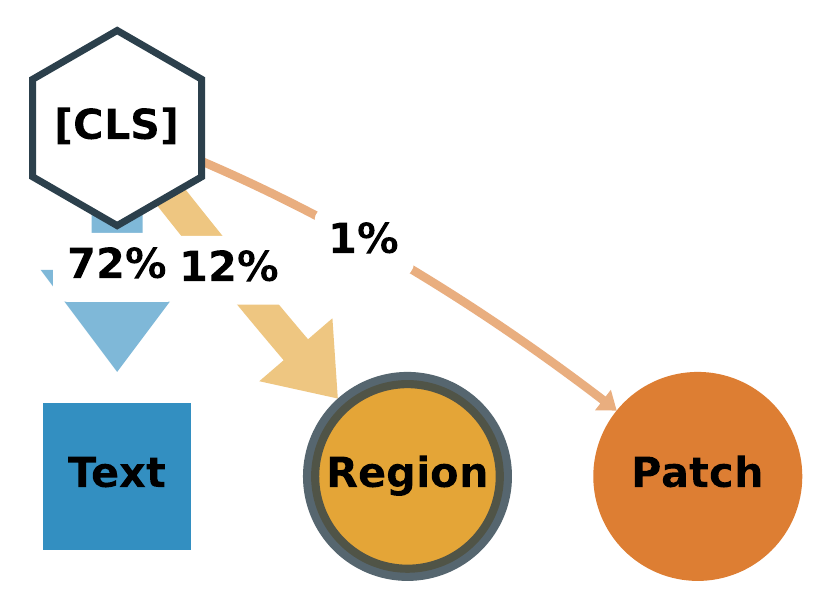}
        \caption{CLS Attention for VQA}
        \label{fig:ana:clsattention:vqa}
    \end{subfigure}
    \begin{subfigure}{.3\linewidth}
    \centering
        \includegraphics[width=.7\linewidth]{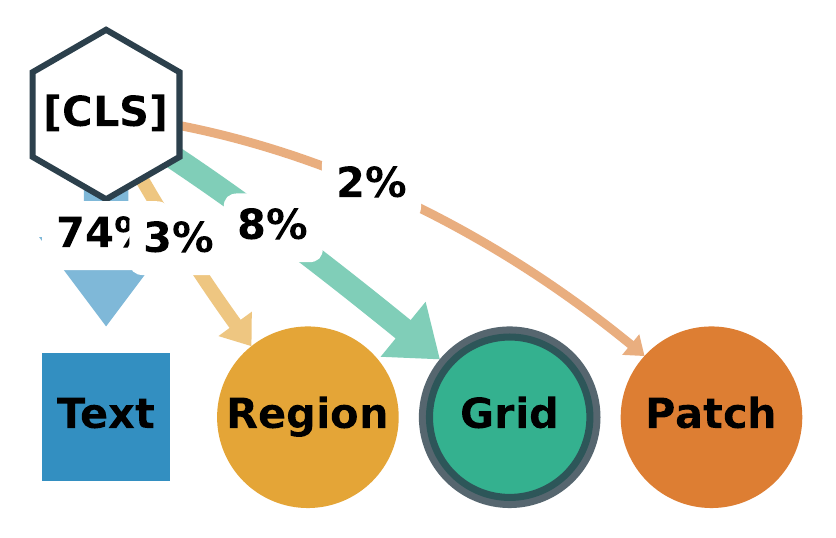}
        \caption{CLS Attention for SNLI-VE}
        \label{fig:ana:clsattention:snlive}
    \end{subfigure}
    \begin{subfigure}{.3\linewidth}
    \centering
        \includegraphics[width=.6\linewidth]{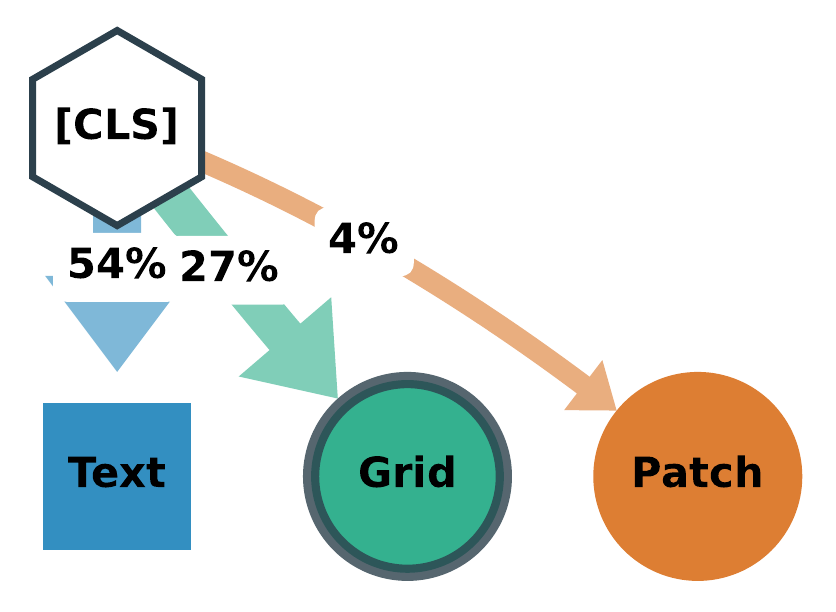}
        \caption{CLS Attention for Hateful Memes}
        \label{fig:ana:clsattention:hateful}
    \end{subfigure}
    
    \caption{CLS attention (in \%) to each modality/ VE averaged over all heads. We add an \textbf{outline} to the VE with the best single-VE results. Numbers do not add to 100\% because of CLS self-attention. We present the best multi-VE results here and all other results in Figure~\ref{fig:appendix:ana:clsattention} in the Appendix.}
\label{fig:ana:clsattention}
 \vspace{-1.5mm}
\end{figure*}

\begin{figure*}[!ht]
\centering
    \begin{subfigure}{.3\linewidth}
    \centering
        \includegraphics[width=.8\linewidth]{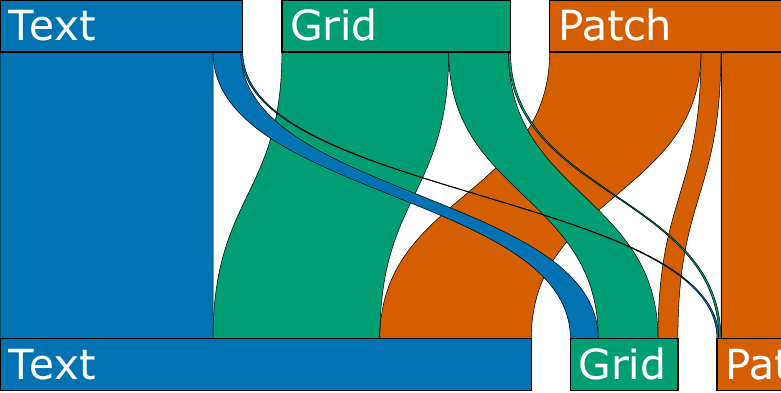}  
        \caption{Cross-Attention for Flickr30k}
        \label{fig:ana:crossattention:flickr}
    \end{subfigure}
    \begin{subfigure}{.3\linewidth}
    \centering
        \includegraphics[width=.8\linewidth]{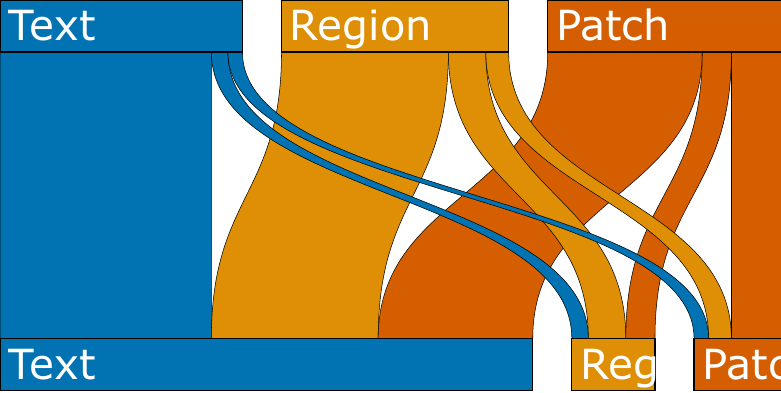}  
        \caption{Cross-Attention for MSCOCO}
        \label{fig:ana:crossattention:mscoco}
    \end{subfigure}
    \begin{subfigure}{.3\linewidth}
    \centering
        \includegraphics[width=.8\linewidth]{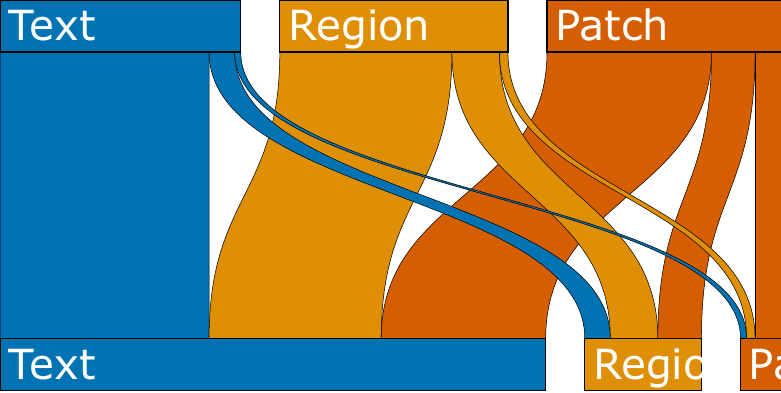}
        \caption{Cross-Attention for GQA}
        \label{fig:ana:crossattention:gqa}
    \end{subfigure}
    \\
    \begin{subfigure}{.3\linewidth}
    \centering
        \includegraphics[width=.8\linewidth]{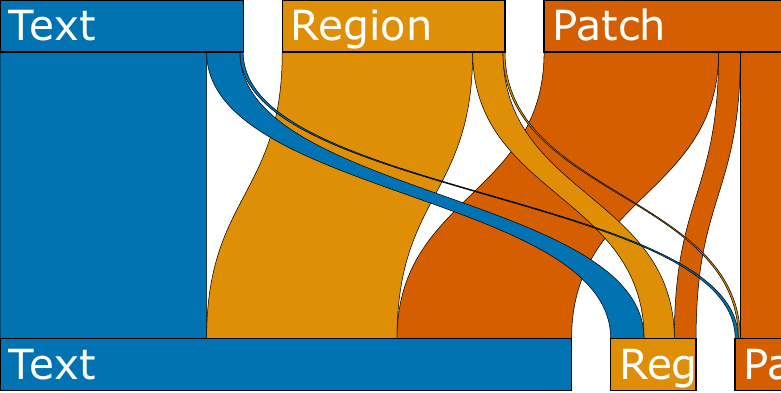}
        \caption{Cross-Attention for VQA}
        \label{fig:ana:crossattention:vqa}
    \end{subfigure}
    \begin{subfigure}{.3\linewidth}
    \centering
        \includegraphics[width=.8\linewidth]{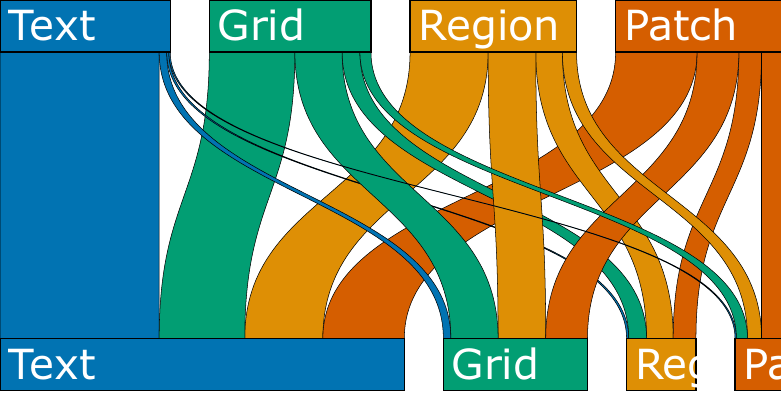}
        \caption{Cross-Attention for SNLI-VE}
        \label{fig:ana:crossattention:snlive}
    \end{subfigure}
    \begin{subfigure}{.3\linewidth}
    \centering
        \includegraphics[width=.8\linewidth]{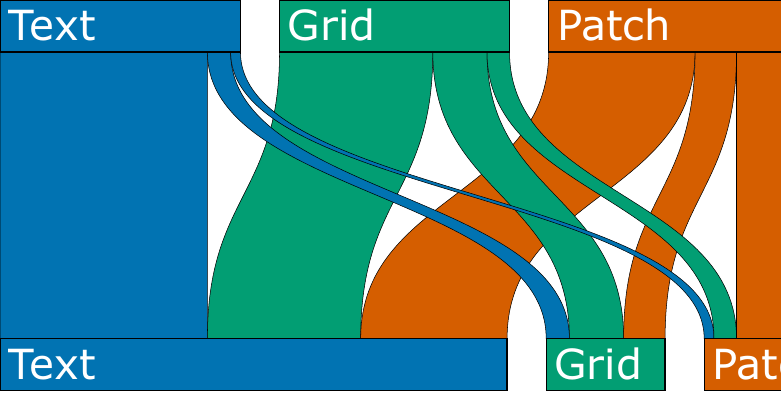}
        \caption{Cross-Attention for Hateful Memes}
        \label{fig:ana:crossattention:hateful}
    \end{subfigure}
    \caption{Cross-modal attention flow (in \%) from each modality/ VEs (top) to all modalities/ VEs (bottom). Flow is the sum of all attention weights between two modalities, averaged over all modality tokens and attention heads. We present the best multi-VE results here and all other results in Figure~\ref{fig:appendix:ana:cross} in the Appendix.}
\label{fig:ana:crossattention}
\vspace{-3mm}
\end{figure*}

The CLS token can be seen as the fused representation of the modalities that are used for the final classification \citep{cao_behind_2020}.
We can thus estimate which VEs are important for classification by considering which VEs the CLS token attends to.\footnote{This is only an estimation because the modalities combine information through attention, too.}
Following \citet{cao_behind_2020}, we compute the sum of attention from the CLS token to each modality and then average those scores over all heads.

We present the CLS attention for the best multi-VE setups in Figure~\ref{fig:ana:clsattention}. 
We see that for most tasks, the VE which performed best in the single-VE setup receives the majority of VE-attention. 
This suggests that one VE dominates in multi-VE setups while the others are complementary.

\subsection{Cross-Modal Attention Flow}
\label{sec:Cross-modal-attention}

The attention flow between the different modalities can indicate which VEs are used by the model to reason over the input.
We assume that more attention to a modality suggests that it contains useful information for others.
We compute the average attention flow between two modalities $M$ and $N$ for an attention head as $\frac{1}{|M|}\sum_{m \in M, n \in N}a_{m \rightarrow n}$
with $a_{m \rightarrow n}$ as the attention weight from token $m$ to $n$ (excluding CLS).
We average over all heads.

We present the attention flow for the best multi-VE setups in Figure~\ref{fig:ana:crossattention}. Similar to \S\ref{sec:CLS_Attetion}, the majority of attention is paid to the VE that achieved better results in the single-VE experiment for that task.

\begin{figure*}[!ht]
\centering
    \begin{subfigure}{.3\linewidth}
    \centering
        \includegraphics[width=.85\linewidth]{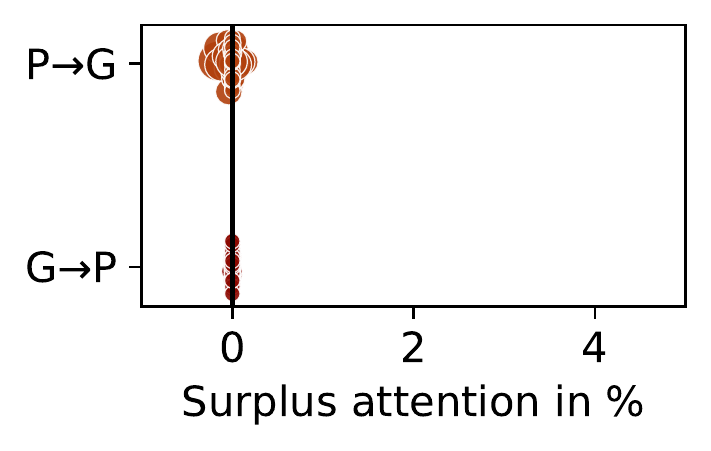}  
        \caption{Surplus Attention for Flickr30k}
        \label{fig:ana:crossattention:flickr}
    \end{subfigure}
    \begin{subfigure}{.3\linewidth}
    \centering
        \includegraphics[width=.85\linewidth]{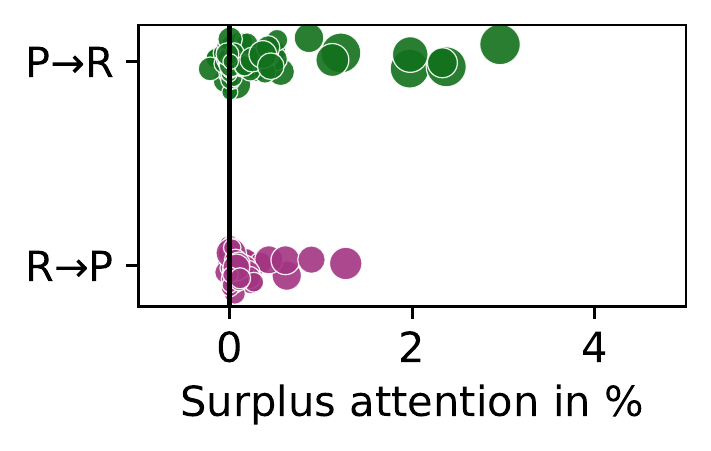}  
        \caption{Surplus Attention for MSCOCO}
        \label{fig:ana:crossattention:mscoco}
    \end{subfigure}
    \begin{subfigure}{.3\linewidth}
    \centering
        \includegraphics[width=.85\linewidth]{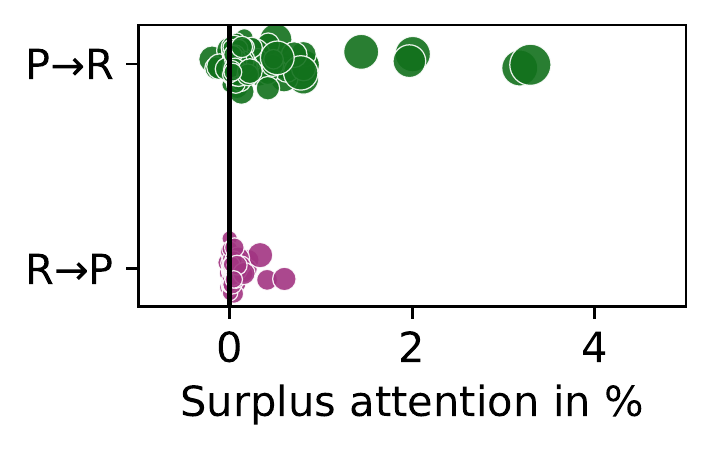}
        \caption{Surplus Attention for GQA}
        \label{fig:ana:crossattention:gqa}
    \end{subfigure}
    \\
    \begin{subfigure}{.3\linewidth}
    \centering
        \includegraphics[width=.85\linewidth]{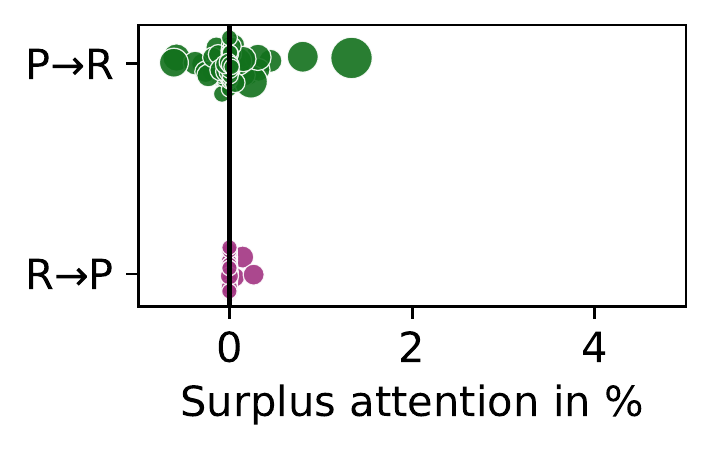}
        \caption{Surplus Attention for VQA}
        \label{fig:ana:crossattention:vqa}
    \end{subfigure}
    \begin{subfigure}{.3\linewidth}
    \centering
        \includegraphics[width=.85\linewidth]{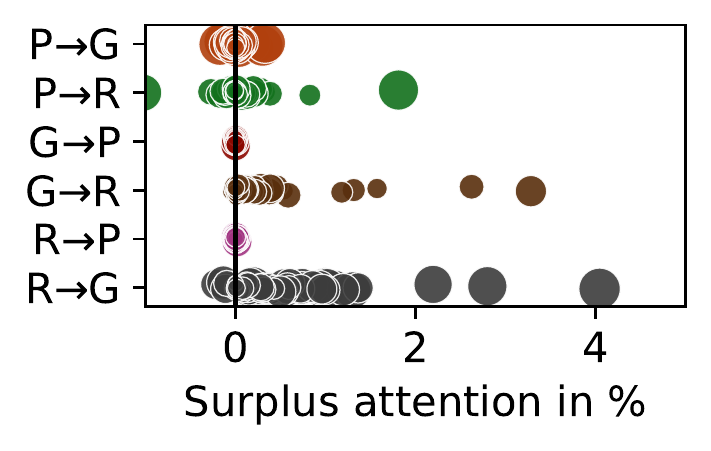}
        \caption{Surplus Attention for SNLI-VE}
        \label{fig:ana:crossattention:snlive}
    \end{subfigure}
    \begin{subfigure}{.3\linewidth}
    \centering
        \includegraphics[width=.85\linewidth]{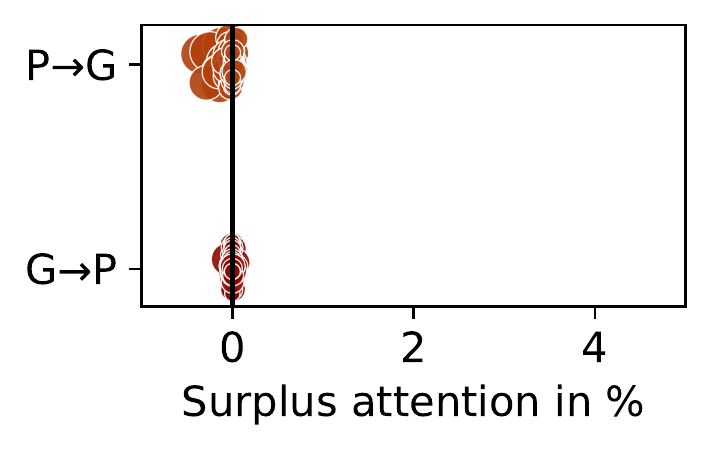}
        \caption{Surplus Attention for Hateful M.}
        \label{fig:ana:crossattention:hateful}
    \end{subfigure}
    \caption{Surplus attention of attention heads from one VE's tokens to another target VE's overlapping tokens compared to the other non-overlapping tokens of the target VE. \textbf{Dot size} represents the average total attention paid to the target VE by the respective head. Positive values indicate that the respective head attributes \textit{more} attention, negative values that \textit{less} attention is attributed  to overlapping tokens.
    We present the best multi-VE results here and all other results in Figure~\ref{fig:appendix:ana:crossimg} in the Appendix. 
    (Abbreviations: \textbf{R}egion, \textbf{G}rid, \textbf{P}atch).}
\label{fig:ana:crossimg}
\end{figure*}
\subsection{Overlapping Token Surplus Attention}
\label{sec:overlapping_attention}
The attention flow between different VEs' visual tokens that \textit{overlap}---i.e. encode the same part of the image---can tell us if the model combines the VEs to complement the image representation.
For each attention head we therefore compute the average per-token attention from a token $t$ of \textit{one} VE to overlapping tokens $I_{|t}$ of \textit{another} VE, and compare this---i.e. compute the surplus---to the  non-overlapping tokens of that VE  $I\textbackslash t$ as $(\frac{1}{|I_{|t}|}\sum_{i \in I_{|t}}a_{t \rightarrow i}) - (\frac{1}{|I\textbackslash t|}\sum_{i \in I\textbackslash t}a_{t \rightarrow i} )$
with $a_{t \rightarrow i}$ being the attention weight between the tokens.
We average over all tokens to get the surplus attention of an attention head for a VE pair.

We present the results for the best performing setups in Figure~\ref{fig:ana:crossimg}.
For most VEs we can identify heads---indicated by larger dot sizes---which attend particularly to \textit{another} VE's tokens. For most settings, these heads are also those which attend to the overlapping tokens of the respective other VE. While we witness a large surplus in attention for overlapping tokens between \textit{Region} and \textit{Grid}/\textit{Patch}, this is not the case for \textit{Grid-Patch}. This indicates that the complementarity of \textit{Region} features is higher for the respective other VEs, which provides more evidence that training the VE on different data and objectives is important.

\subsection{Visual Entity Grounding}
\label{sec:grounding}
\begin{figure}[]
    \centering
    \begin{subfigure}{.47\linewidth}
    \centering
        \includegraphics[width=.99\linewidth]{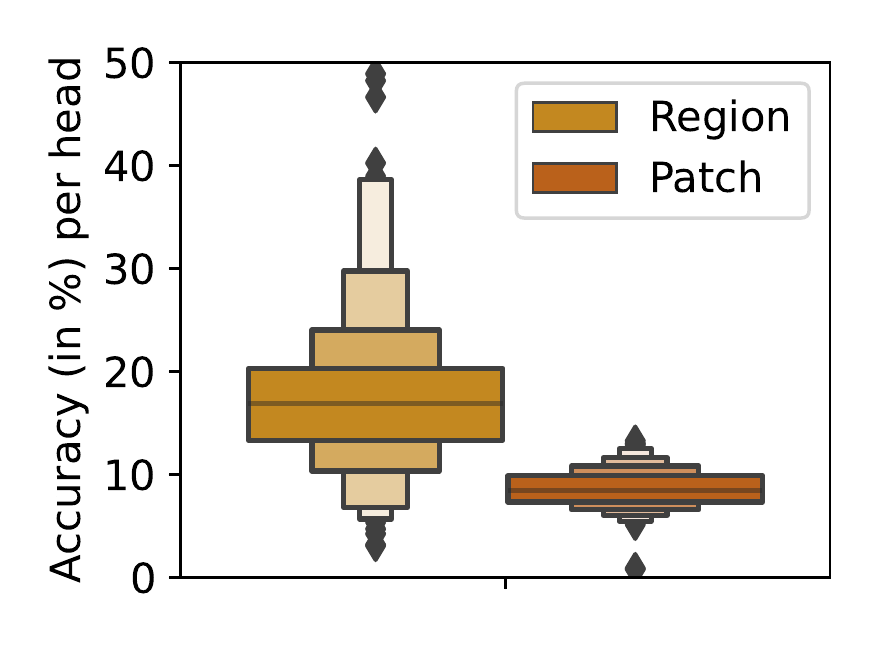}
        \caption{Visual Grounding GQA}
        \label{fig:ana:grounding:gqa}
    \end{subfigure}
    \begin{subfigure}{.47\linewidth}
    \centering
        \includegraphics[width=.99\linewidth]{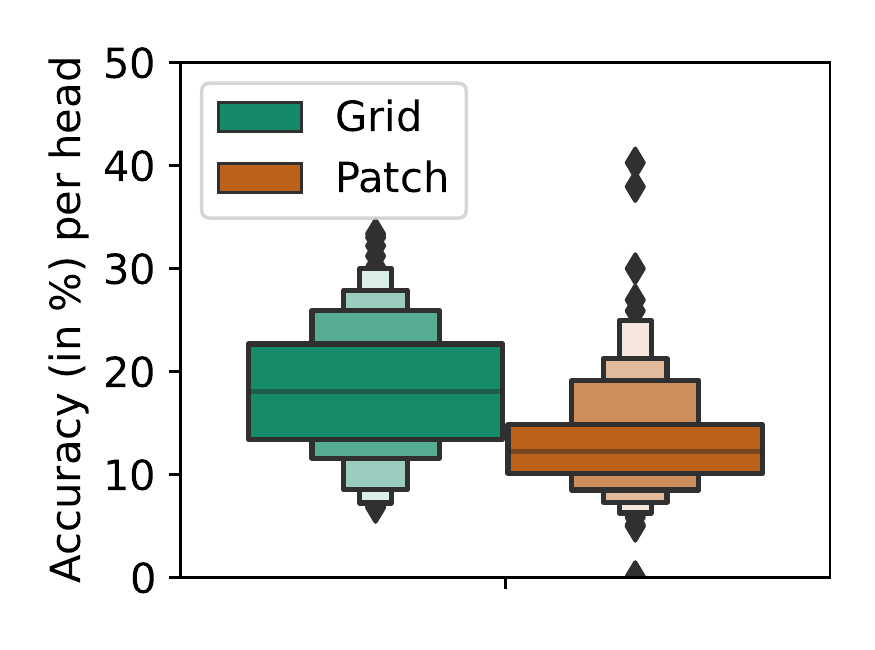}  
        \caption{Visual Ground. Flickr30k}
        \label{fig:ana:grounding:flickr}
    \end{subfigure}
    \caption{Visual Entity Grounding accuracy of all attention heads. An entity is grounded correctly to a VE if the attention weight from the phrase to the matching visual tokens is the highest over all the VE's tokens. Other results can be found in Figure~\ref{fig:appendix:ana:grounding} in the Appendix.}
\label{fig:ana:grounding}
\end{figure}
Visual grounding is the task of matching text phrases to their corresponding parts in the image.
To analyze whether or not there are dominant VEs which learn to ground, we follow  \citet{li_what_2020} and count how often the highest attention weight from a text phrase is assigned to the corresponding visual token.
We use the gold phrase-to-bounding box annotations available for Flickr30k \citep{plummer_flickr30k_2015} and GQA.
Formally, a head correctly grounds a phrase to the gold box $g$ if the \textit{maximum attention} from the last phrase token $t$ to any of a VE's tokens $I$ goes to any token $I_{|g}$ overlapping with the gold box,\footnote{We consider all visual tokens with Intersection over Union between the token box and the gold box over 0.5 for region tokens (based on \citet{li_what_2020}) and over 0.1 for grid and patch tokens (because a single object can be distributed over many tokens due to the uniform grid).} i.e. if $\mathrm{arg\,max}_{i \in I} a_{t \rightarrow i} \in I_{|g}$. 
We calculate the accuracy by counting the number of correct  groundings where $I_{|g}$ is not empty.

We report the results for all heads of the best GQA and Flickr30k models in Figure~\ref{fig:ana:grounding}. We can see that the accuracy of the dominant VE (\textit{Grid} and \textit{Region}, respectively) is generally higher than for \textit{Patch}.
While there is a clear pattern of the dominant VE achieving significantly higher accuracy, the complementary VE also achieves  accuracies beyond 10\%, indicating that the  model learns to reason over, and utilize \textit{both} VE's representations.

\begin{figure}[]
    \begin{subfigure}{.47\linewidth}
    \centering
        \includegraphics[width=.9\linewidth]{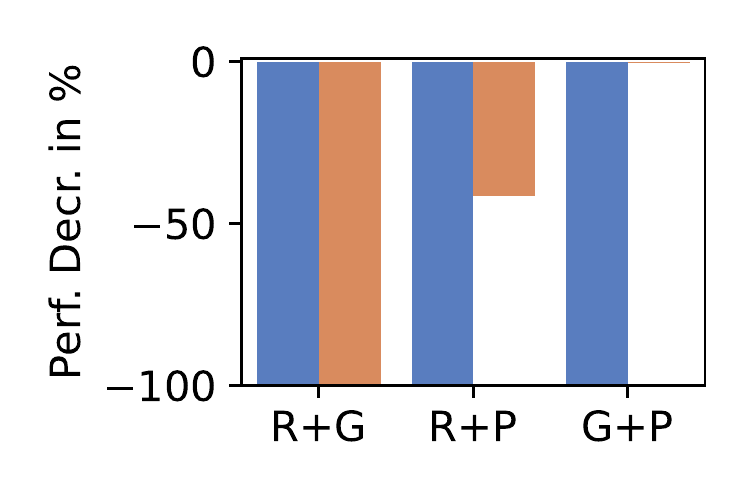}  
        \caption{Flickr30k VE Dropping}
        \label{fig:ana:grounding:flickr}
    \end{subfigure}
    \begin{subfigure}{.47\linewidth}
    \centering
        \includegraphics[width=.9\linewidth]{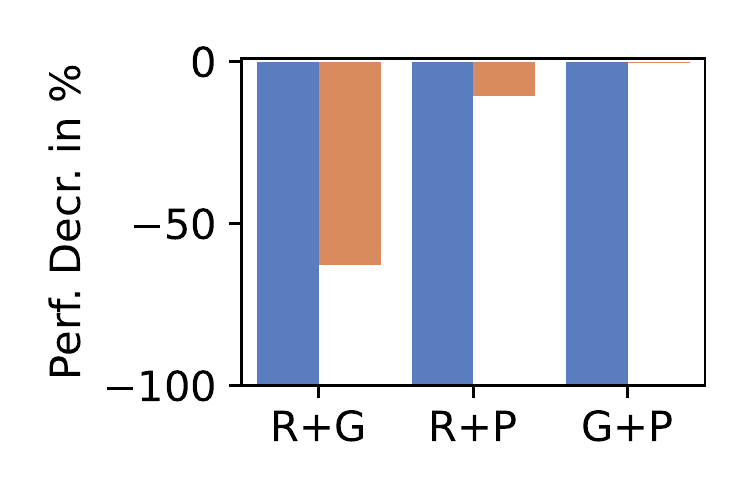}  
        \caption{MSCOCO VE Dropping}
        \label{fig:ana:grounding:mscoco}
    \end{subfigure}
    \\
    \begin{subfigure}{.47\linewidth}
    \centering
        \includegraphics[width=.9\linewidth]{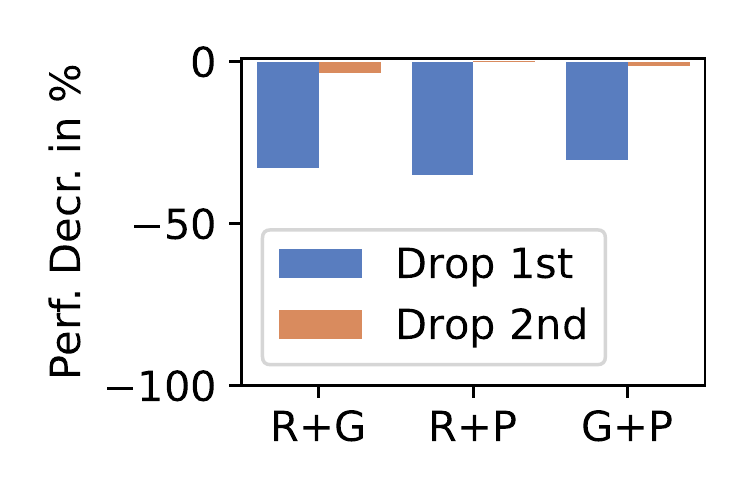}  
        \caption{GQA VE Dropping}
        \label{fig:ana:grounding:gqa}
    \end{subfigure}
    \begin{subfigure}{.47\linewidth}
    \centering
        \includegraphics[width=.9\linewidth]{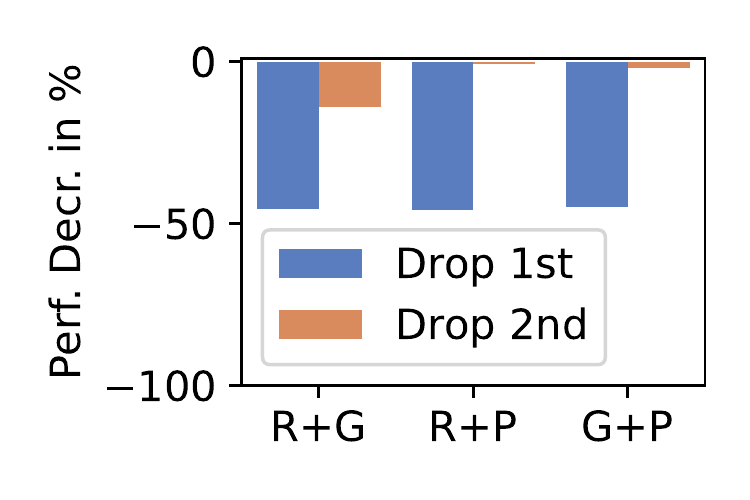}  
        \caption{VQA VE Dropping}
        \label{fig:ana:grounding:vqa}
    \end{subfigure}
    \\
    \begin{subfigure}{.47\linewidth}
    \centering
        \includegraphics[width=.9\linewidth]{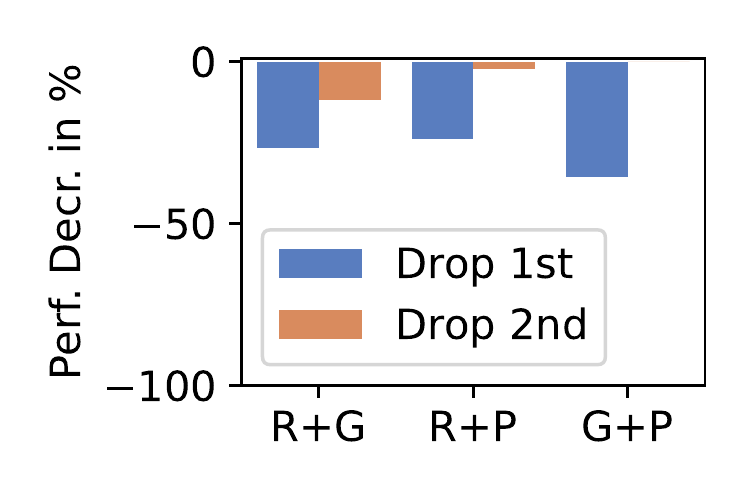}  
        \caption{SNLI-VE VE Dropping}
        \label{fig:ana:grounding:snlive}
    \end{subfigure}
    \begin{subfigure}{.47\linewidth}
    \centering
        \includegraphics[width=.9\linewidth]{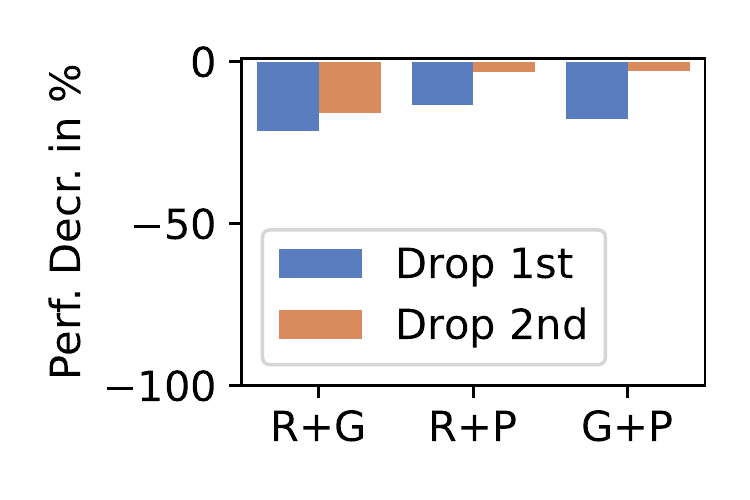}  
        \caption{Hateful M. VE Dropping}
        \label{fig:ana:grounding:snlive}
    \end{subfigure}
    \caption{Relative performance decrease of 2-encoder models after dropping the entire first or second VE from the input compared to evaluation with both encoders in use.
    (Abbreviations: \textbf{R}egion, \textbf{G}rid, \textbf{P}atch).}
\label{fig:ana:pruning}
\end{figure}

\subsection{VE-Dropout}
\label{sec:ana:dropouttrain}
Our analyses in the previous sections suggest that there are dominant and complementary encoders, the former contributing the most to the model's performance on the target task.  To further evaluate the importance of the respective VEs we experiment with dropping all VE-specific features during test time. The results in Figure~\ref{fig:ana:pruning} show that dropping the dominant encoder results in a catastrophic performance decrease, especially for the retrieval tasks. While QA and reasoning tasks have a  20\% to 40\% decrease in performance, for retrieval tasks R@1 decreases by almost 100\%. %

\begin{figure}[]
    \centering
    \begin{subfigure}{.47\linewidth}
    \centering
        \includegraphics[width=.9\linewidth]{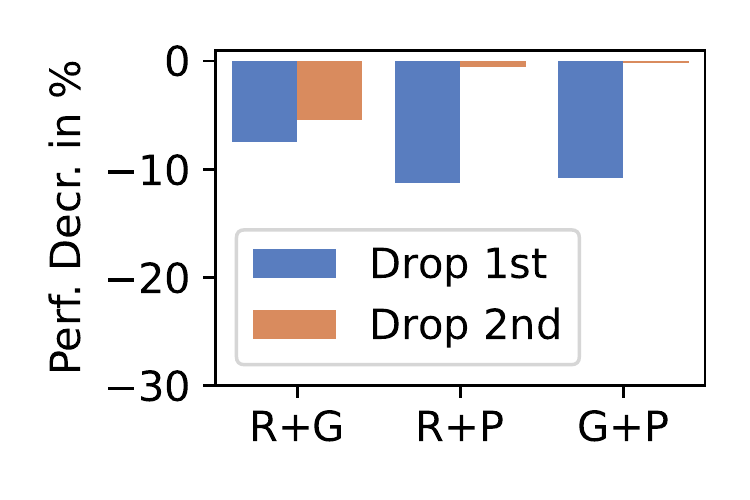}  
        \caption{MSCOCO VE Dropping}
        \label{fig:ana:grounding:mscoco}
    \end{subfigure}
    \begin{subfigure}{.47\linewidth}
    \centering
        \includegraphics[width=.9\linewidth]{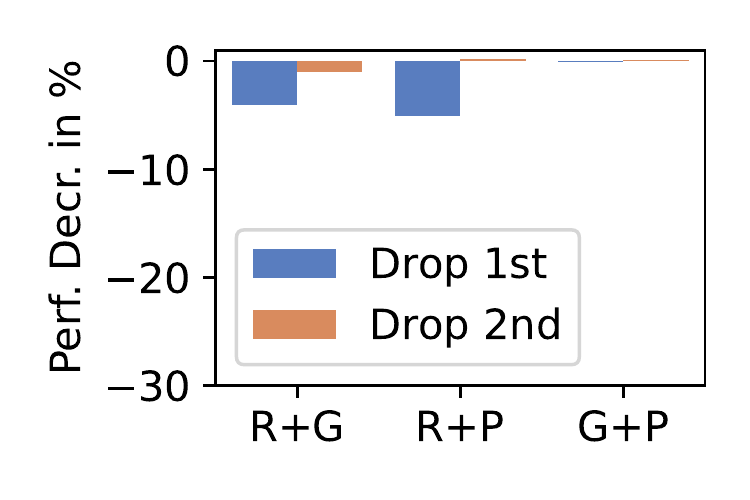}  
        \caption{GQA VE Dropping}
        \label{fig:ana:grounding:gqa}
    \end{subfigure}
    \caption{Relative performance decrease of 2-encoder models as in Figure~\ref{fig:ana:pruning} \textbf{after} VE-Dropout Training. Other tasks in Figure~\ref{fig:appendix:ana:pruning} in the Appendix.
    }
\label{fig:ana:drop:pruning}
\end{figure}

However, the detrimental performance of dropping out of the VE features at test time might be a result of the multi-VE models never being trained for this setting. Consequently, we train the 2-encoder models with VE-wise dropout per batch.\footnote{Uniformly dropping the \textit{first, second}, or \textit{no} VE.} We hypothesize that this would force the model to take the complementary VE into account while being more robust during inference. As reported in Figure~\ref{fig:ana:drop:pruning}, the robustness in terms of dropping VEs improves, however, we see a slight drop in the final task performance, as reported in Table~\ref{tbl:res:scores}.\footnote{We  notice no significant changes in the attention patterns after VE-Dropout training (see Appendix~\ref{sec:appendix:ana:drop}).}

\section{Discussion and Future Directions}
Our analyses demonstrate that, while combining multiple VEs consistently outperforms single VE setups, there is not a single VE or a fixed strategy on combining VEs that works  best for all tasks.
In particular, simply ensembling all VEs is rarely the optimal choice; consequently,  best-performing combinations of VEs need to be identified for each individual task
(\S~\ref{sec:results_discussion}). 
By further analyzing the attention patterns, we find a clear dominating VE (\S~\ref{sec:ana:dropouttrain}, Figures~\ref{fig:ana:pruning} \& \ref{fig:ana:drop:pruning}) that both the [CLS] (\S~\ref{sec:CLS_Attetion}, Figure~\ref{fig:ana:clsattention}) and the multimodal tokens (\S~\ref{sec:Cross-modal-attention}, Figure~\ref{fig:ana:crossattention}) predominantly attend to, whereas the secondary VEs provide complementary information, supporting the model's overall performance.  The complementarity of VEs is highlighted by analyzing the cross-modal attention patterns for overlapping parts of the image (\S~\ref{sec:overlapping_attention}, Figure~\ref{fig:ana:crossimg}). VEs trained on different data and objectives (e.g. \textit{Grid} and \textit{Region}) cross-attend to the tokens of the respective \textit{other} VE that encode the same parts of the image, aggregating their information.
Further,  the model learns to visually ground the text representations to all VEs, as demonstrated by their attention patterns (\S~\ref{sec:grounding}, Figure~\ref{fig:ana:grounding}), and the text modality aggregates information from different VEs.

In summary, 
our results indicate that  VEs,  trained on different data and objectives, encode complementary information, resulting in improvements over approaches which only utilize a single VE. This indicates that VEs, explicitly designed for V+L tasks---e.g. by incorporating more diverse training data and objectives during pre-training---have the potential to significantly impact the performance on the target V+L tasks. 

\section{Conclusion}
In this work, we investigated whether different VEs---based on repurposed pre-trained vision models---encode complementary information, which improves the performance on downstream V+L tasks. We experimented with three popular VE classes \textit{Region}, \textit{Grid}, and \textit{Patch}, and trained models with all possible combinations on six different V+L tasks. While combining VEs improve over single-VE setups, our further analysis of attention patterns reveals that diverse VEs encode complementary information, which motivates future work on designing VEs explicitly for V+L tasks---e.g. by incorporating more diverse datasets, and training objectives.

\FloatBarrier

\section*{Limitations}

The main limitation of our concatenation-based multi-VE models is efficiency:
The models are significantly slower than single-VE models because of the additional visual tokens in the input; the 3-VE model requires almost twice the time to train (in real time, not training steps) compared to the single-VE models. Also, in cases where images are not pre-encoded, multi-VE setups are significantly slower at inference time.
However, as mentioned before, we concatenate the tokens for analysis purposes only (\S\ref{sec:ana}) and leave more efficient alternatives like resampling~\citep{alayrac_flamingo_2022} to the future.

Several limitations could be investigated in the future, assuming access to a larger computational budget:
\begin{enumerate}
    \item We focused on single-stream Transformers and did not take into account dual-stream or other multimodal Transformer architectures. 
    \item We only experimented with three popular VEs (one version per VE class). There are many other VEs we could investigate in the future.
    \item We do not pre-train our multimodal models on intermediate, auxiliary multimodal tasks \cite[][\textit{inter alia}]{tan_lxmert_2019, lu_vilbert_2019,chen_uniter_2020} as achieving state-of-the-art is not our goal.  
\end{enumerate}

\section*{Acknowledgements}
This work has been funded by the German Federal Ministry of Education and Research (BMBF) under the promotional reference 13N15897 (MISRIK) and by the LOEWE initiative (Hesse, Germany) within the emergenCITY center. 

We thank Mert Tiftikci, Pooneh Mousavi, and Neha Warikoo for insightful feedback and suggestions on a draft of this paper.

\bibliography{anthology,custom}

\newpage
\appendix

\section{Training and Hyperparameters}
\label{sec:appendix:hyper}
We report the hyperparameters along with the task-specific training details.

\subsection{Hyperparameters}
We report our hyperparameters in Table~\ref{tab:appendix:hyperparam1},\ref{tab:appendix:hyperparam2}.
For each task, we select the learning rate in $\{2e-5, 3e-5, 5e-5 \}$ with the best validation performance for the model trained with all three VEs.
We train all VE combinations for one task with the same hyperparameters.
We use the training checkpoint with the best validation performance (computed each epoch) for testing.
\begin{table}[!ht]
    \centering
    \footnotesize
    \begin{tabular}{l r}
    \toprule
    \bf Name & \bf Value \\
    \midrule
    Optimizer     & AdamW \\
    & \cite{loshchilov_decoupled_2019} \\
    Schedule     & linear \\
    Warmup steps & 5\% \\
    Weight decay & 0.05 \\
    Batch size &   64 \\
    Max. text sequence & 96 \\
    \bottomrule
    \end{tabular}
    \caption{Shared hyperparameters used during training for the different tasks.}
    \label{tab:appendix:hyperparam1}
\end{table}
\begin{table}[ht!]
    \centering
    \footnotesize
    \begin{tabular}{l rr}
    \toprule
    \bf Task & \bf Learning Rate & \bf Epochs \\
    \midrule
    Flickr30k & 2e-5 & 10  \\
    MSCOCO & 2e-5 & 3 \\
    GQA & 5e-5 & 6 \\
    VQA & 5e-5 & 10 \\
    SNLI-VE & 5e-5 & 10 \\
    Hateful Memes & 3e-5 & 50 \\
    \bottomrule
    \end{tabular}
    \caption{The per-task hyperparameters used during training.}
    \label{tab:appendix:hyperparam2}
\end{table}

\subsection{Task Details}
We describe the training details for each task.
Unless noted otherwise, we use the standard splits of all tasks as described in \citet{li_oscar_2020,zhang_vinvl_2021}.

\paragraph{Flickr30k, MSCOCO:}
We use a cross-encoder following \citet{li_oscar_2020,zhang_vinvl_2021}.
The model either receives the caption and the paired image or a random image (each with 50\% chance). The task is to predict if the caption and image match.
We use cross entropy as loss as the training objective.

During evaluation, we compute the logits for all possible image-caption pairs and use these scores as ranking to compute the recall at $k$.
We evaluate MSCOCO on the 1k image test set.

\paragraph{GQA, VQA, SNLI-VE, Hateful Memes:}
We train all three tasks as standard classification task with cross entropy loss.
For GQA, each class corresponds to a label appearing in the train, test, or validation set.
We also use the \textit{balanced} training data for GQA as it produces similar results to the much larger \textit{unbalanced} training set with a fraction of the training time.
For VQA, we follow \citet{li_oscar_2020} and use the top-3000 labels for classification and we train the model with a multilabel objective using the relevance scores as soft probabilities.
For testing, we use the maximum logit as the single predicted class.

\setcounter{topnumber}{2}
\setcounter{bottomnumber}{2}
\setcounter{totalnumber}{4}
\renewcommand{\topfraction}{0.85}
\renewcommand{\bottomfraction}{0.85}
\renewcommand{\floatpagefraction}{0.7}

\section{Full Analysis Results}
\label{sec:appendix:plot}
\begin{figure*}[!ht]
\centering
    \begin{subfigure}{.99\linewidth}
    \centering
        \includegraphics[width=.23\linewidth]{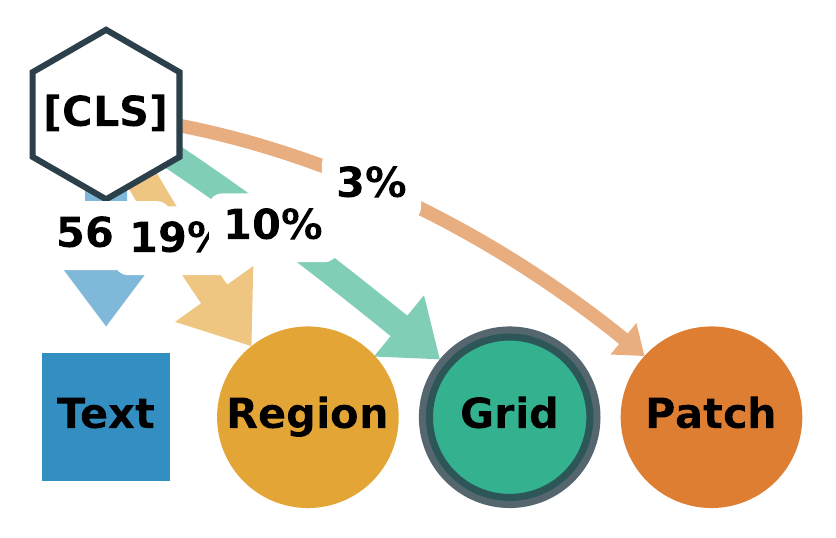}
        \includegraphics[width=.23\linewidth]{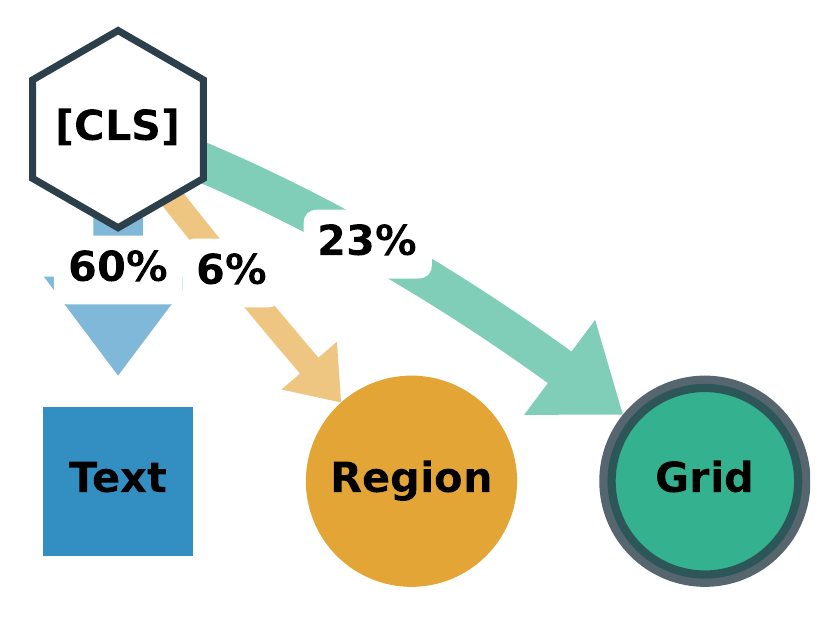}
        \includegraphics[width=.23\linewidth]{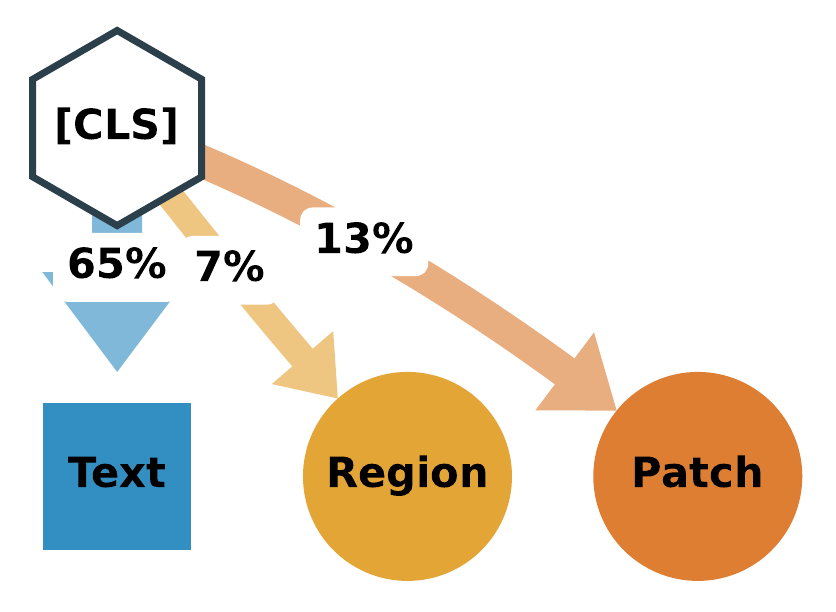}
        \includegraphics[width=.23\linewidth]{media/plots/cls_graph/Flickr30k+Grid,Patch_graph.pdf}
        \caption{CLS Attention Flickr30k}
    \end{subfigure}
    \begin{subfigure}{.99\linewidth}
    \centering
        \includegraphics[width=.23\linewidth]{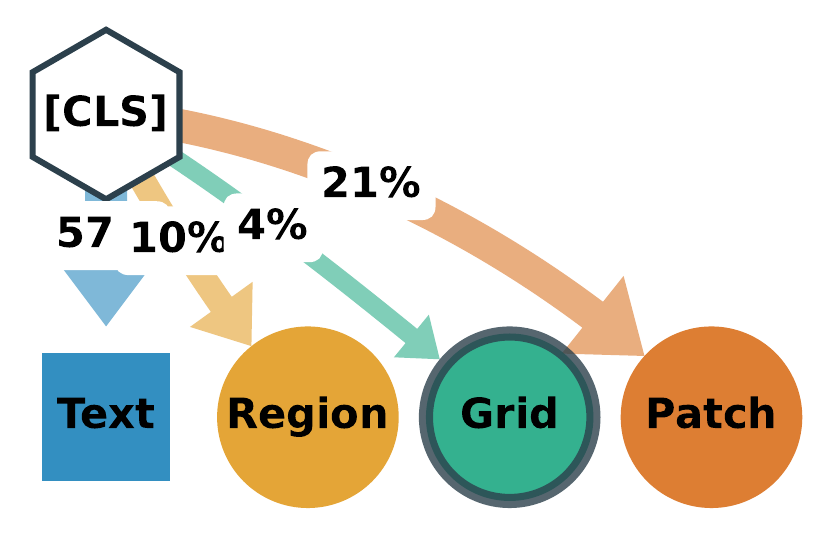}
        \includegraphics[width=.23\linewidth]{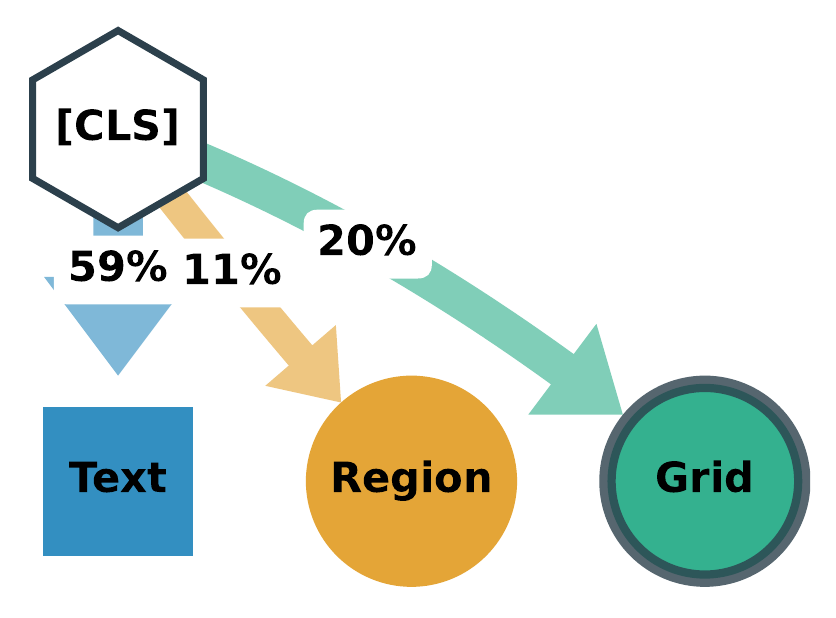}
        \includegraphics[width=.23\linewidth]{media/plots/cls_graph/MSCOCO+Region,Patch_graph.pdf}
        \includegraphics[width=.23\linewidth]{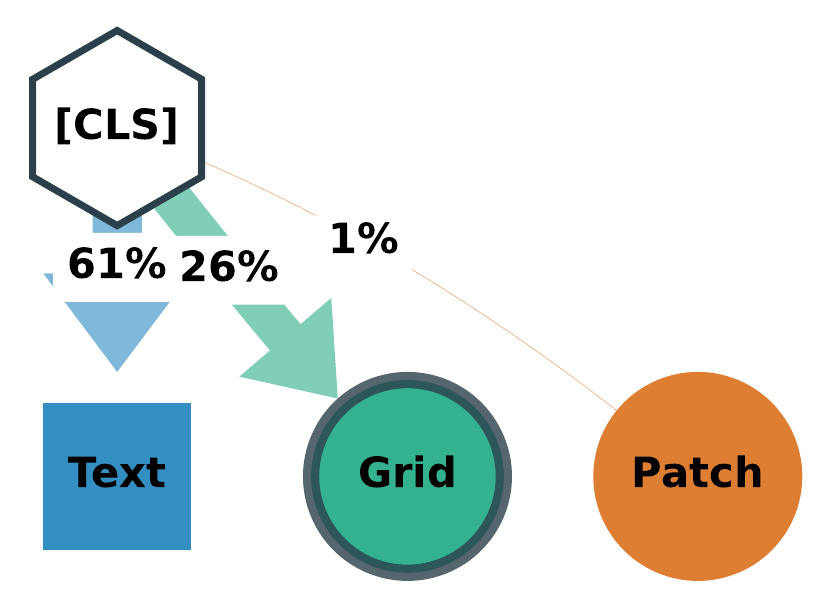}
        \caption{CLS Attention MSCOCO}
    \end{subfigure}
    \begin{subfigure}{.99\linewidth}
    \centering
        \includegraphics[width=.23\linewidth]{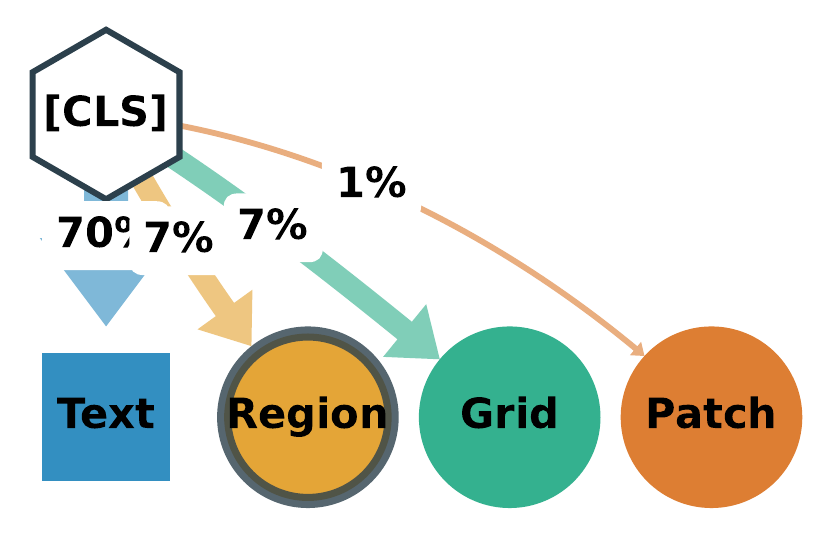}
        \includegraphics[width=.23\linewidth]{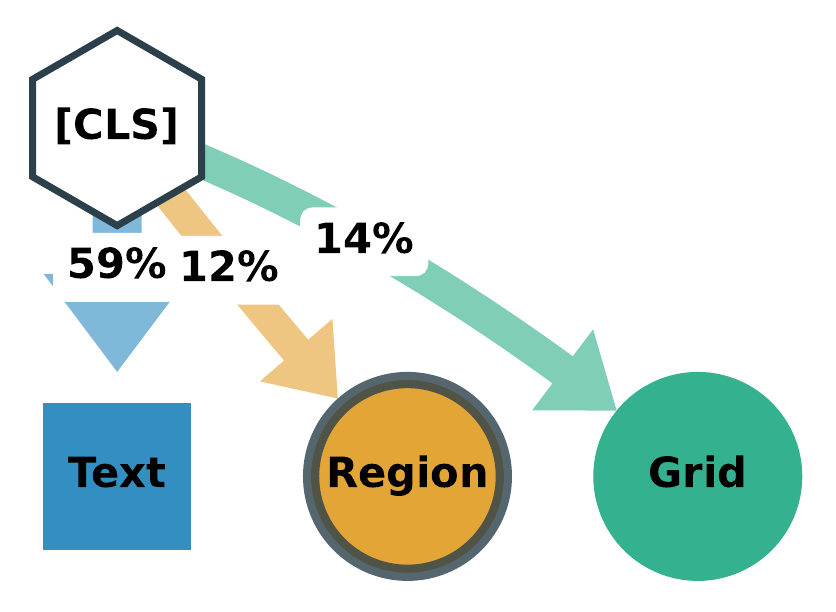}
        \includegraphics[width=.23\linewidth]{media/plots/cls_graph/GQA+Region,Patch_graph.pdf}
        \includegraphics[width=.23\linewidth]{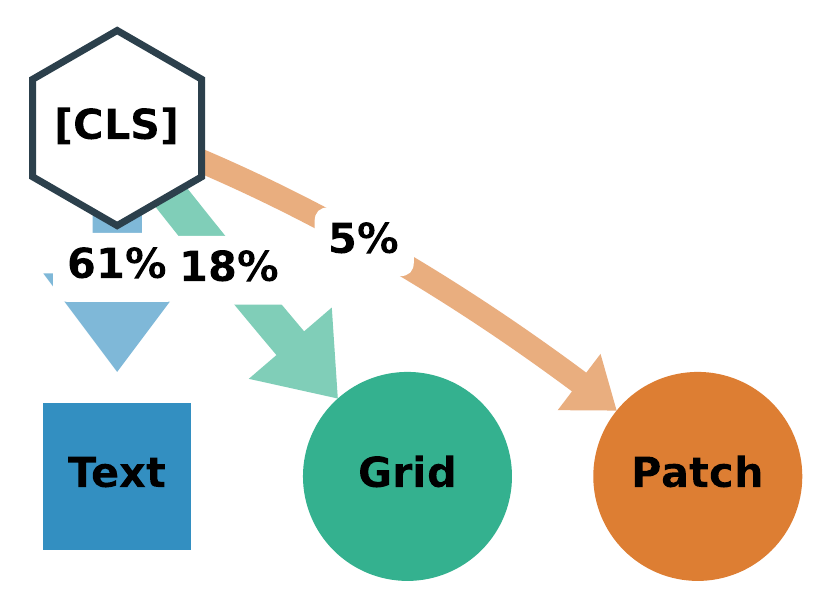}
        \caption{CLS Attention GQA}
    \end{subfigure}
    \begin{subfigure}{.99\linewidth}
    \centering
        \includegraphics[width=.23\linewidth]{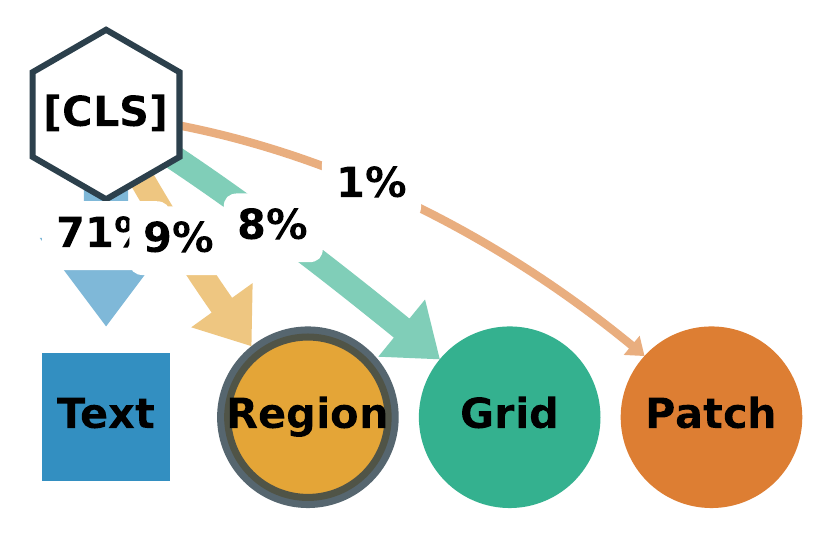}
        \includegraphics[width=.23\linewidth]{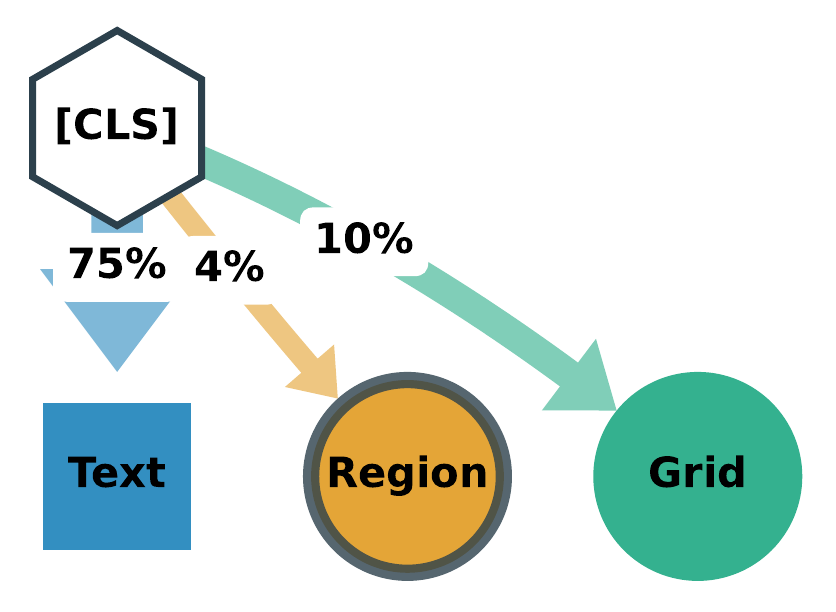}
        \includegraphics[width=.23\linewidth]{media/plots/cls_graph/VQA+Region,Patch_graph.pdf}
        \includegraphics[width=.23\linewidth]{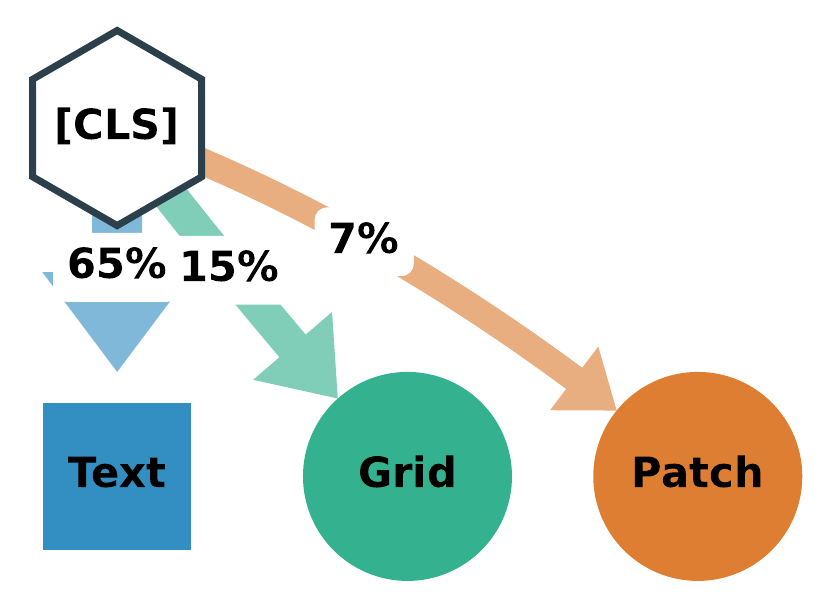}
        \caption{CLS Attention VQA}
    \end{subfigure}
    \begin{subfigure}{.99\linewidth}
    \centering
        \includegraphics[width=.23\linewidth]{media/plots/cls_graph/SNLI-VE+Region,Grid,Patch_graph.pdf}
        \includegraphics[width=.23\linewidth]{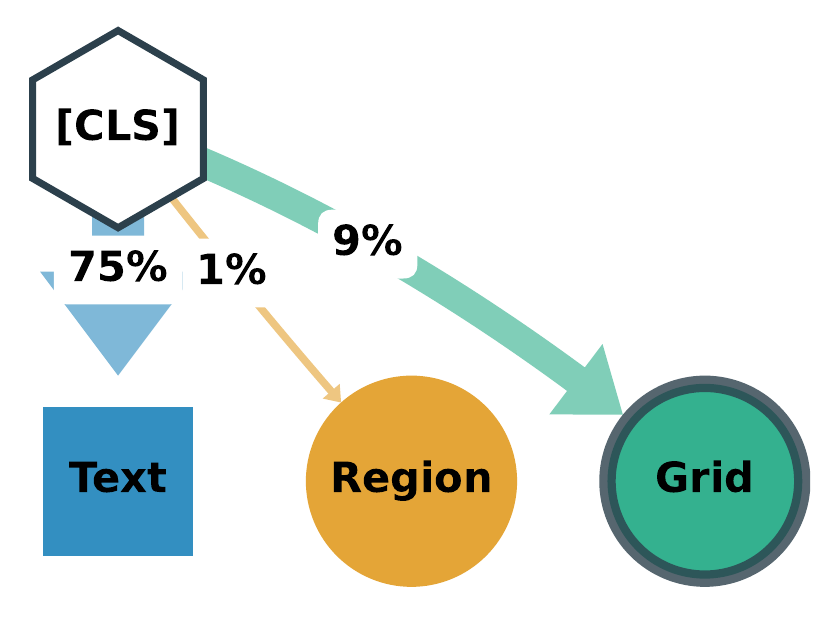}
        \includegraphics[width=.23\linewidth]{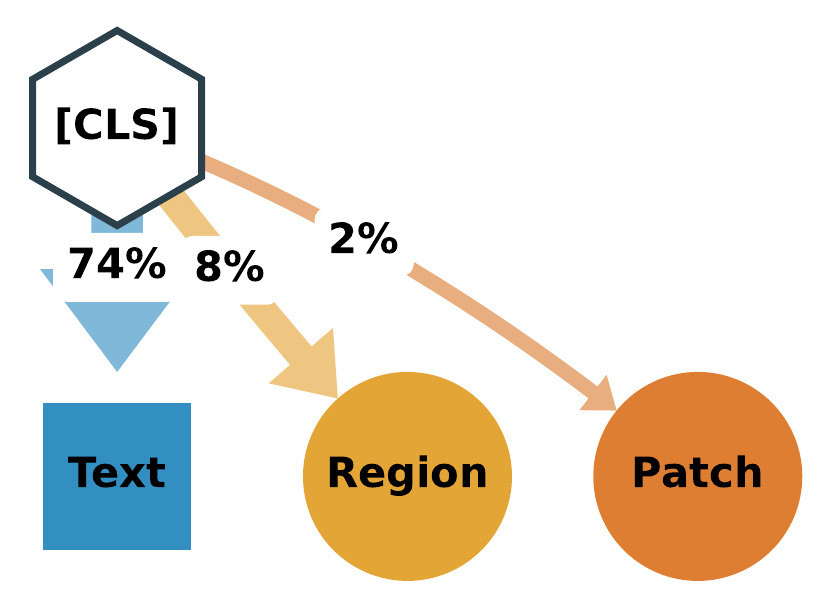}
        \includegraphics[width=.23\linewidth]{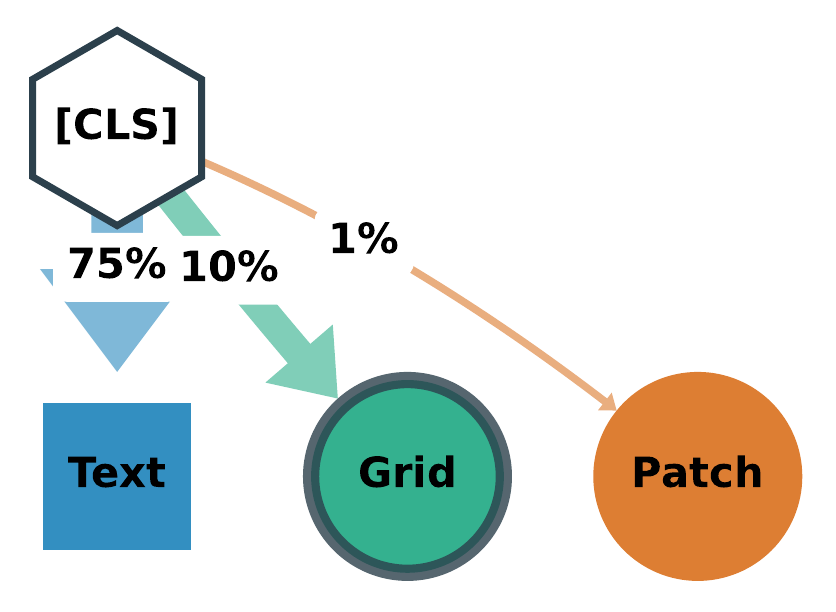}
        \caption{CLS Attention SNLI-VE}
    \end{subfigure}
    \begin{subfigure}{.99\linewidth}
    \centering
        \includegraphics[width=.23\linewidth]{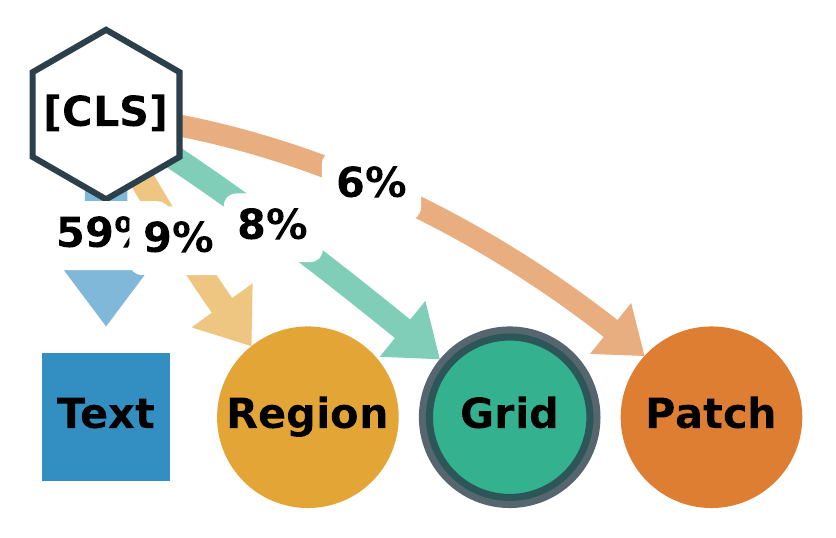}
        \includegraphics[width=.23\linewidth]{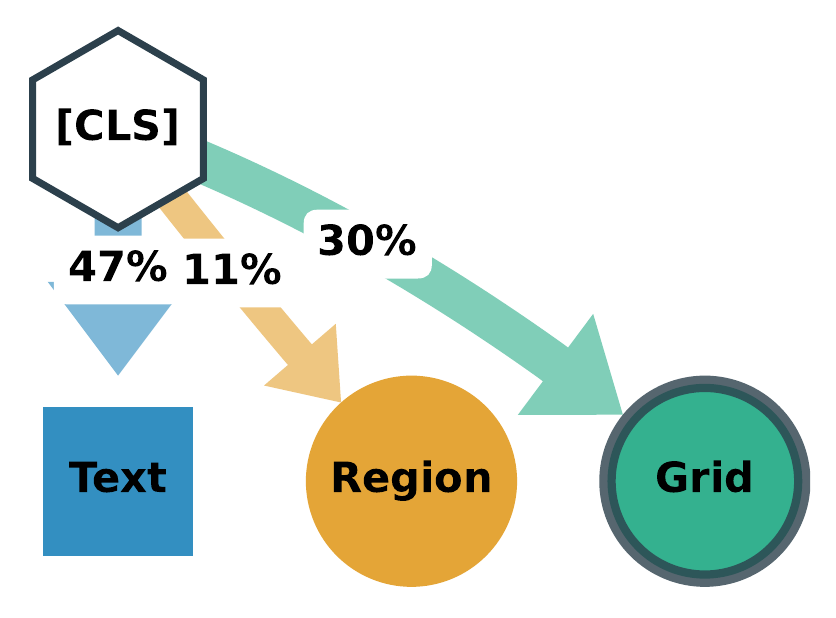}
        \includegraphics[width=.23\linewidth]{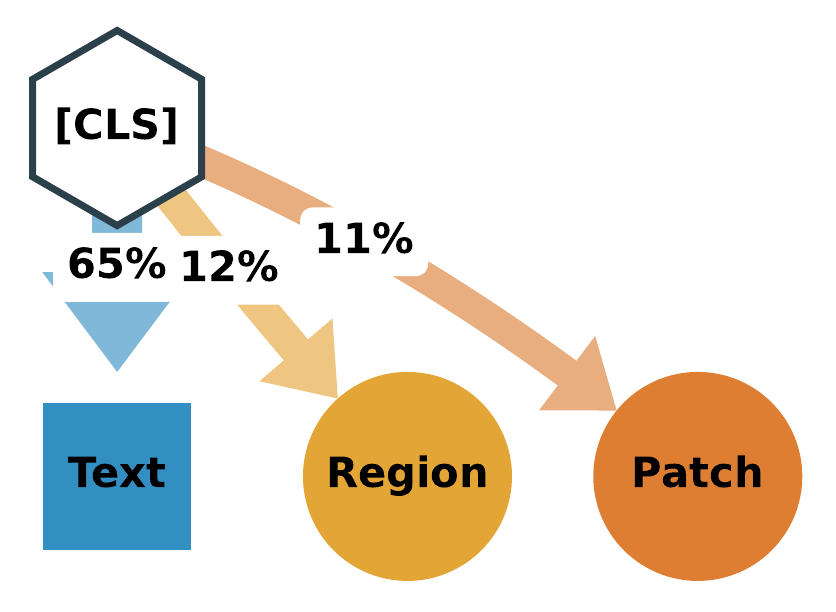}
        \includegraphics[width=.23\linewidth]{media/plots/cls_graph/Hateful+Grid,Patch_graph.pdf}
        \caption{CLS Attention Hateful Memes}
    \end{subfigure}
    \caption{CLS attention weights (in \%) averaged over all heads to the modalities. Numbers do not add to 100\% because of CLS self-attention.}
\label{fig:appendix:ana:clsattention}
\end{figure*}

\begin{figure*}[!ht]
\centering
    \begin{subfigure}{.99\linewidth}
    \centering
        \includegraphics[width=.23\linewidth]{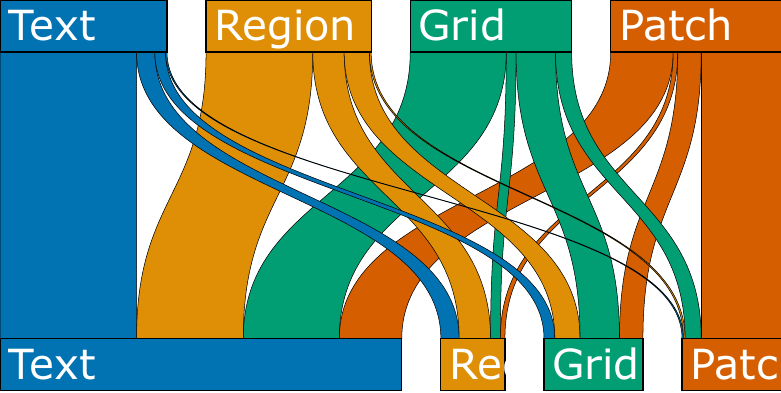}
        \includegraphics[width=.23\linewidth]{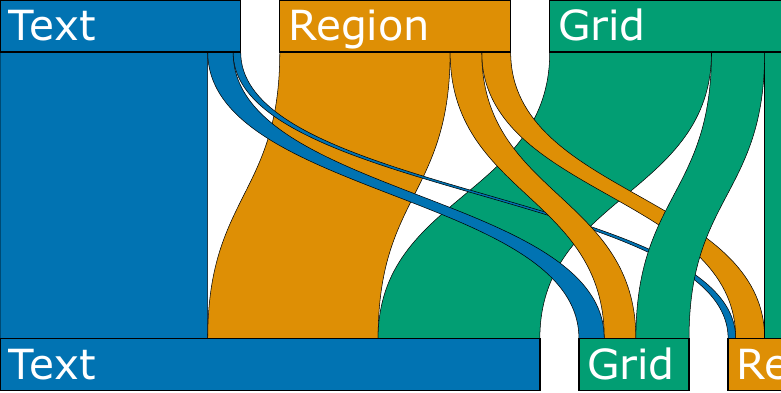}
        \includegraphics[width=.23\linewidth]{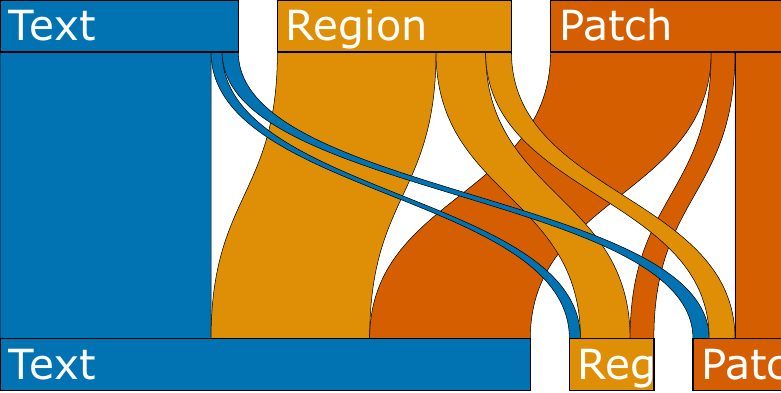}
        \includegraphics[width=.23\linewidth]{media/plots/cross_head/Flickr30k+Grid,Patch_graph.pdf}
        \caption{Cross-Attention for Flickr30k}
    \end{subfigure}
    \begin{subfigure}{.99\linewidth}
    \centering
        \includegraphics[width=.23\linewidth]{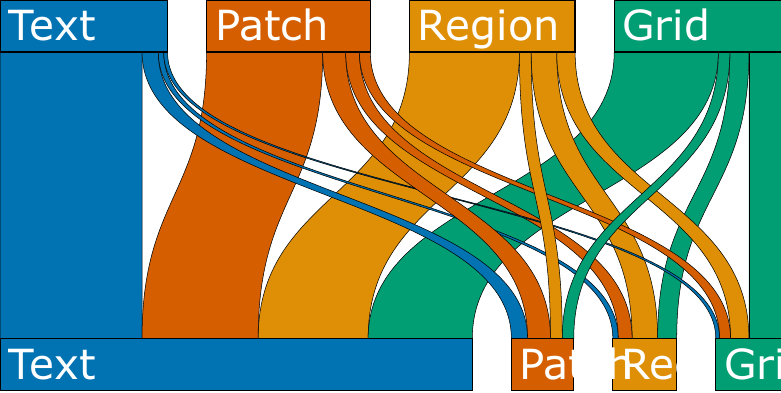}
        \includegraphics[width=.23\linewidth]{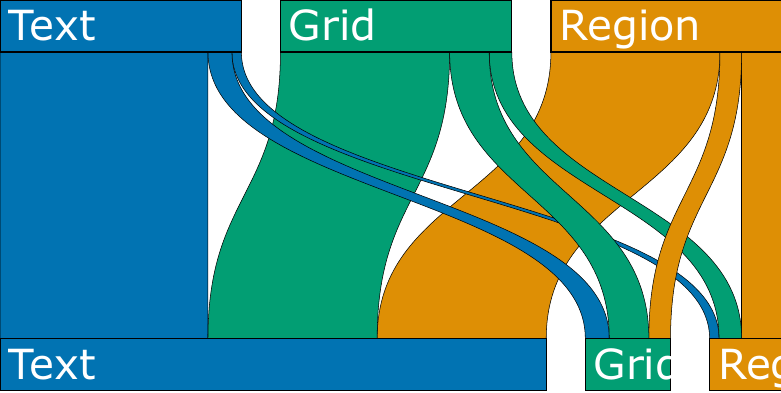}
        \includegraphics[width=.23\linewidth]{media/plots/cross_head/MSCOCO+Region,Patch_graph.pdf}
        \includegraphics[width=.23\linewidth]{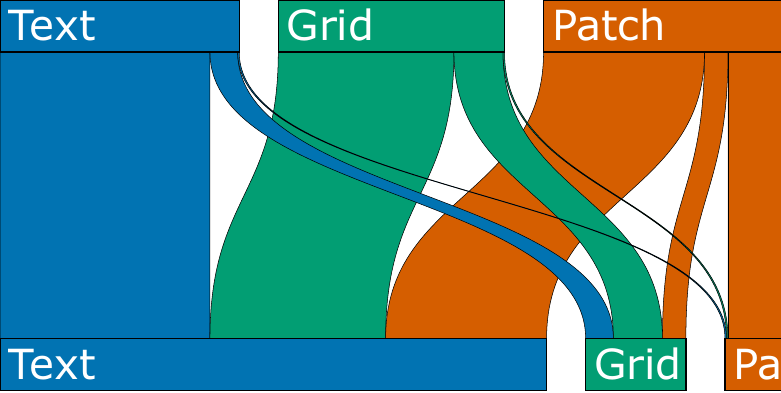}
        \caption{Cross-Attention for MSCOCO}
    \end{subfigure}
    \begin{subfigure}{.99\linewidth}
    \centering
        \includegraphics[width=.23\linewidth]{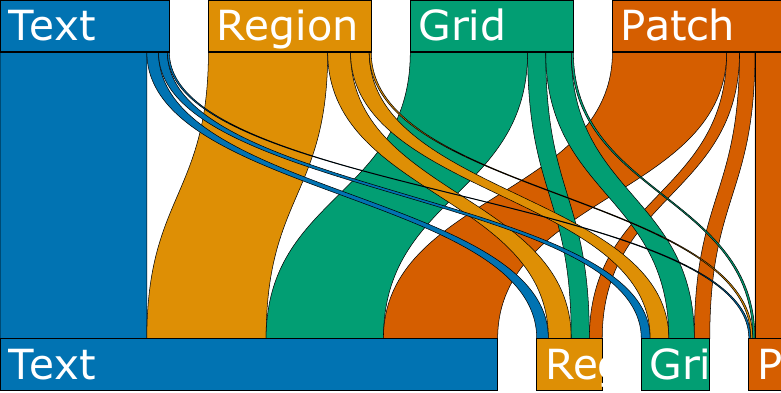}
        \includegraphics[width=.23\linewidth]{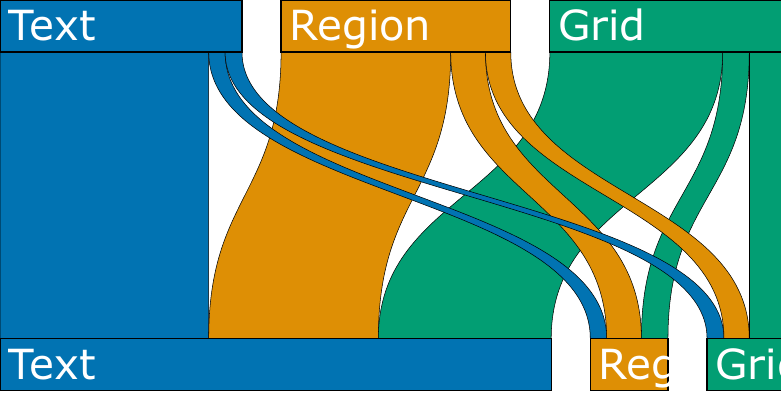}
        \includegraphics[width=.23\linewidth]{media/plots/cross_head/GQA+Region,Patch_graph.pdf}
        \includegraphics[width=.23\linewidth]{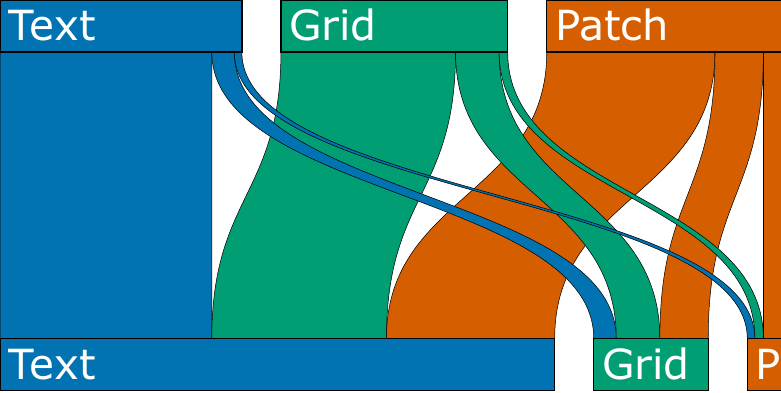}
        \caption{Cross-Attention for GQA}
    \end{subfigure}
    \begin{subfigure}{.99\linewidth}
    \centering
        \includegraphics[width=.23\linewidth]{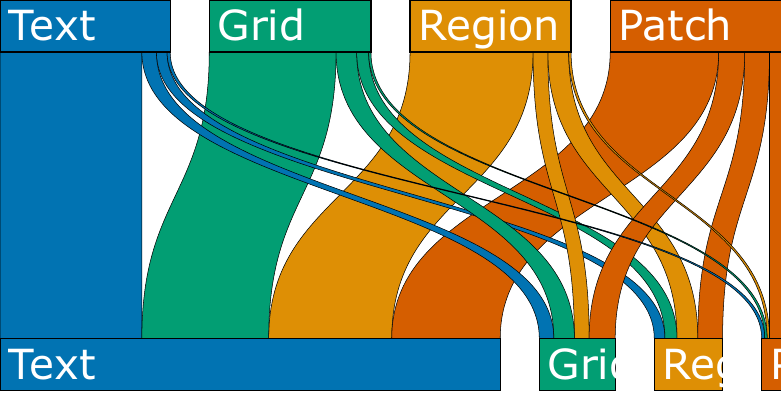}
        \includegraphics[width=.23\linewidth]{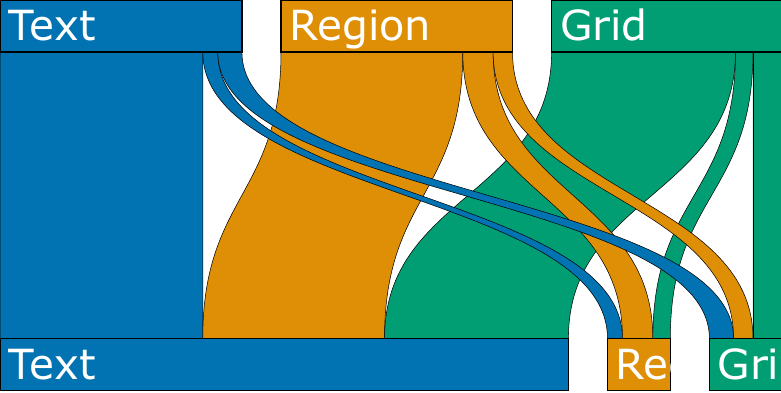}
        \includegraphics[width=.23\linewidth]{media/plots/cross_head/VQA+Region,Patch_graph.pdf}
        \includegraphics[width=.23\linewidth]{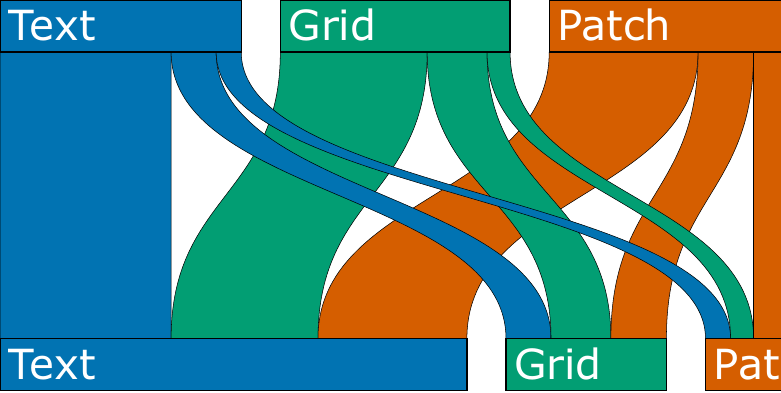}
        \caption{Cross-Attention for VQA}
    \end{subfigure}
    \begin{subfigure}{.99\linewidth}
    \centering
        \includegraphics[width=.23\linewidth]{media/plots/cross_head/SNLI-VE+Region,Grid,Patch_graph.pdf}
        \includegraphics[width=.23\linewidth]{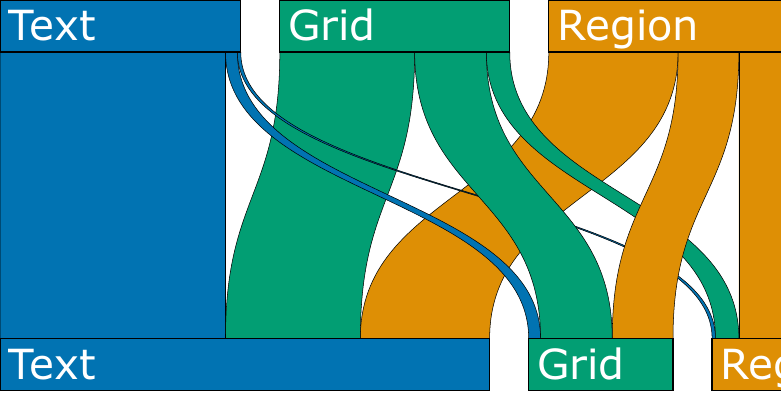}
        \includegraphics[width=.23\linewidth]{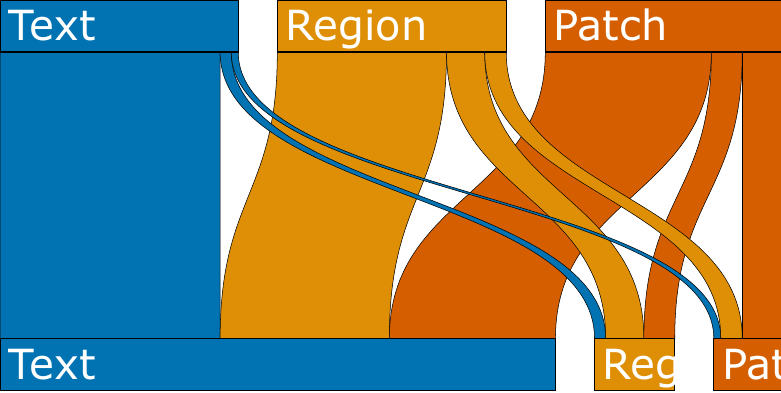}
        \includegraphics[width=.23\linewidth]{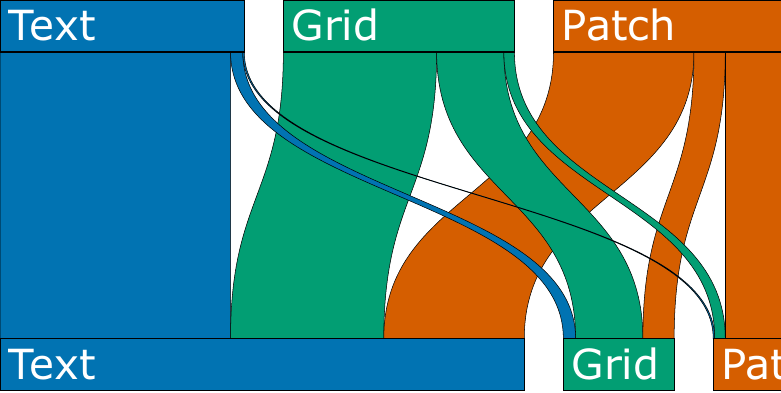}
        \caption{Cross-Attention for SNLI-VE}
    \end{subfigure}
    \begin{subfigure}{.99\linewidth}
    \centering
        \includegraphics[width=.23\linewidth]{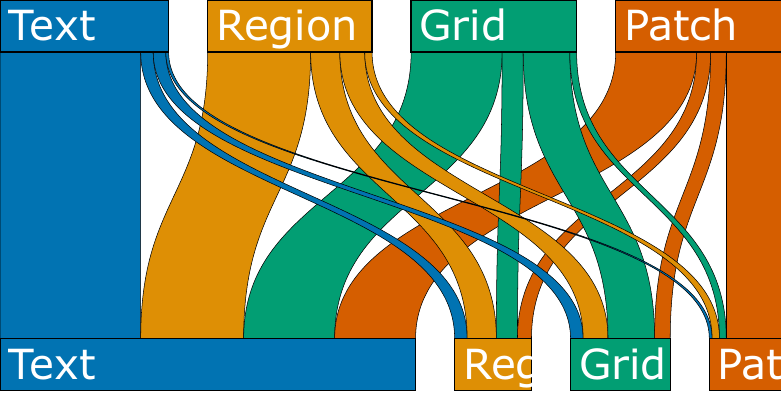}
        \includegraphics[width=.23\linewidth]{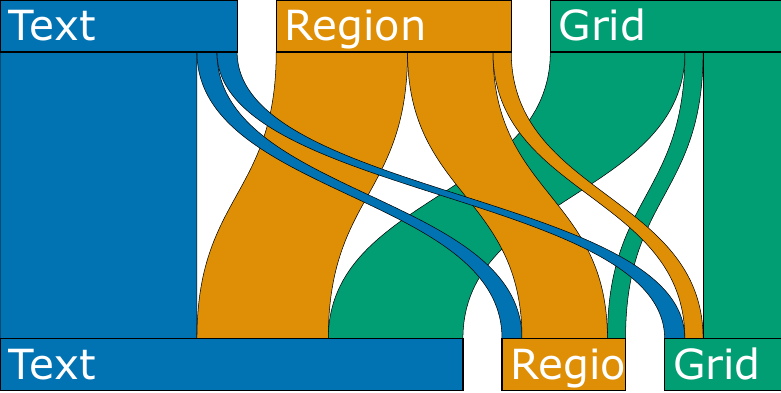}
        \includegraphics[width=.23\linewidth]{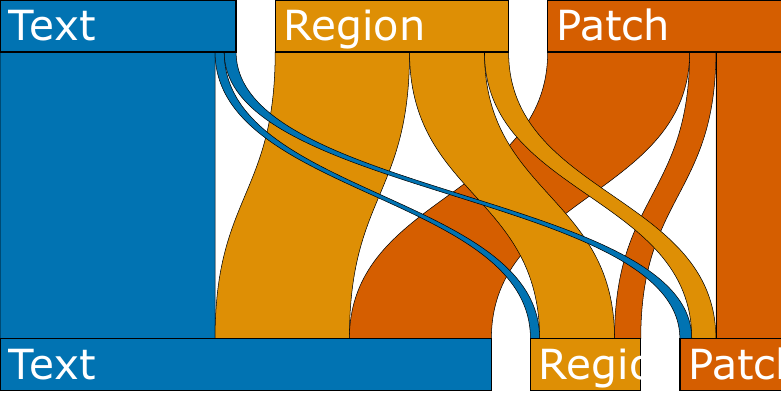}
        \includegraphics[width=.23\linewidth]{media/plots/cross_head/Hateful+Grid,Patch_graph.pdf}
        \caption{Cross-Attention for Hateful Memes}
    \end{subfigure}
    \caption{Attention flow (in \%) from each modality (top) to all modalities (bottom). Flow is the sum of all attention
weights between the modalities, averaged over all modality tokens and all attention heads.}
\label{fig:appendix:ana:cross}
\end{figure*}

\begin{figure*}[!ht]
\centering
    \begin{subfigure}{.99\linewidth}
    \centering
        \includegraphics[width=.23\linewidth]{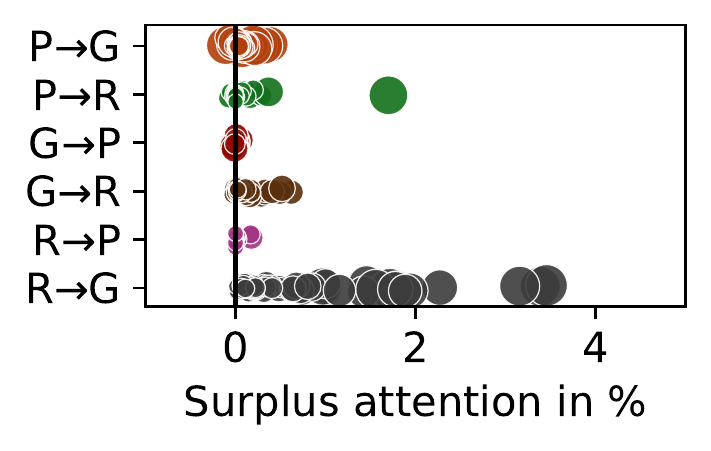}
        \includegraphics[width=.23\linewidth]{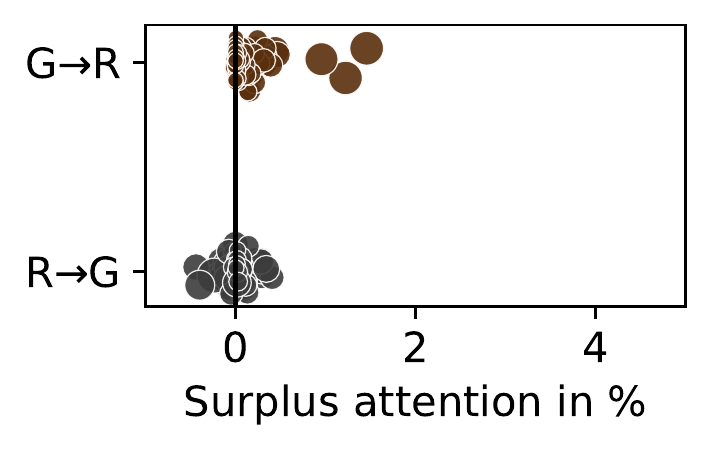}
        \includegraphics[width=.23\linewidth]{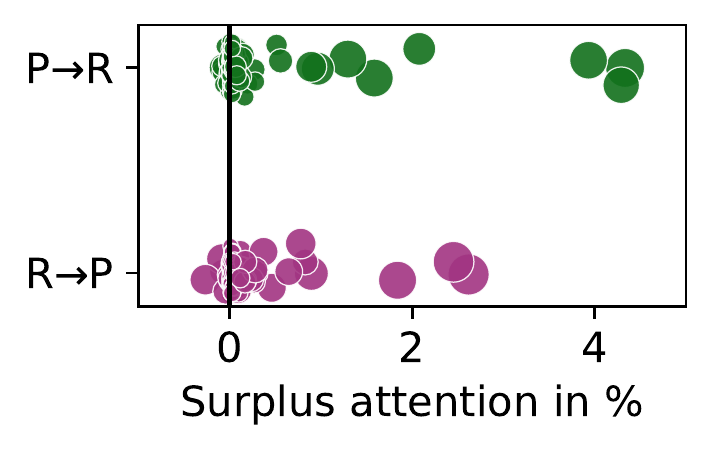}
        \includegraphics[width=.23\linewidth]{media/plots/cross_img/Flickr30k+Grid,Patch.pdf}
        \caption{Surplus Attention Flickr30k}
    \end{subfigure}
    \begin{subfigure}{.99\linewidth}
    \centering
        \includegraphics[width=.23\linewidth]{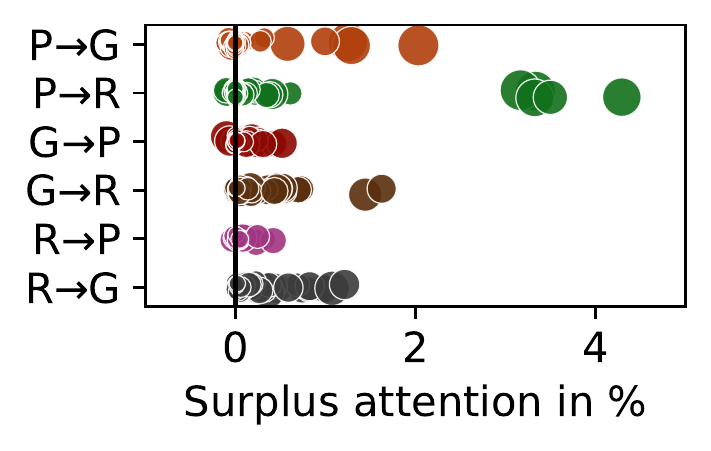}
        \includegraphics[width=.23\linewidth]{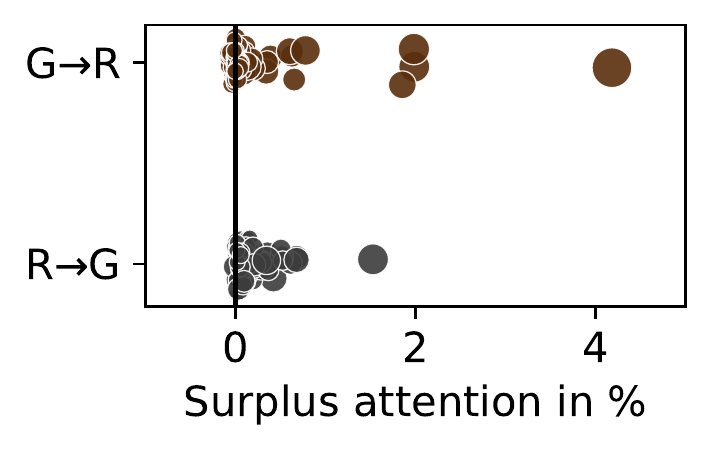}
        \includegraphics[width=.23\linewidth]{media/plots/cross_img/MSCOCO+Region,Patch.pdf}
        \includegraphics[width=.23\linewidth]{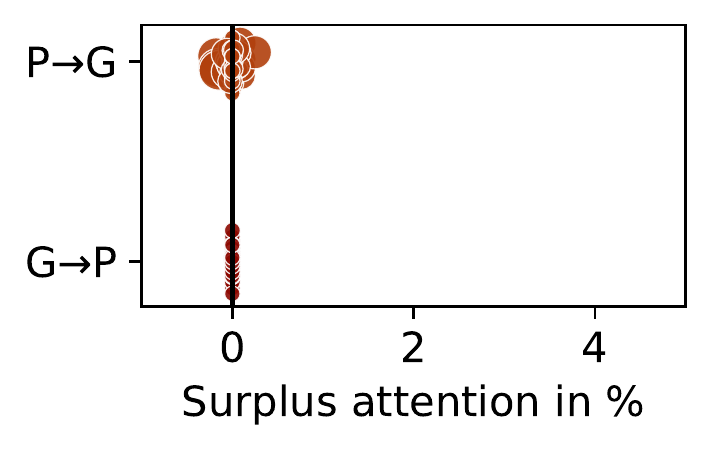}
        \caption{Surplus Attention MSCOCO}
    \end{subfigure}
    \begin{subfigure}{.99\linewidth}
    \centering
        \includegraphics[width=.23\linewidth]{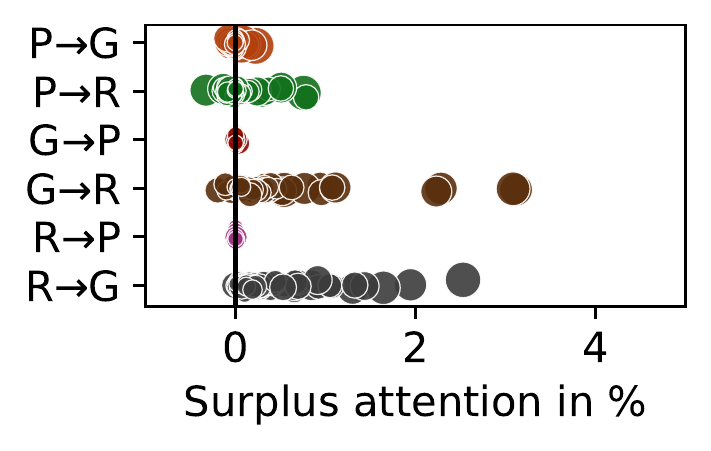}
        \includegraphics[width=.23\linewidth]{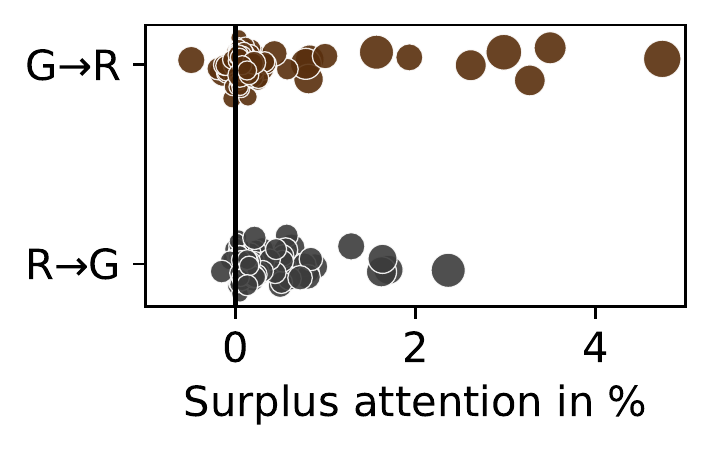}
        \includegraphics[width=.23\linewidth]{media/plots/cross_img/GQA+Region,Patch.pdf}
        \includegraphics[width=.23\linewidth]{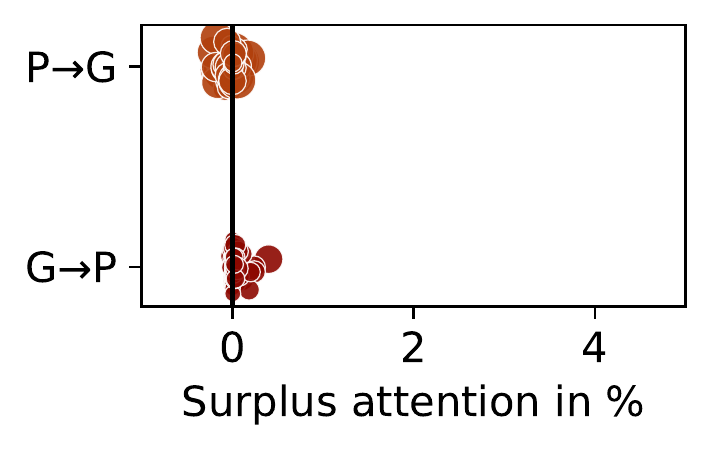}
        \caption{Surplus Attention GQA}
    \end{subfigure}
    \begin{subfigure}{.99\linewidth}
    \centering
        \includegraphics[width=.23\linewidth]{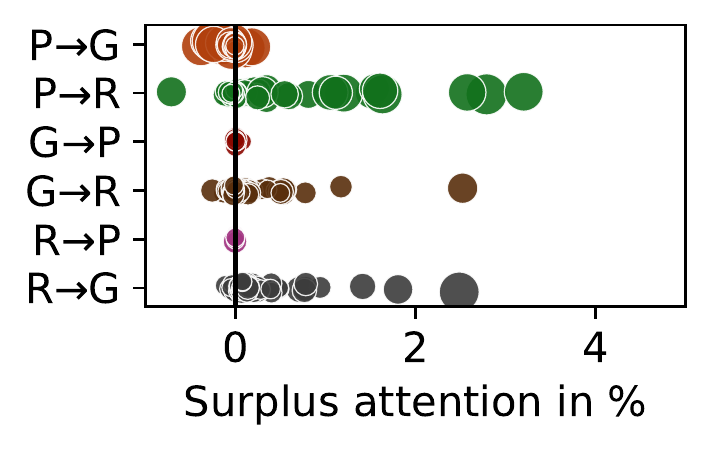}
        \includegraphics[width=.23\linewidth]{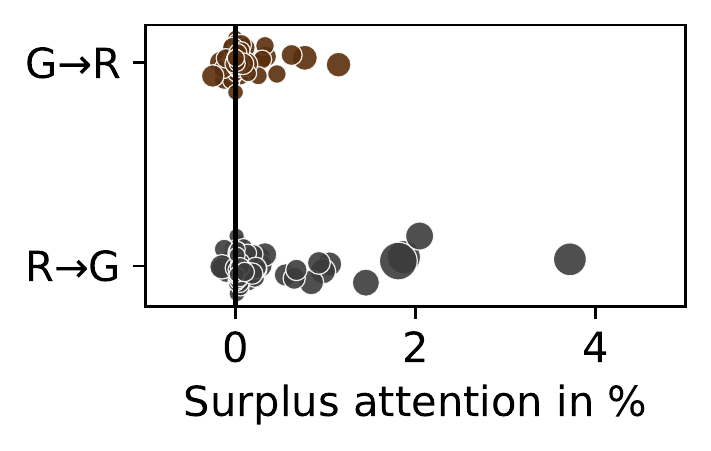}
        \includegraphics[width=.23\linewidth]{media/plots/cross_img/VQA+Region,Patch.pdf}
        \includegraphics[width=.23\linewidth]{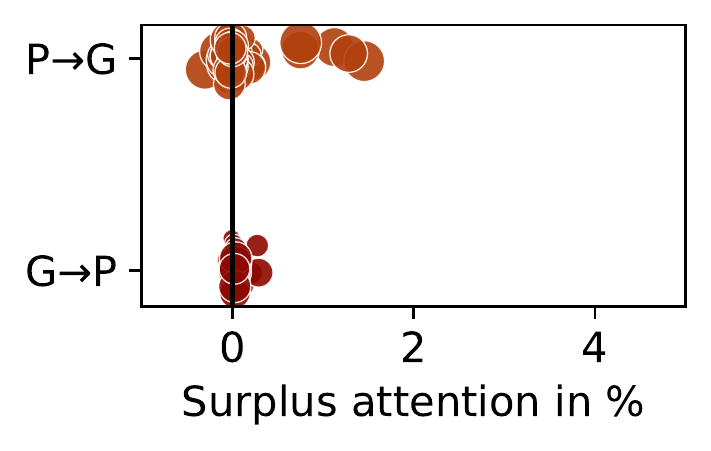}
        \caption{Surplus Attention VQA}
    \end{subfigure}
    \begin{subfigure}{.99\linewidth}
    \centering
        \includegraphics[width=.23\linewidth]{media/plots/cross_img/SNLI-VE+Region,Grid,Patch.pdf}
        \includegraphics[width=.23\linewidth]{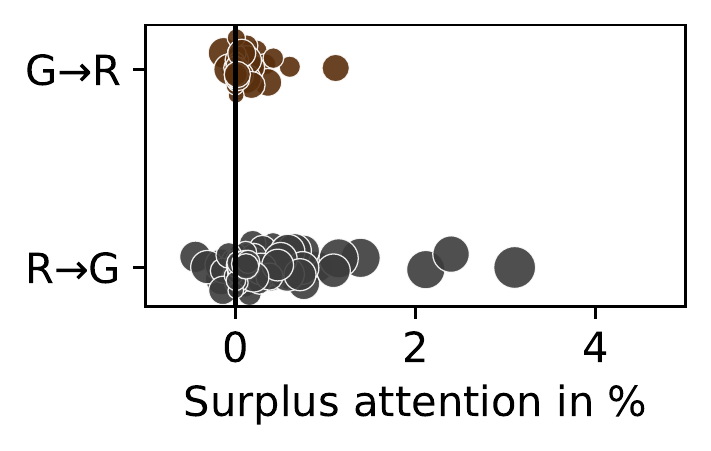}
        \includegraphics[width=.23\linewidth]{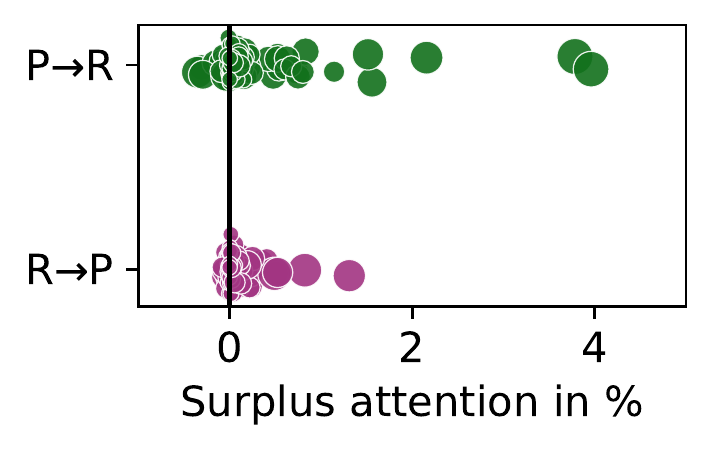}
        \includegraphics[width=.23\linewidth]{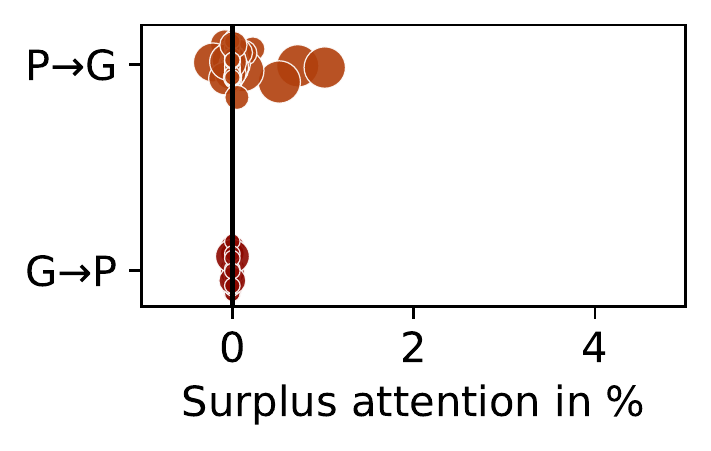}
        \caption{Surplus Attention SNLI-VE}
    \end{subfigure}
    \begin{subfigure}{.99\linewidth}
    \centering
        \includegraphics[width=.23\linewidth]{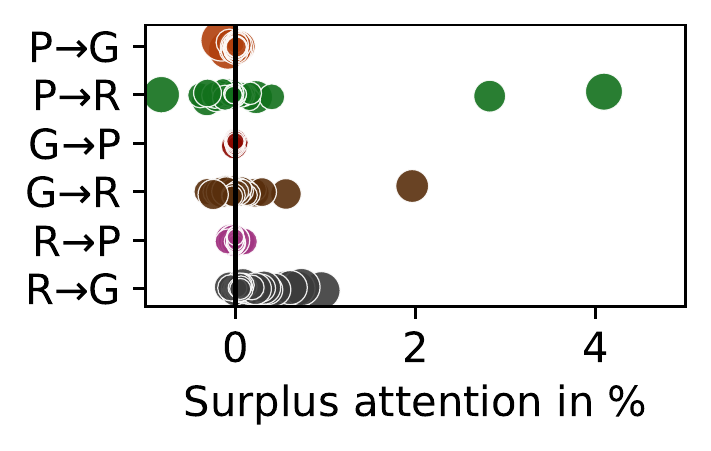}
        \includegraphics[width=.23\linewidth]{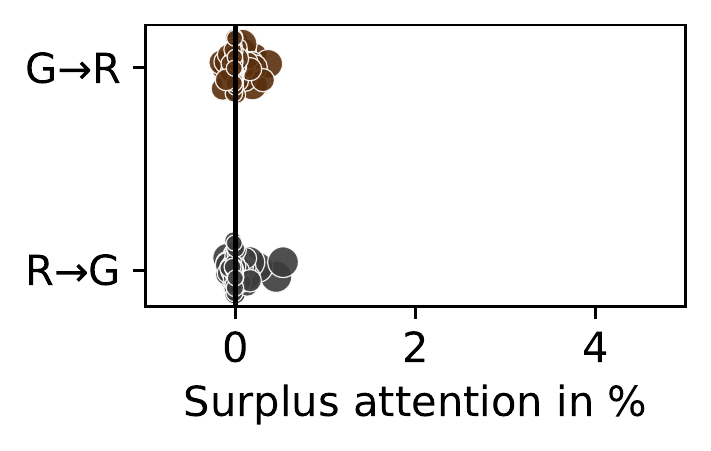}
        \includegraphics[width=.23\linewidth]{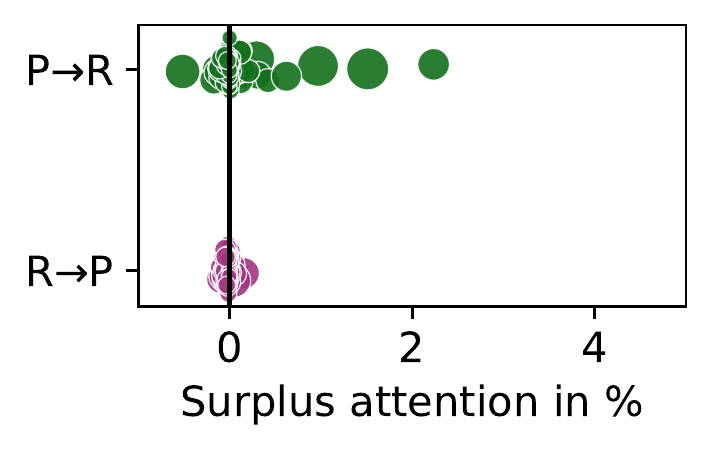}
        \includegraphics[width=.23\linewidth]{media/plots/cross_img/Hateful+Grid,Patch.pdf}
        \caption{Surplus Attention Hateful Memes}
    \end{subfigure}
    \caption{Surplus attention of attention heads from one VE's tokens to another target VE's overlapping tokens compared to the other non-overlapping tokens of the target VE. \textbf{Dot size} represents the average total attention paid to the target VE by each head.  (Abbreviations: \textbf{R}egion, \textbf{G}rid, \textbf{P}atch).}
\label{fig:appendix:ana:crossimg}
\end{figure*}

\begin{figure*}[ht!]
\centering
    \begin{subfigure}{.99\linewidth}
    \centering
        \includegraphics[width=.23\linewidth]{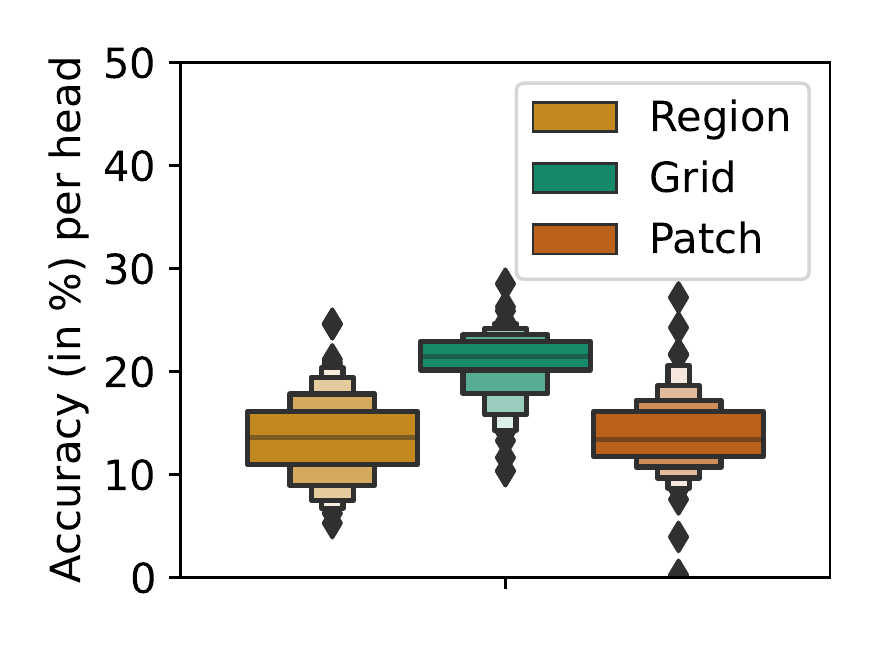}
        \includegraphics[width=.23\linewidth]{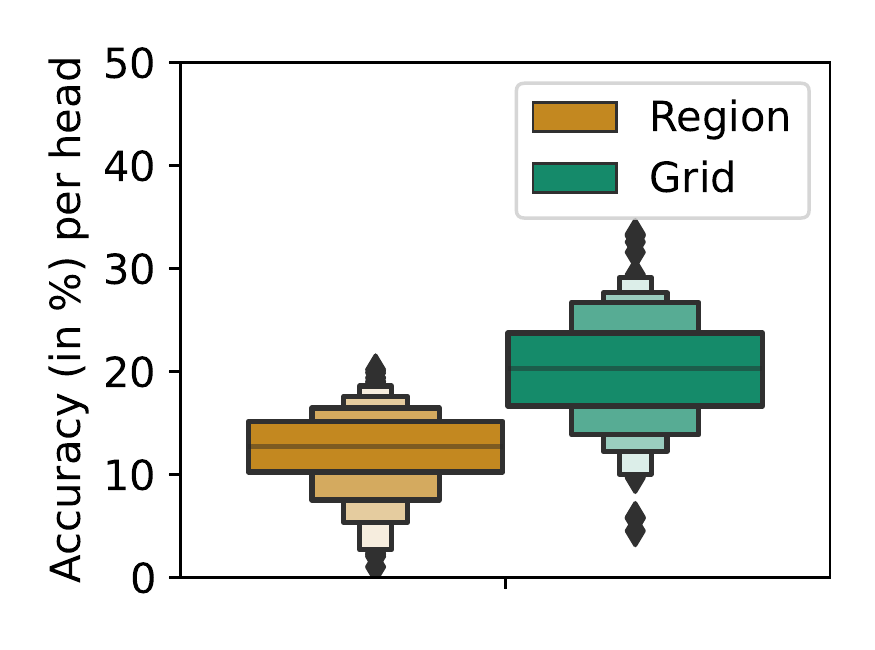} 
        \includegraphics[width=.23\linewidth]{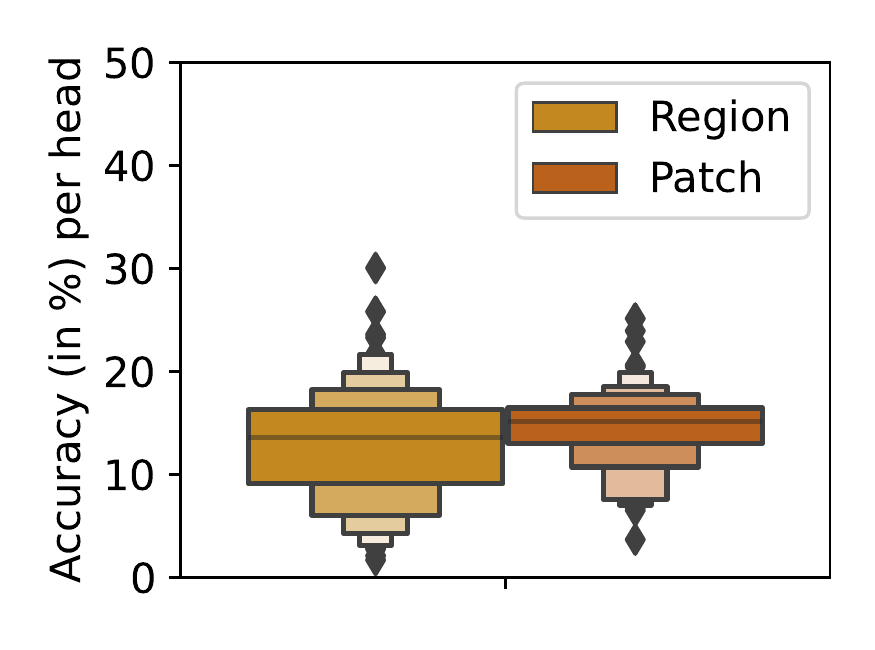}
        \includegraphics[width=.23\linewidth]{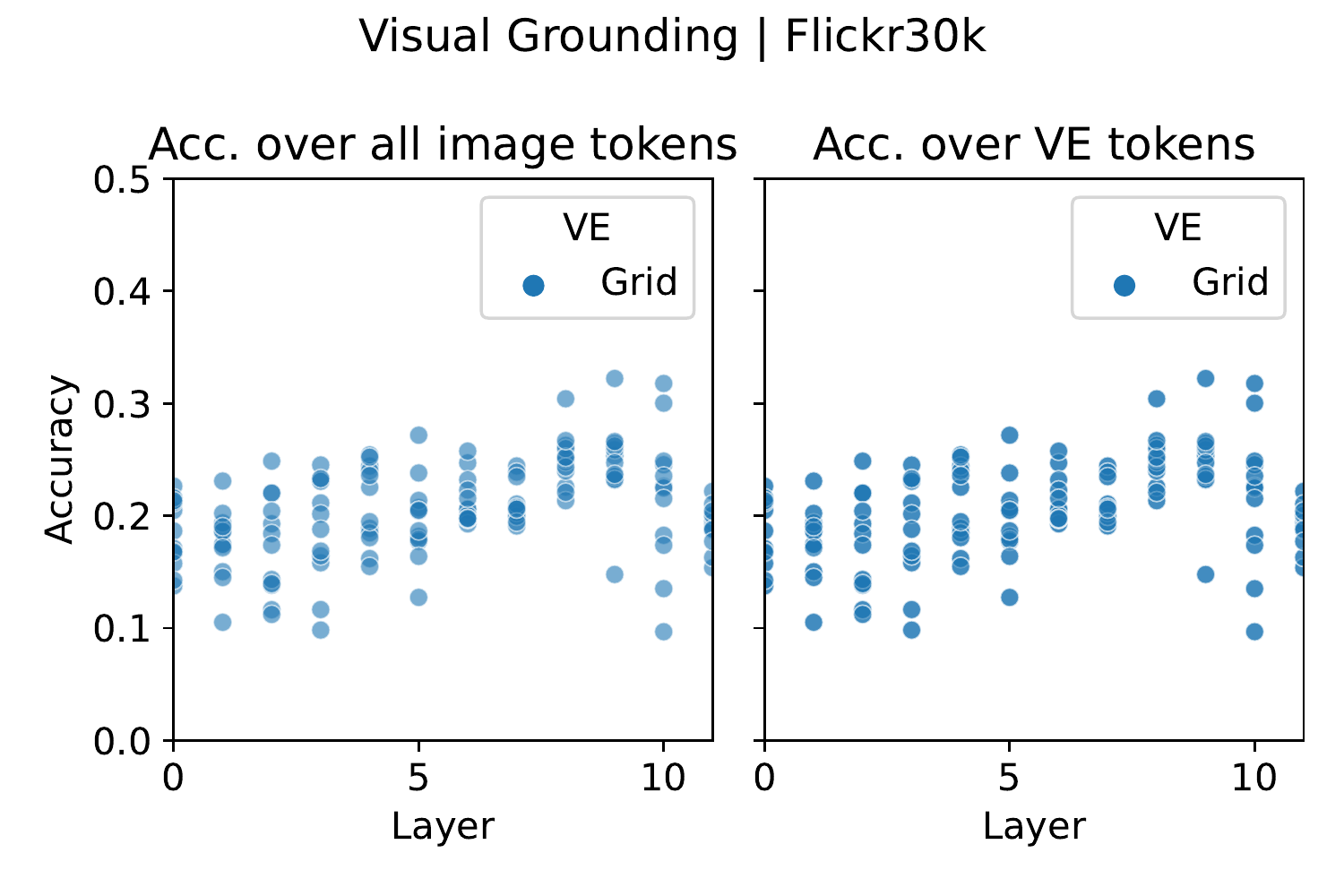,Patch_per-ve.pdf}
        \caption{Visual Grounding of attention heads on  Flickr30k}
    \end{subfigure}
    \begin{subfigure}{.99\linewidth}
    \centering
        \includegraphics[width=.23\linewidth]{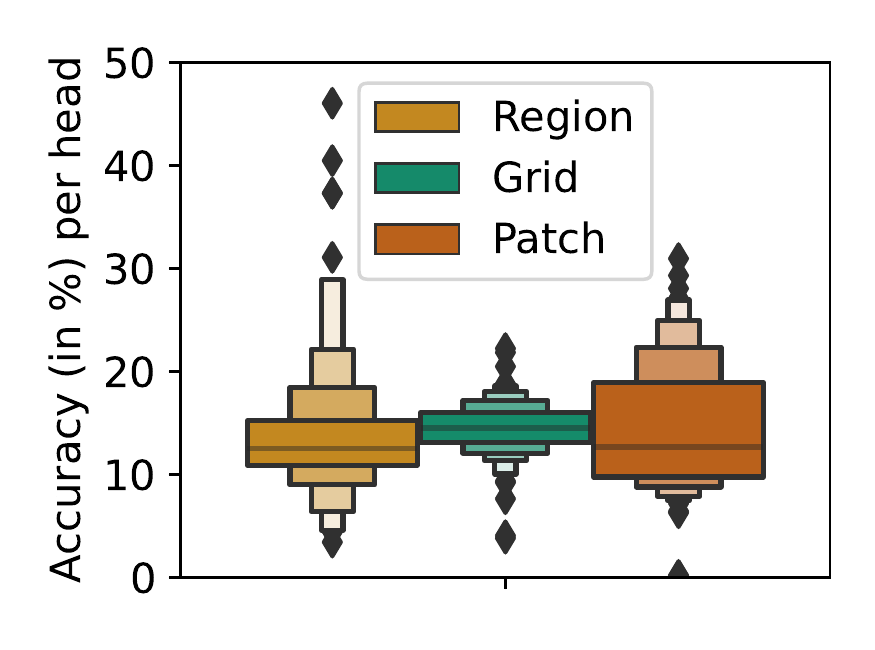}
        \includegraphics[width=.23\linewidth]{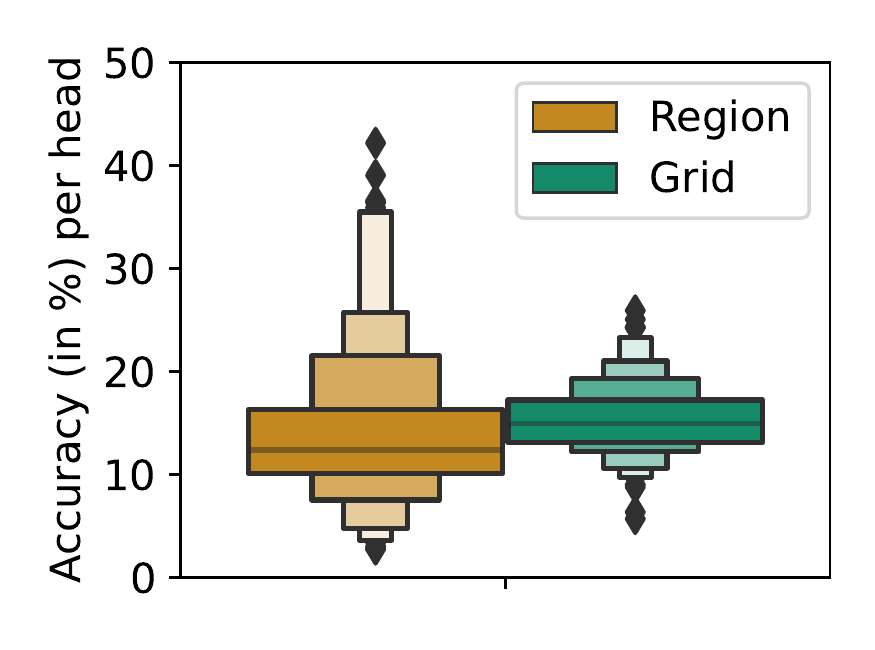}
        \includegraphics[width=.23\linewidth]{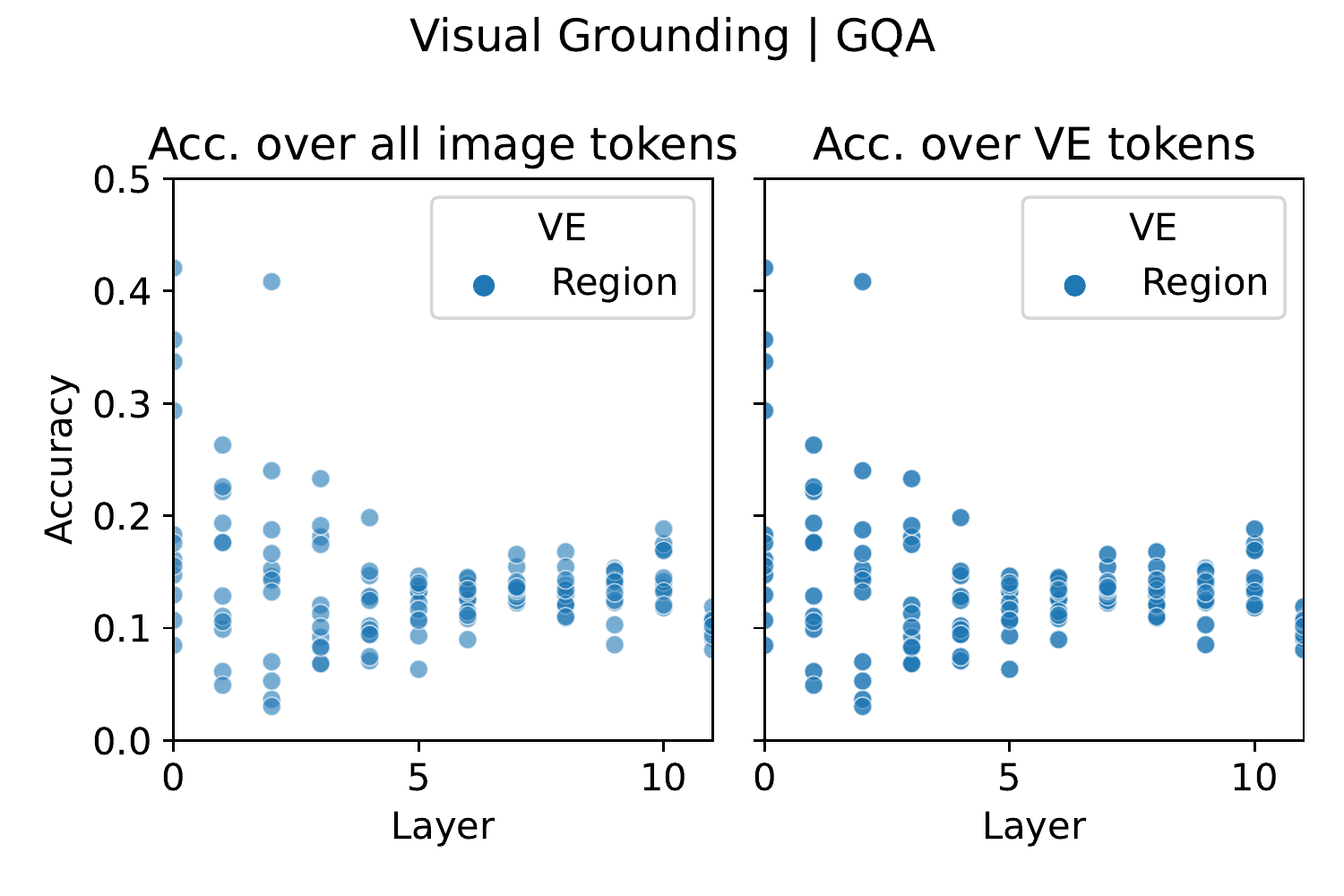,Patch_per-ve.pdf}
        \includegraphics[width=.23\linewidth]{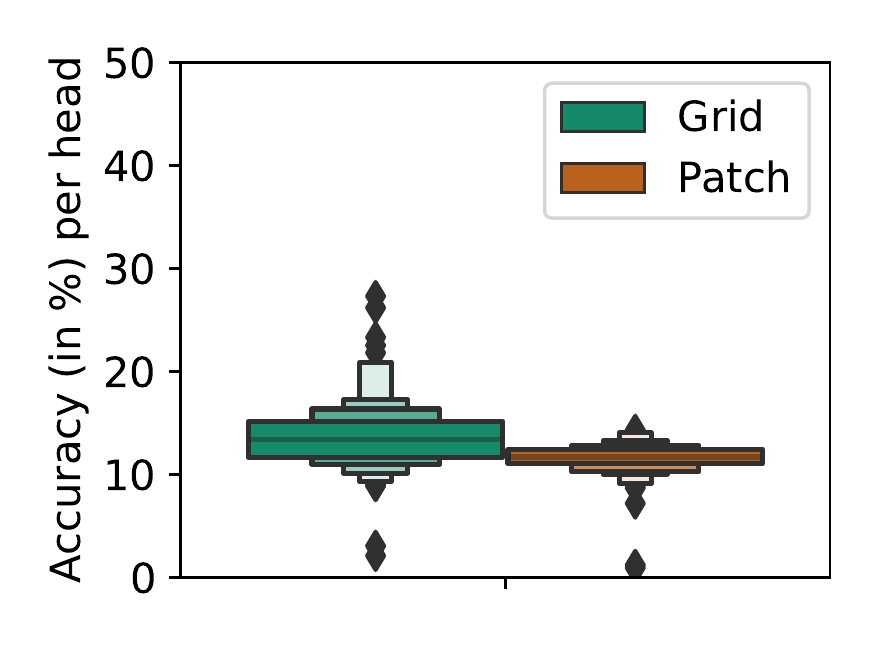}
        \caption{Visual Grounding of attention heads on  GQA}
    \end{subfigure}
    \caption{Visual Entity Grounding accuracy of all attention heads. An entity is grounded correctly to a VE if the attention weight from the phrase to the matching visual tokens is the highest over all the VE's tokens.}
\label{fig:appendix:ana:grounding}
\end{figure*}

We present the full results for all VE combinations from the analysis of \S\ref{sec:ana}.
Figure \ref{fig:appendix:ana:clsattention} shows the CLS attention, Figure \ref{fig:appendix:ana:cross} the attention flow, Figure \ref{fig:appendix:ana:crossimg} the surplus attention for overlapping tokens, and Figure \ref{fig:appendix:ana:grounding} the visual grounding.

\section{Analysis Results after VE-Dropout Training}
\label{sec:appendix:ana:drop}
\begin{figure*}[ht!]
    \centering
    \begin{subfigure}{.3\linewidth}
    \centering
        \includegraphics[width=.99\linewidth]{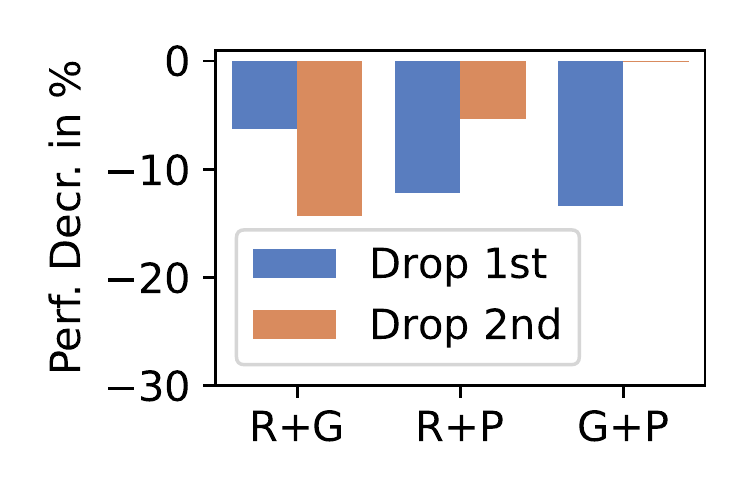}  
        \caption{Flickr30k Pruning}
        \label{fig:ana:grounding:flickr}
    \end{subfigure}
    \begin{subfigure}{.3\linewidth}
    \centering
        \includegraphics[width=.99\linewidth]{media/plots/pruning/MSCOCO_drop.pdf}  
        \caption{MSCOCO Pruning}
        \label{fig:ana:grounding:mscoco}
    \end{subfigure} 
    \begin{subfigure}{.3\linewidth}
    \centering
        \includegraphics[width=.99\linewidth]{media/plots/pruning/GQA_drop.pdf}  
        \caption{GQA Pruning}
        \label{fig:ana:grounding:gqa}
    \end{subfigure} \\
    \begin{subfigure}{.3\linewidth}
    \centering
        \includegraphics[width=.99\linewidth]{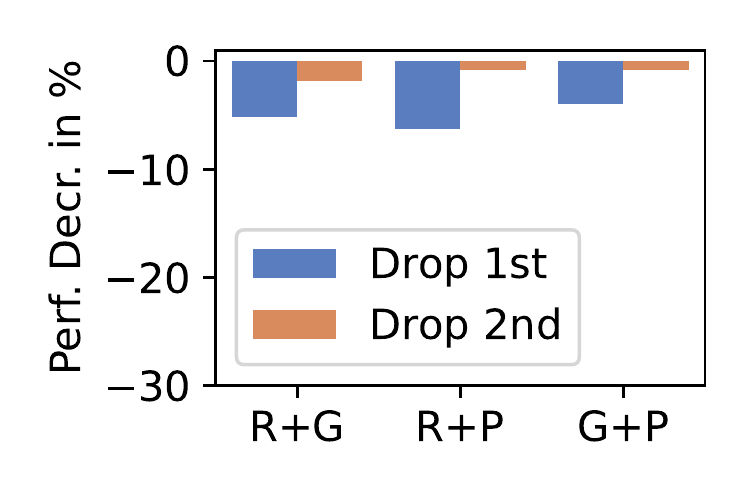}  
        \caption{VQA Pruning} 
        \label{fig:ana:grounding:vqa}
    \end{subfigure} 
    \begin{subfigure}{.3\linewidth}
    \centering
        \includegraphics[width=.99\linewidth]{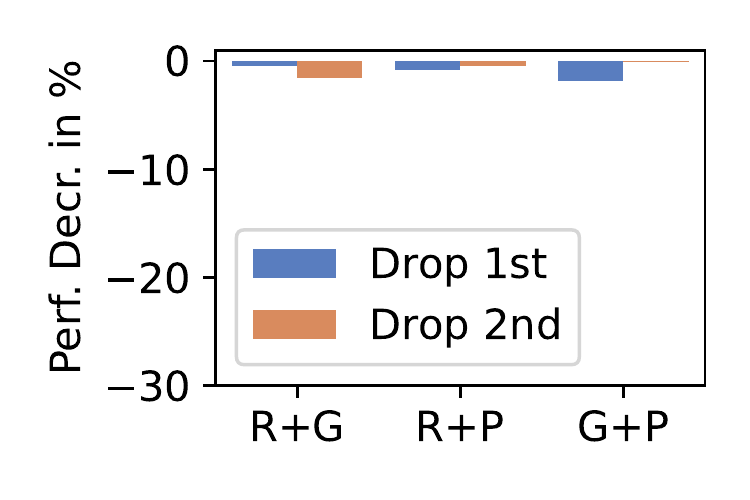}  
        \caption{SNLI-VE Pruning}
        \label{fig:ana:grounding:snlive}
    \end{subfigure}
    \begin{subfigure}{.3\linewidth}
    \centering
        \includegraphics[width=.99\linewidth]{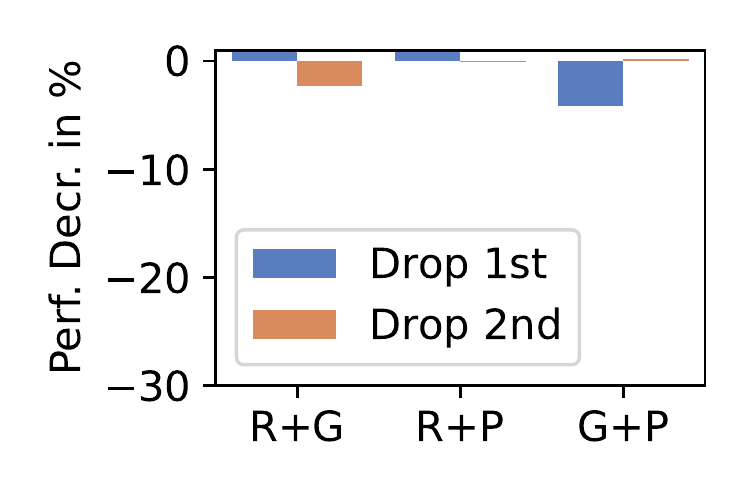}  
        \caption{Hateful Memes Pruning}
        \label{fig:ana:grounding:snlive}
    \end{subfigure}
    \caption{(After VE-Dropout Training) Relative performance decrease of 2-encoder models after dropping the entire first or second VE from the input compared to evaluation with both encoders in use.
    (Abbreviations: \textbf{R}egion, \textbf{G}rid, \textbf{P}atch).}
\label{fig:appendix:ana:pruning}
\end{figure*}

\begin{figure*}[]
\centering
    \begin{subfigure}{.99\linewidth}
    \centering
        \includegraphics[width=.3\linewidth]{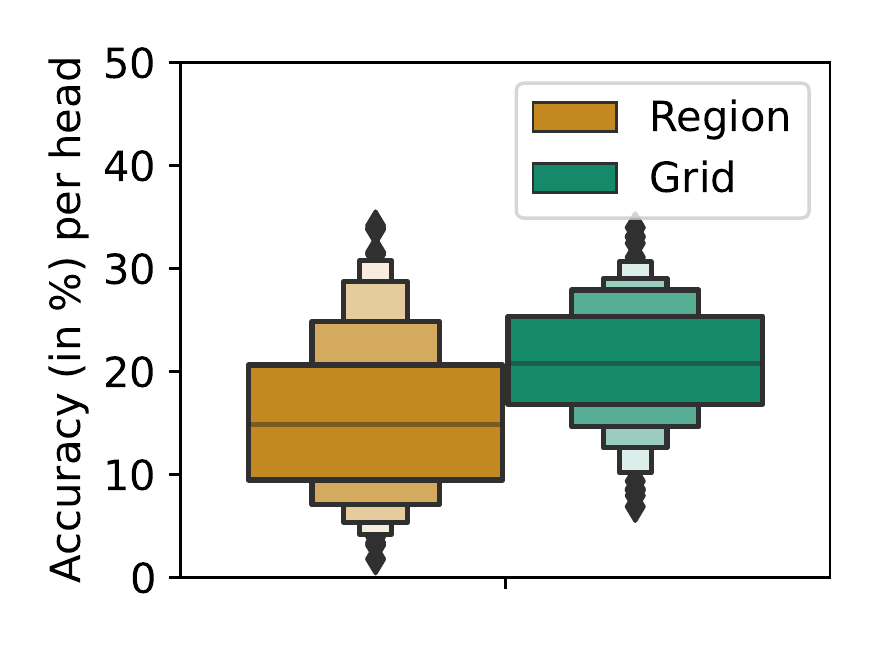} 
        \includegraphics[width=.3\linewidth]{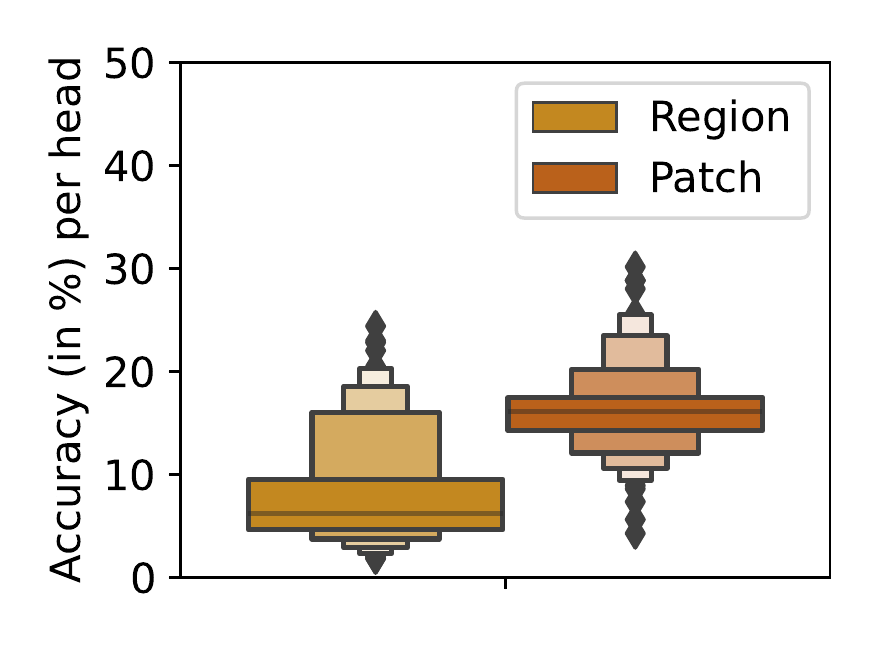}
        \includegraphics[width=.3\linewidth]{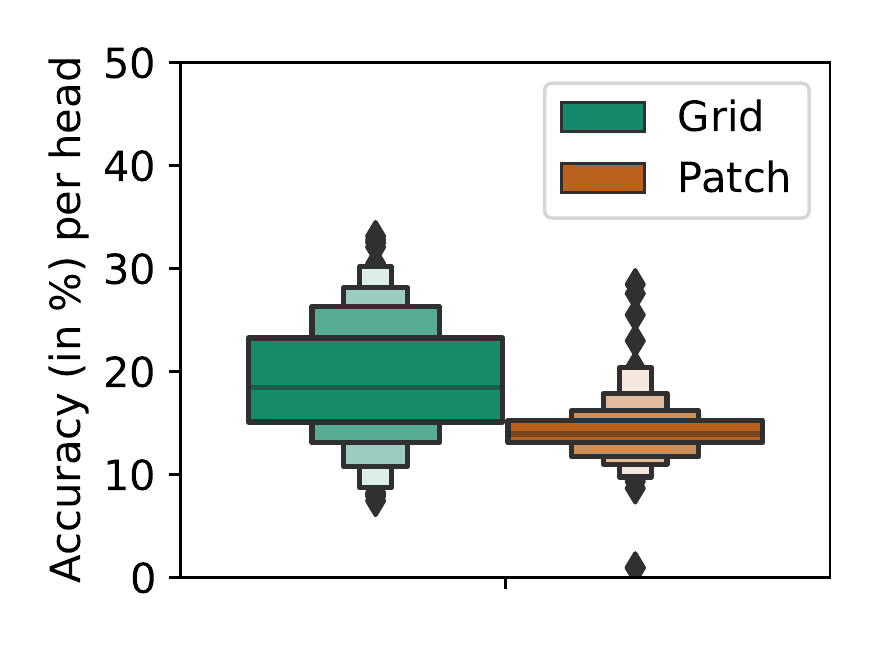}
        \caption{Visual Grounding of attention heads on  Flickr30k}
    \end{subfigure}
    \begin{subfigure}{.99\linewidth}
    \centering
        \includegraphics[width=.3\linewidth]{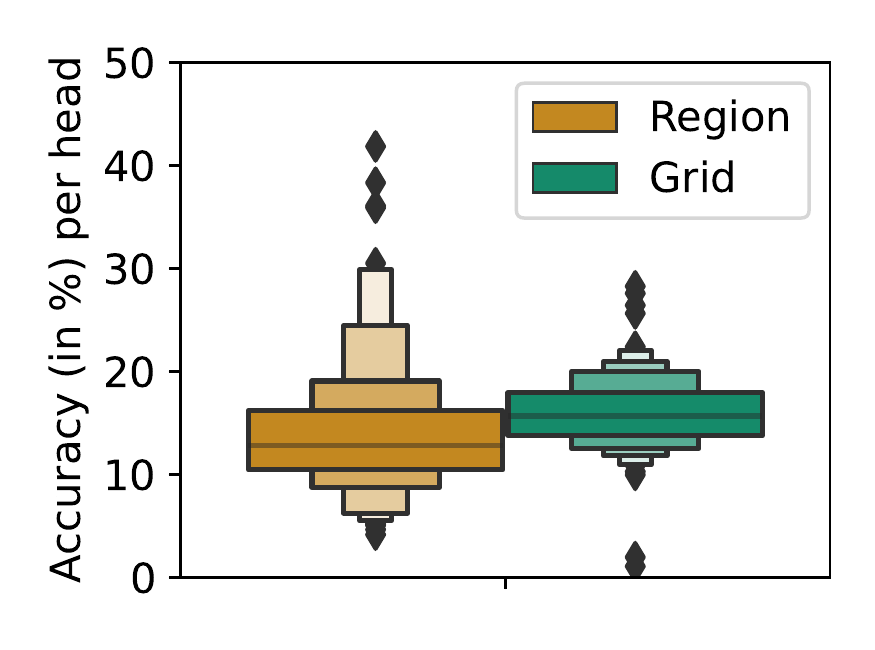}
        \includegraphics[width=.3\linewidth]{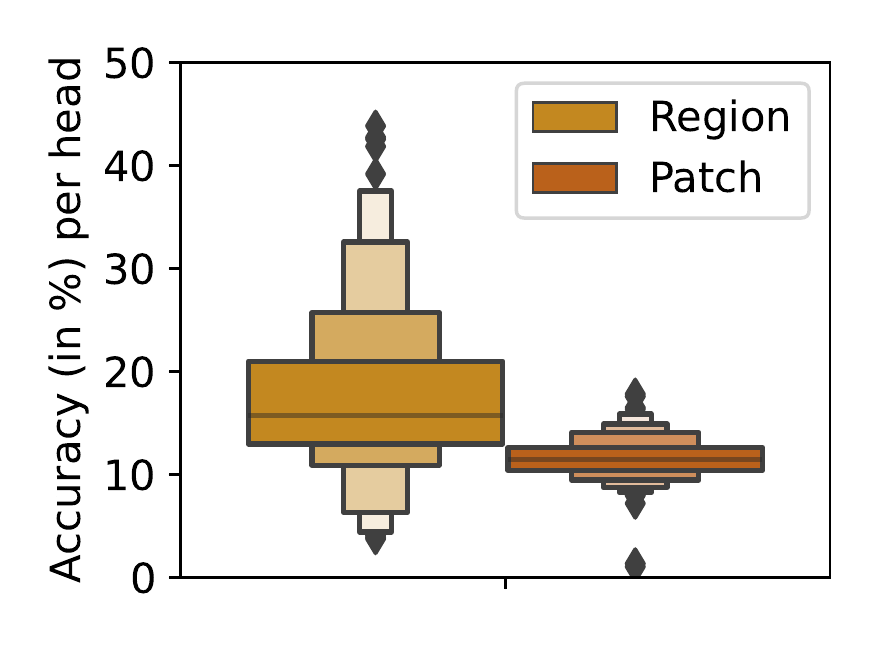}
        \includegraphics[width=.3\linewidth]{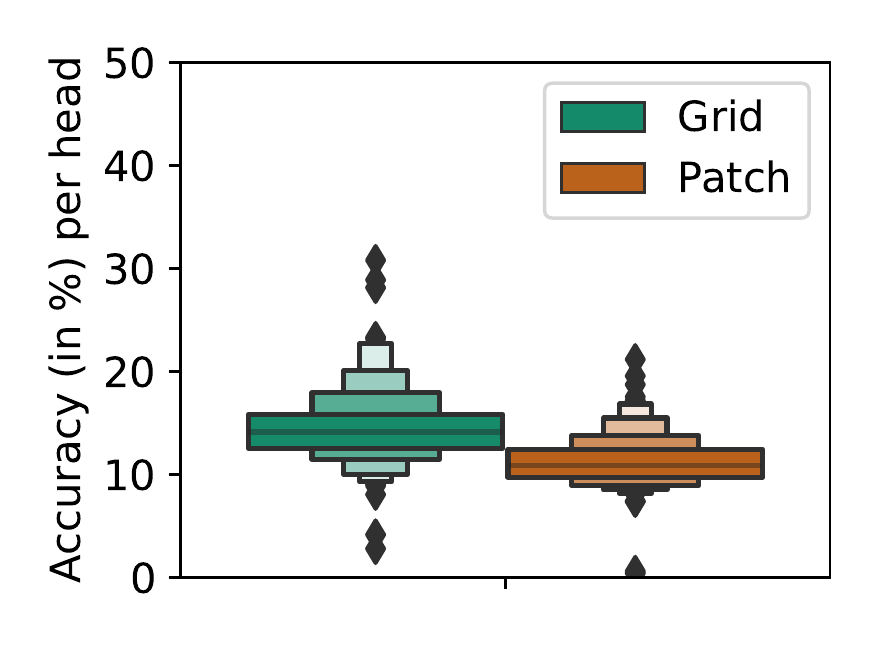}
        \caption{Visual Grounding of attention heads on  GQA}
    \end{subfigure}
    \caption{(After VE-Dropout Training) Visual Entity Grounding accuracy of all attention heads. An entity is grounded correctly to a VE if the attention weight from the phrase to the matching visual tokens is the highest over all the VE's tokens.}
\label{fig:appendix:ana:grounding:drop}
\end{figure*}

\begin{figure*}[!ht]
\centering
    \begin{subfigure}{.99\linewidth}
    \centering
        \includegraphics[width=.23\linewidth]{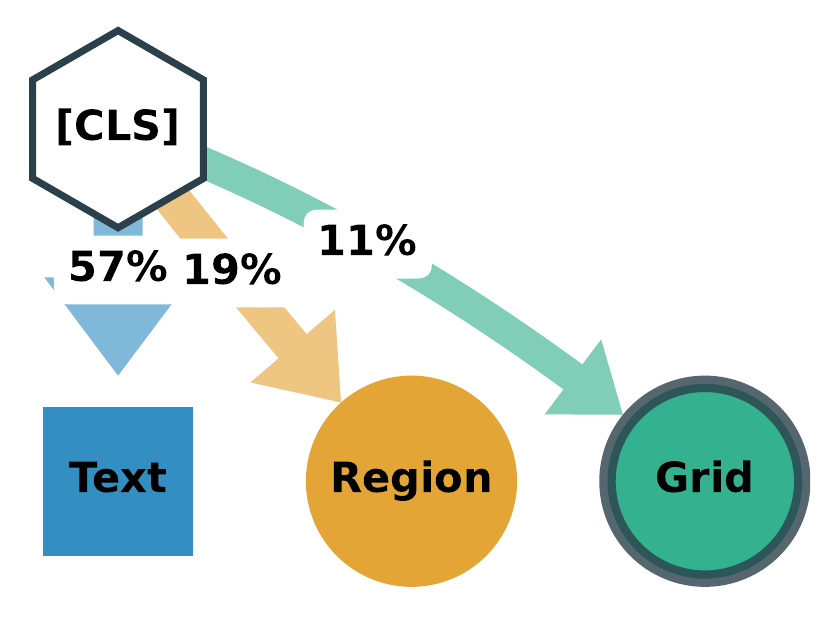}
        \includegraphics[width=.23\linewidth]{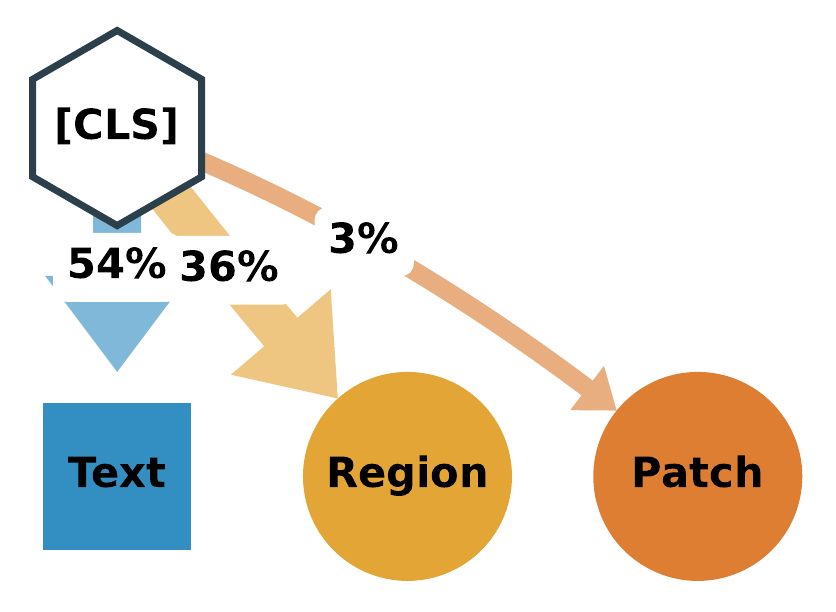}
        \includegraphics[width=.23\linewidth]{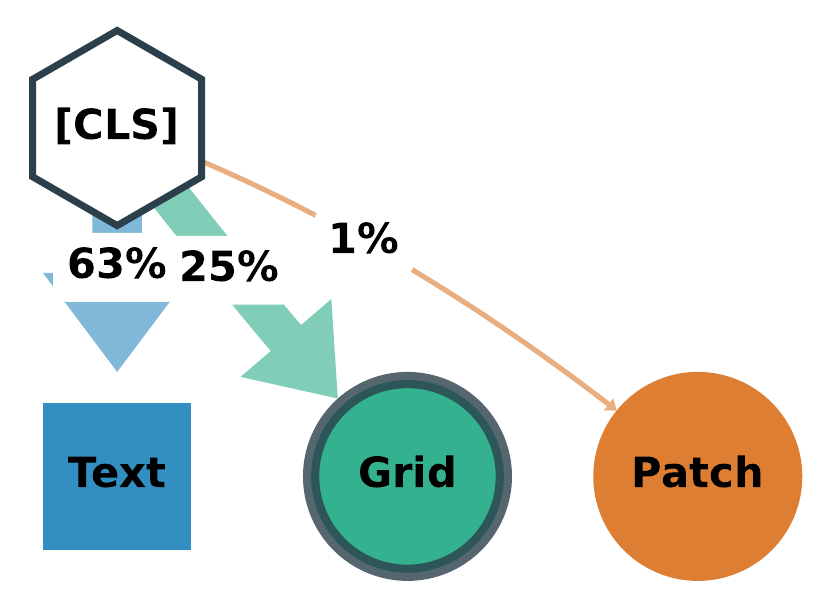}
        \caption{CLS Attention Flickr30k}
    \end{subfigure}
    \begin{subfigure}{.99\linewidth}
    \centering
        \includegraphics[width=.23\linewidth]{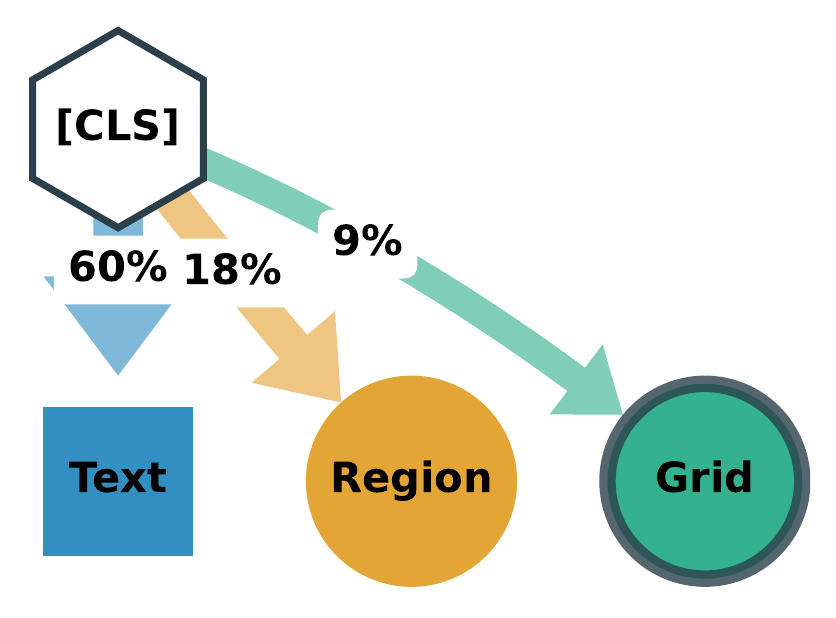}
        \includegraphics[width=.23\linewidth]{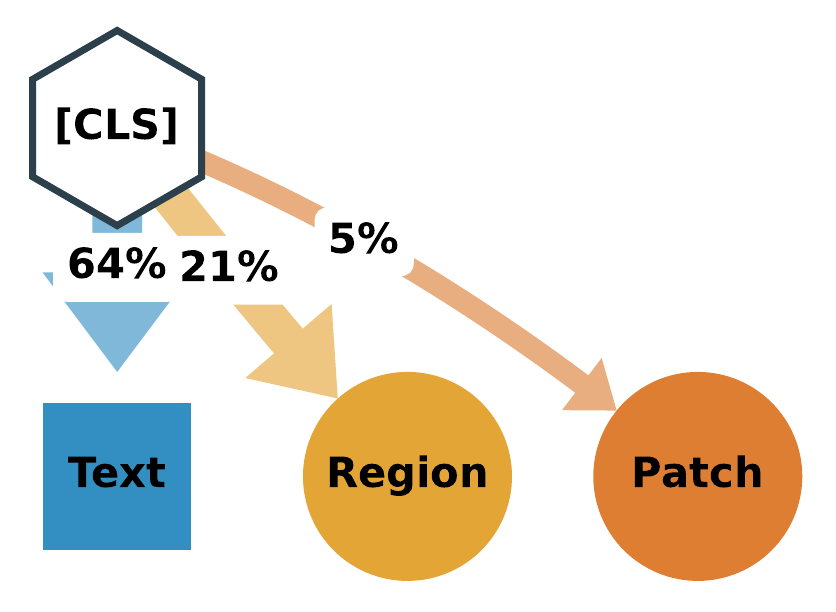}
        \includegraphics[width=.23\linewidth]{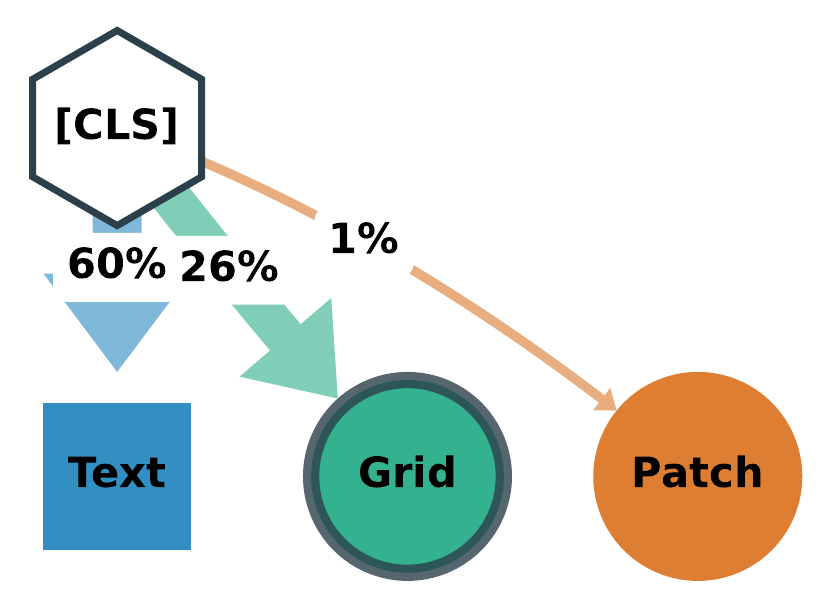}
        \caption{CLS Attention MSCOCO}
    \end{subfigure}
    \begin{subfigure}{.99\linewidth}
    \centering
        \includegraphics[width=.23\linewidth]{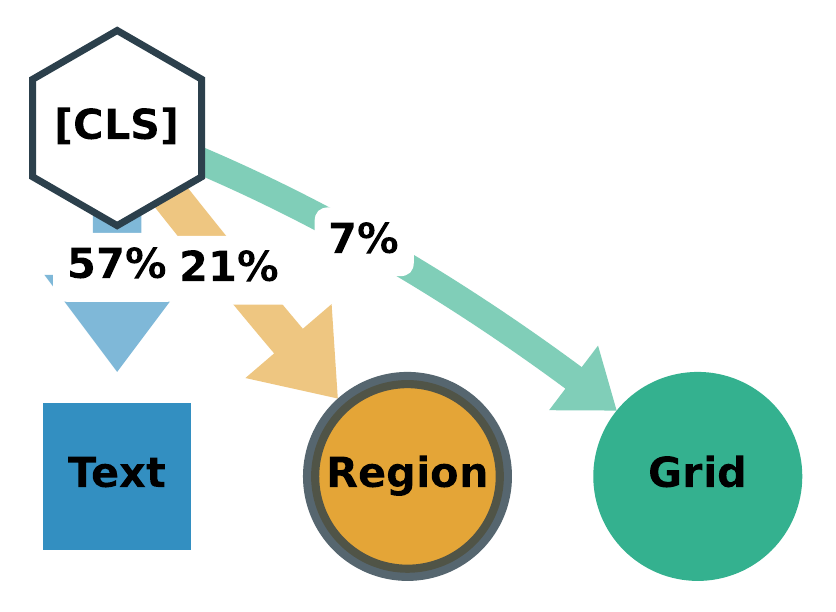}
        \includegraphics[width=.23\linewidth]{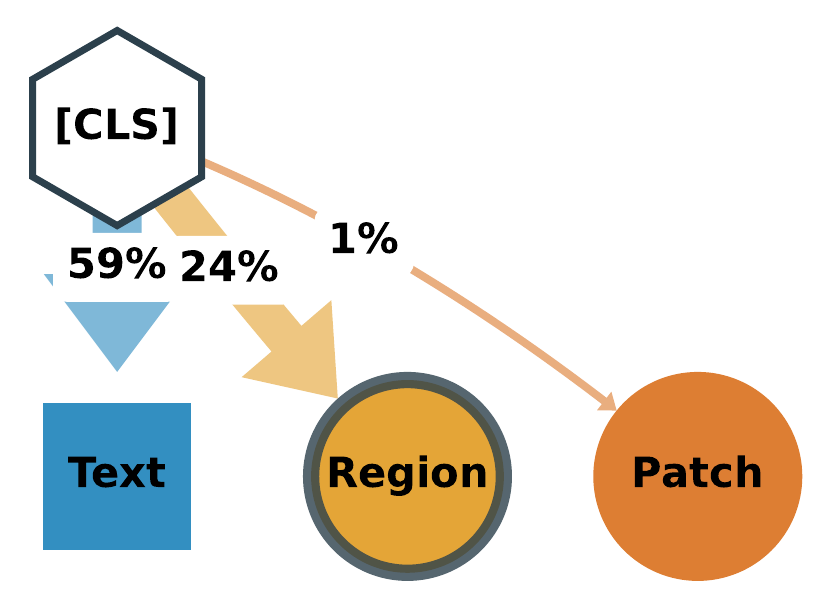}
        \includegraphics[width=.23\linewidth]{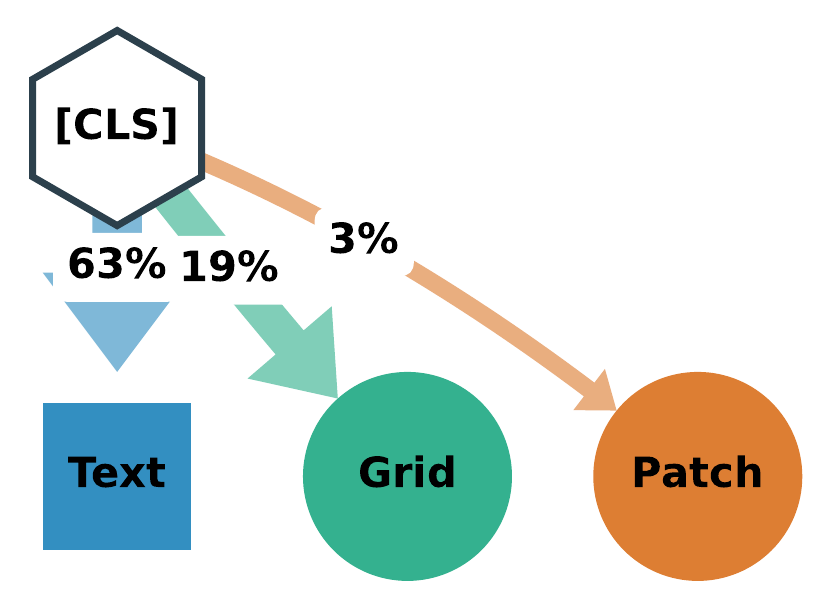}
        \caption{CLS Attention GQA}
    \end{subfigure}
    \begin{subfigure}{.99\linewidth}
    \centering
        \includegraphics[width=.23\linewidth]{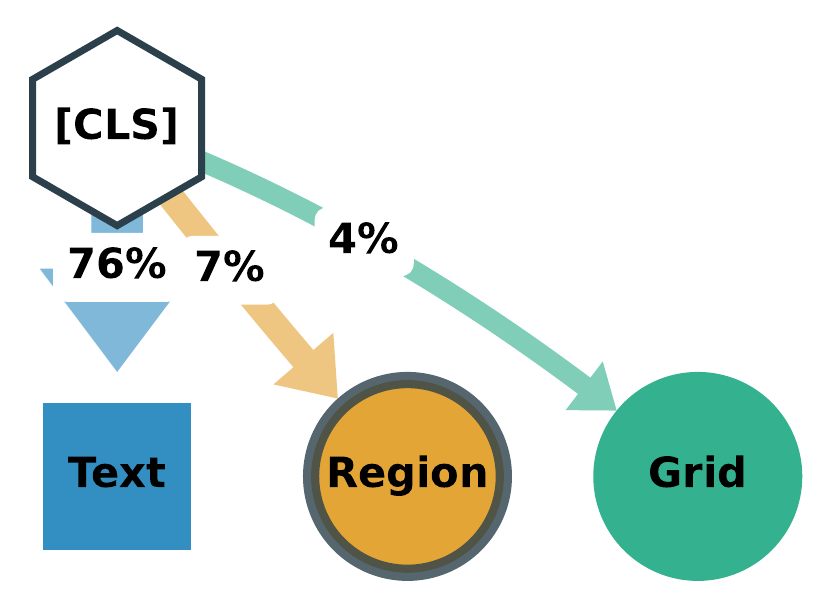}
        \includegraphics[width=.23\linewidth]{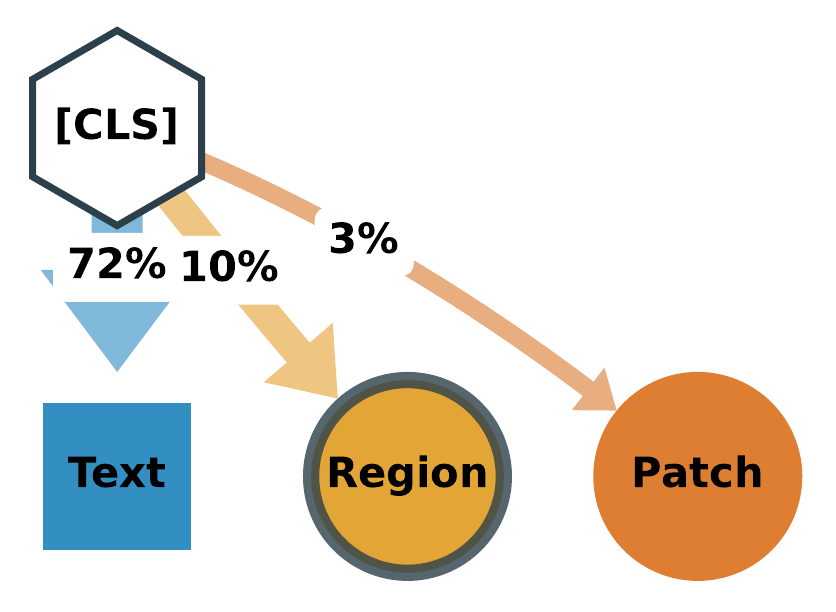}
        \includegraphics[width=.23\linewidth]{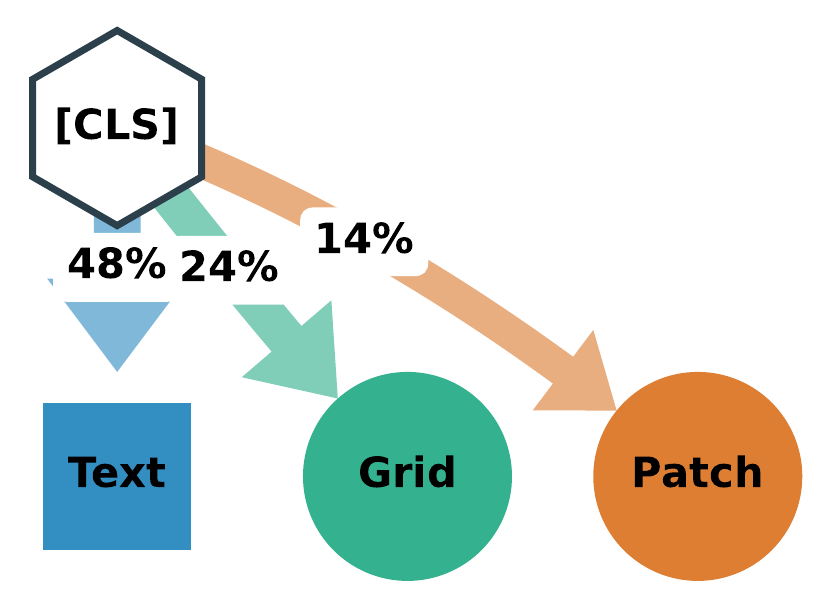}
        \caption{CLS Attention VQA}
    \end{subfigure}
    \begin{subfigure}{.99\linewidth}
    \centering
        \includegraphics[width=.23\linewidth]{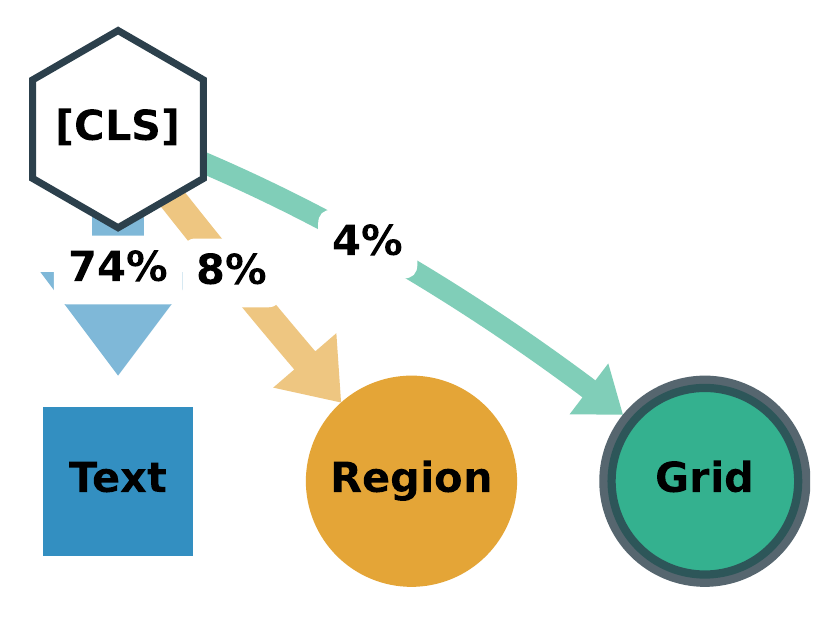}
        \includegraphics[width=.23\linewidth]{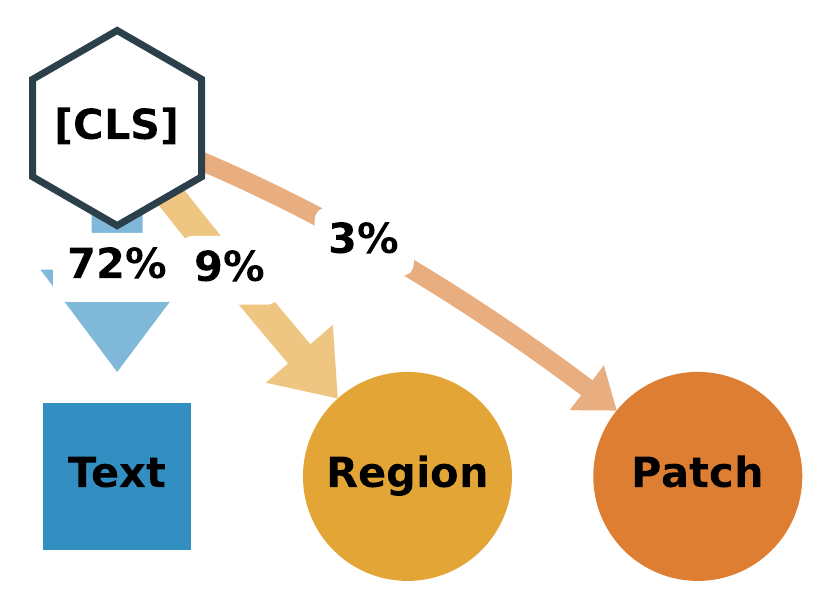}
        \includegraphics[width=.23\linewidth]{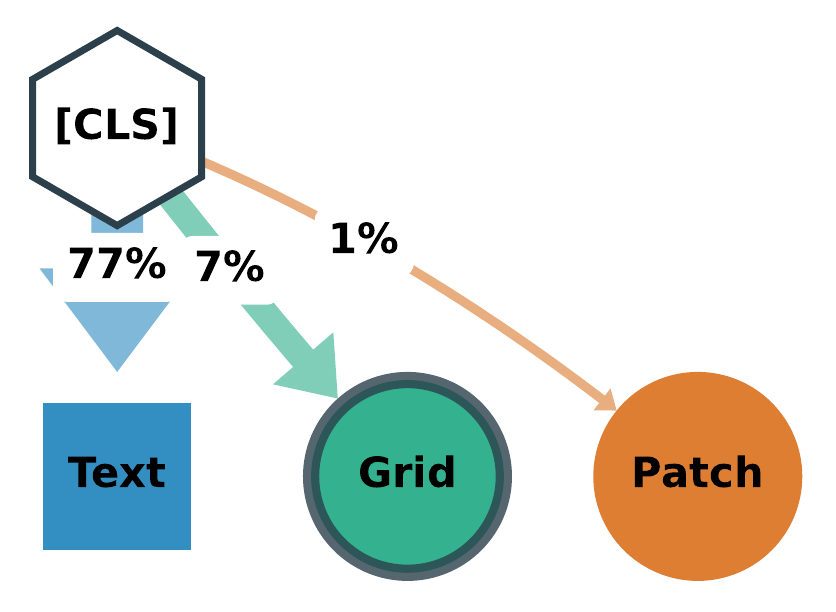}
        \caption{CLS Attention SNLI-VE}
    \end{subfigure}
    \begin{subfigure}{.99\linewidth}
    \centering
        \includegraphics[width=.23\linewidth]{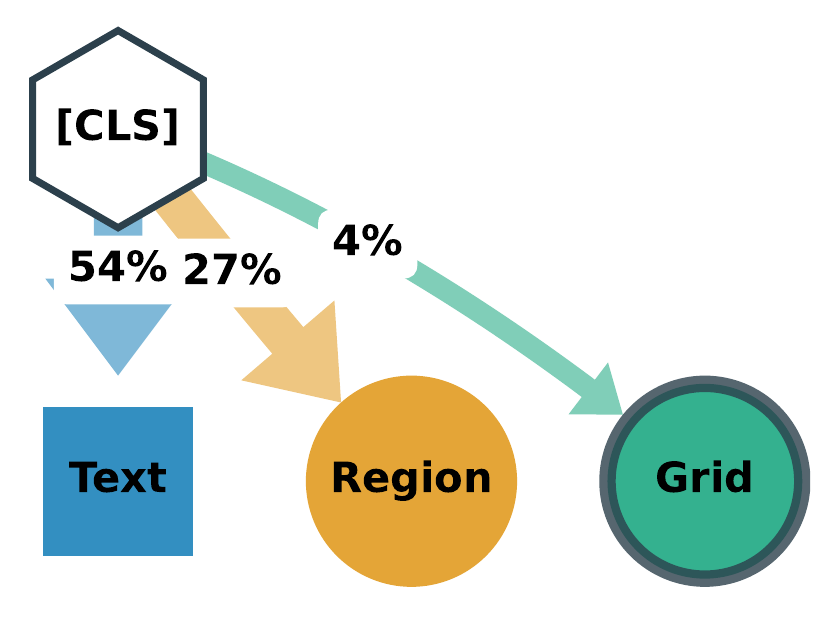}
        \includegraphics[width=.23\linewidth]{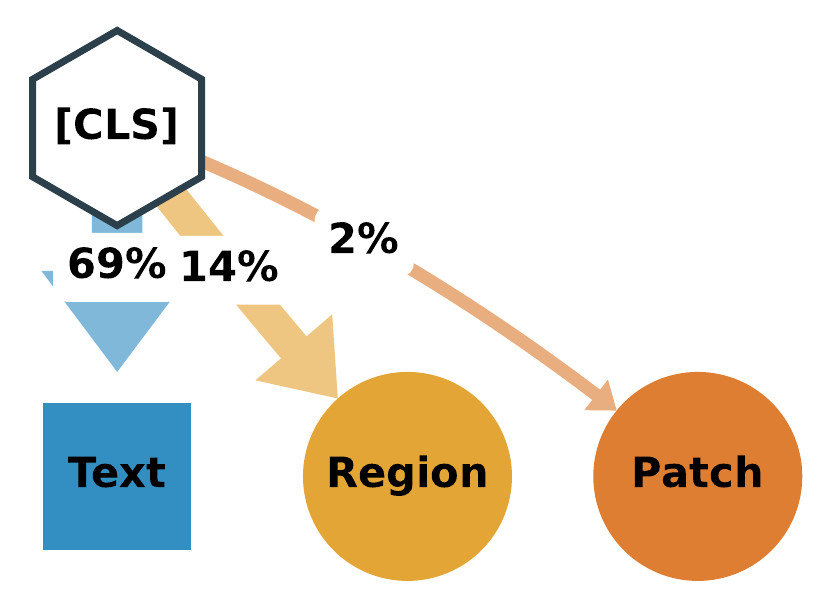}
        \includegraphics[width=.23\linewidth]{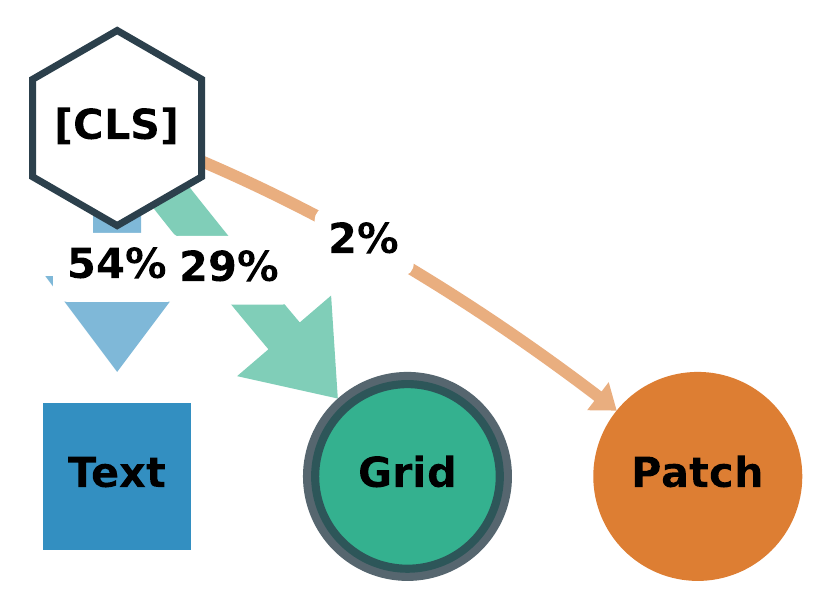}
        \caption{CLS Attention Hateful Memes}
    \end{subfigure}
    \caption{(After VE-Dropout Training) CLS attention weights (in \%) averaged over all heads to the modalities. Numbers do not add to 100\% because of CLS self-attention.}
\label{fig:appendix:ana:clsattention:drop}
\end{figure*}

\begin{figure*}[!ht]
\centering
    \begin{subfigure}{.99\linewidth}
    \centering
        \includegraphics[width=.23\linewidth]{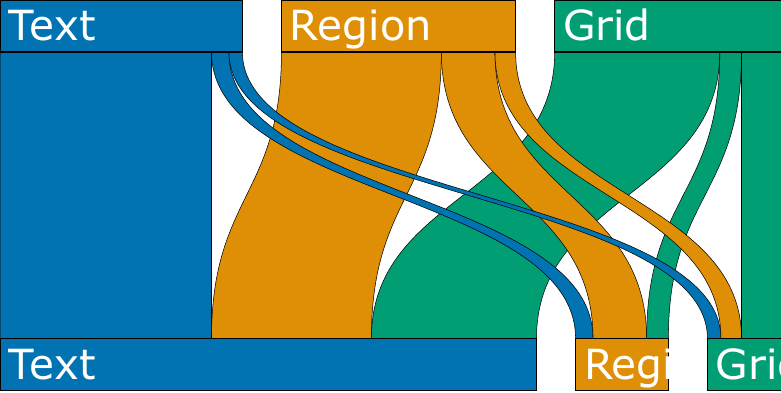}
        \includegraphics[width=.23\linewidth]{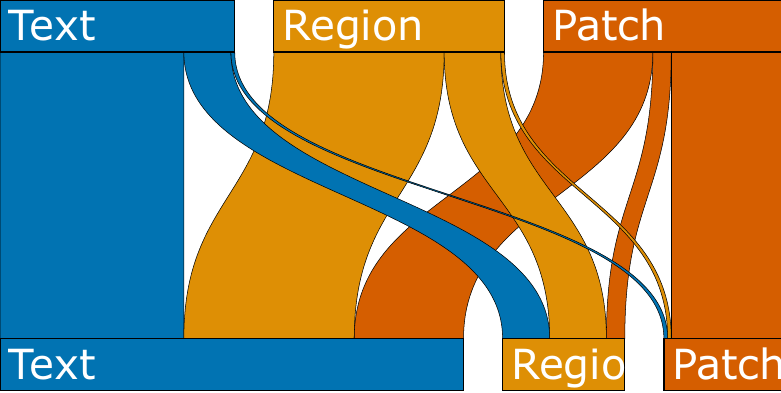}
        \includegraphics[width=.23\linewidth]{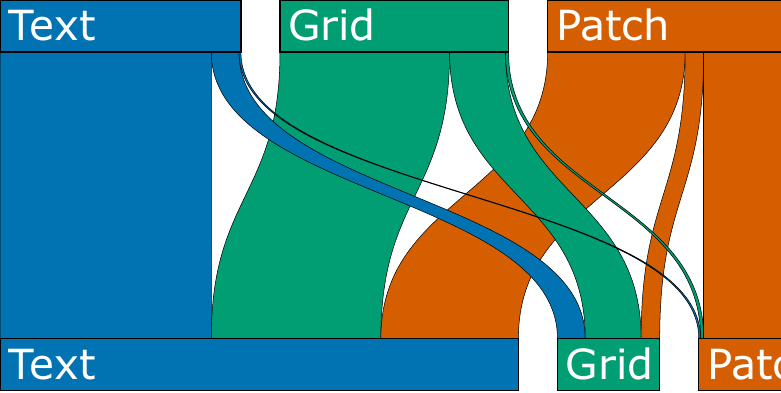}
        \caption{Cross-Attention for Flickr30k}
    \end{subfigure}
    \begin{subfigure}{.99\linewidth}
    \centering
        \includegraphics[width=.23\linewidth]{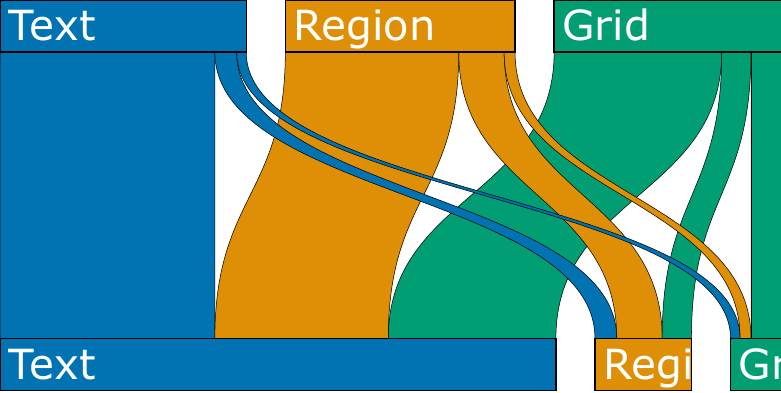}
        \includegraphics[width=.23\linewidth]{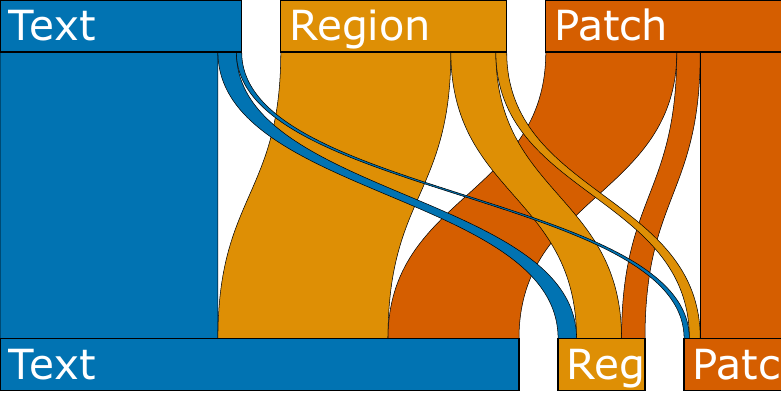}
        \includegraphics[width=.23\linewidth]{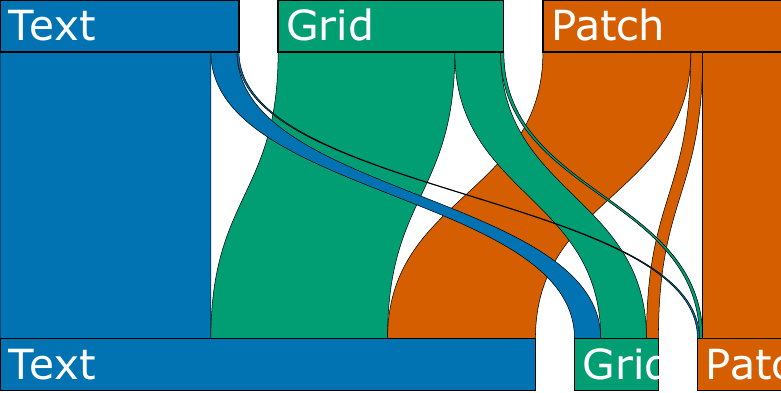}
        \caption{Cross-Attention for MSCOCO}
    \end{subfigure}
    \begin{subfigure}{.99\linewidth}
    \centering
        \includegraphics[width=.23\linewidth]{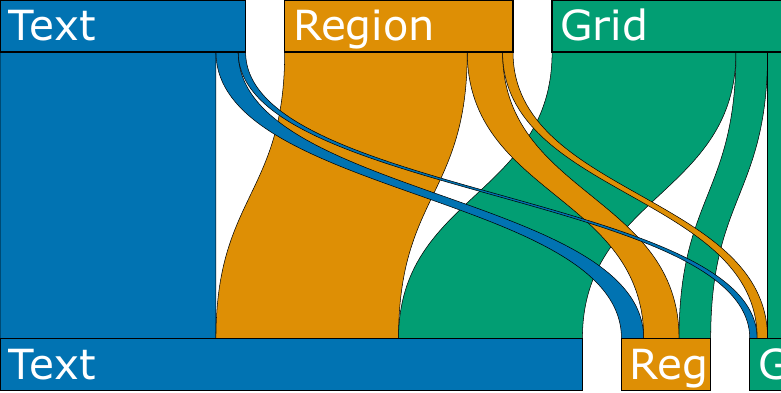}
        \includegraphics[width=.23\linewidth]{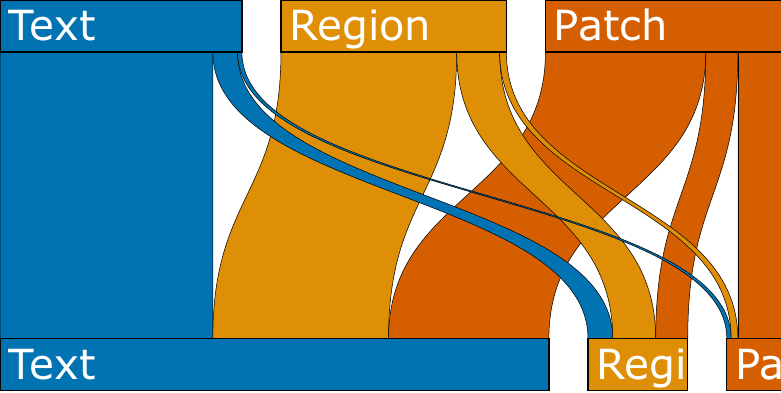}
        \includegraphics[width=.23\linewidth]{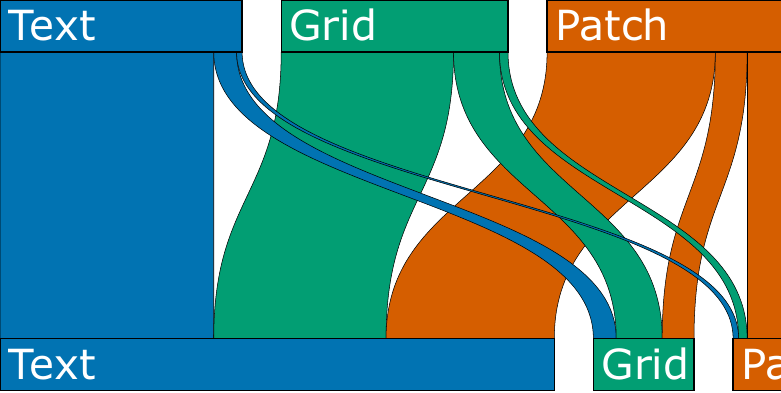}
        \caption{Cross-Attention for GQA}
    \end{subfigure}
    \begin{subfigure}{.99\linewidth}
    \centering
        \includegraphics[width=.23\linewidth]{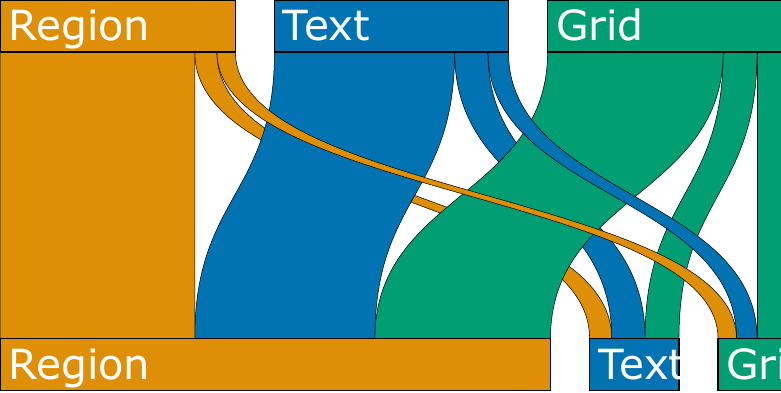}
        \includegraphics[width=.23\linewidth]{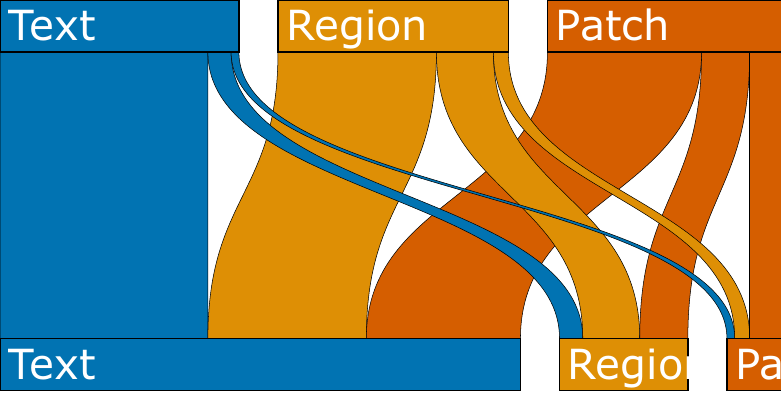}
        \includegraphics[width=.23\linewidth]{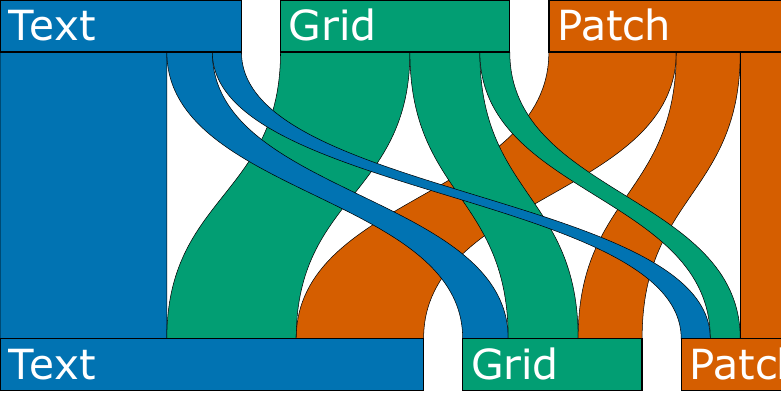}
        \caption{Cross-Attention for VQA}
    \end{subfigure}
    \begin{subfigure}{.99\linewidth}
    \centering
        \includegraphics[width=.23\linewidth]{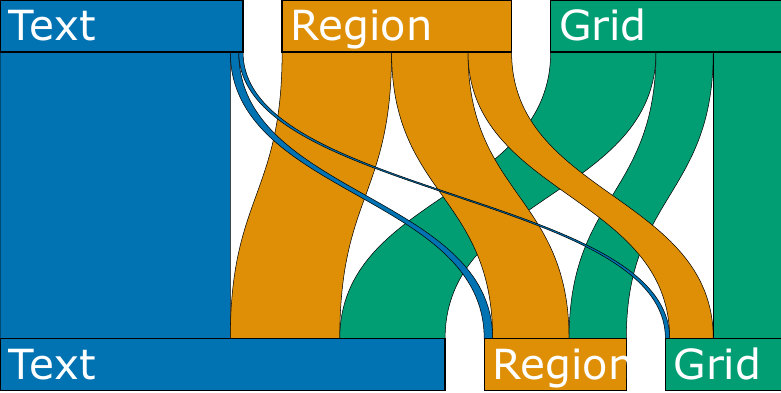}
        \includegraphics[width=.23\linewidth]{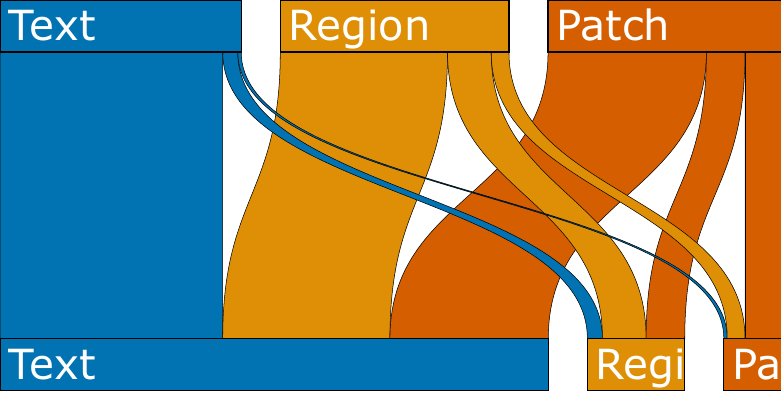}
        \includegraphics[width=.23\linewidth]{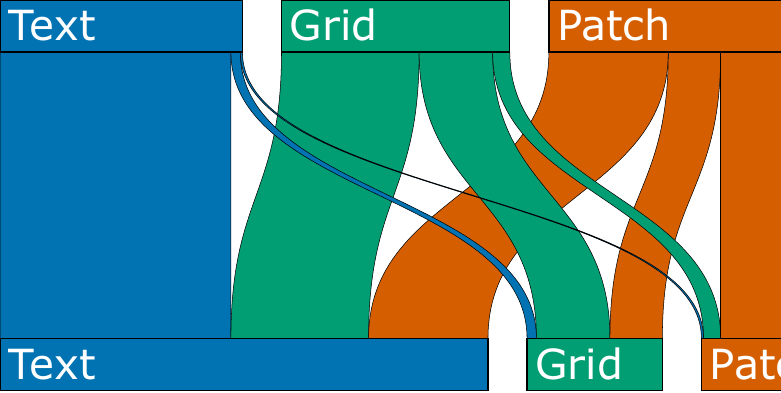}
        \caption{Cross-Attention for SNLI-VE}
    \end{subfigure}
    \begin{subfigure}{.99\linewidth}
    \centering
        \includegraphics[width=.23\linewidth]{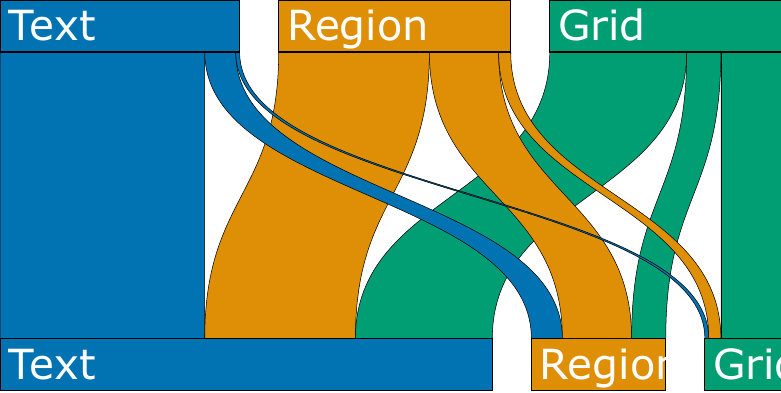}
        \includegraphics[width=.23\linewidth]{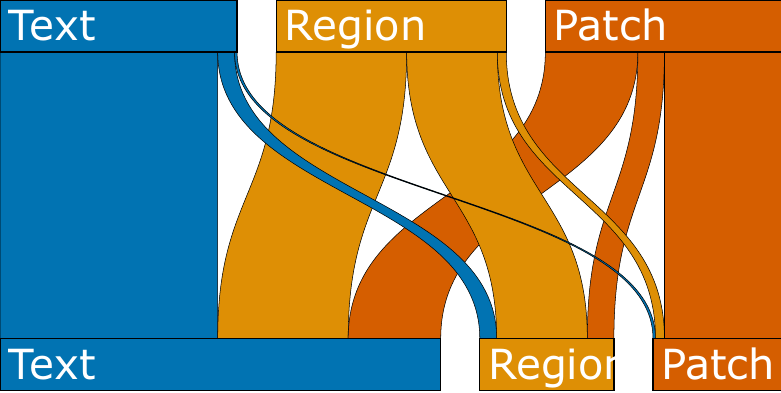}
        \includegraphics[width=.23\linewidth]{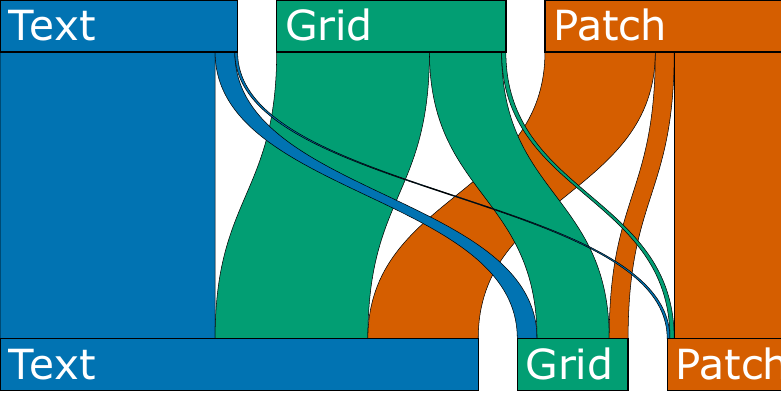}
        \caption{Cross-Attention for Hateful Memes}
    \end{subfigure}
    \caption{(After VE-Dropout Training) Attention flow (in \%) from each modality (top) to all modalities (bottom). Flow is the sum of all attention
weights between the modalities, averaged over all modality tokens and all attention heads.}
\label{fig:appendix:ana:cross:drop}
\end{figure*}

\begin{figure*}[!ht]
\centering
    \begin{subfigure}{.99\linewidth}
    \centering
        \includegraphics[width=.23\linewidth]{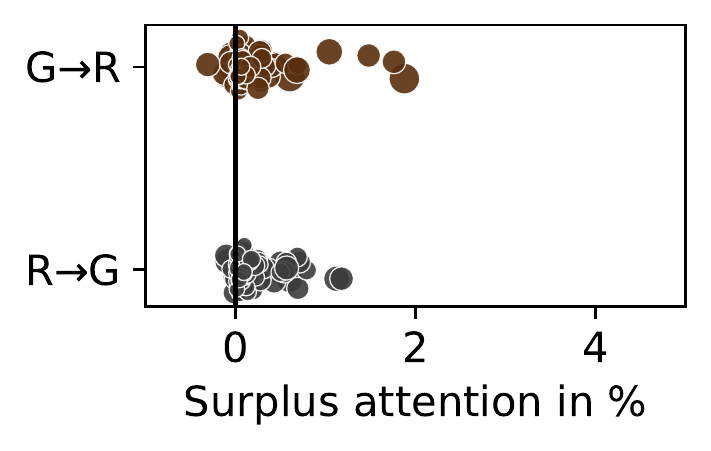}
        \includegraphics[width=.23\linewidth]{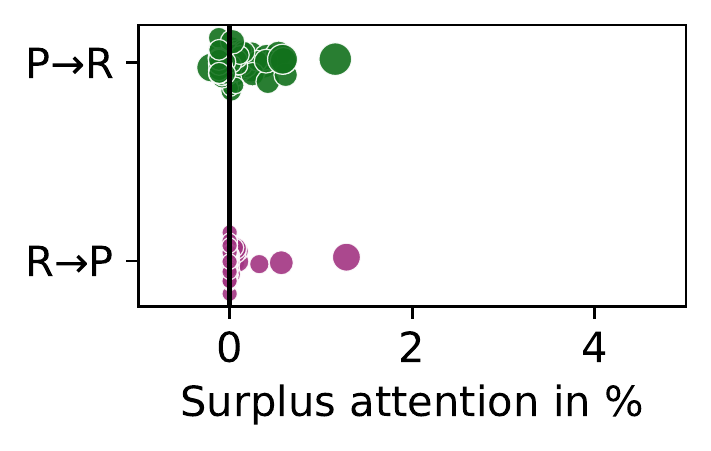}
        \includegraphics[width=.23\linewidth]{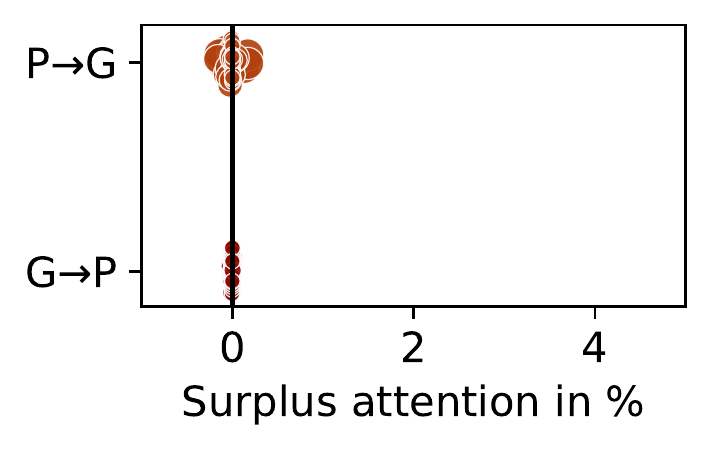}
        \caption{Surplus Attention Flickr30k}
    \end{subfigure}
    \begin{subfigure}{.99\linewidth}
    \centering
        \includegraphics[width=.23\linewidth]{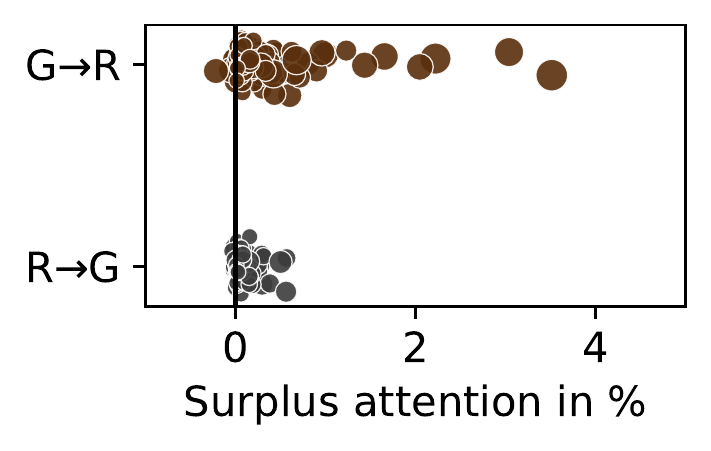}
        \includegraphics[width=.23\linewidth]{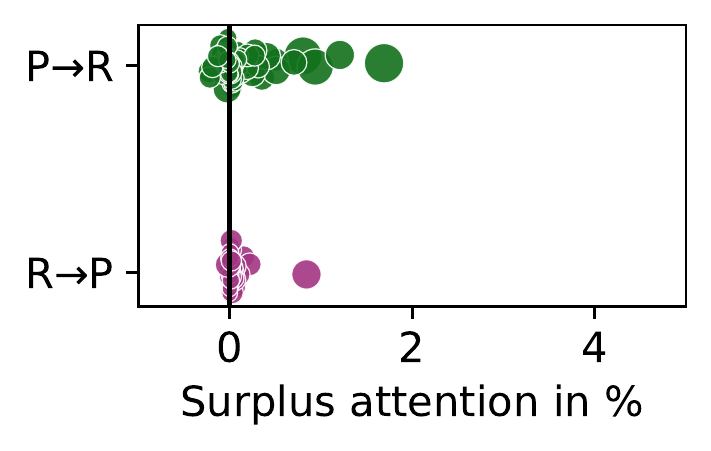}
        \includegraphics[width=.23\linewidth]{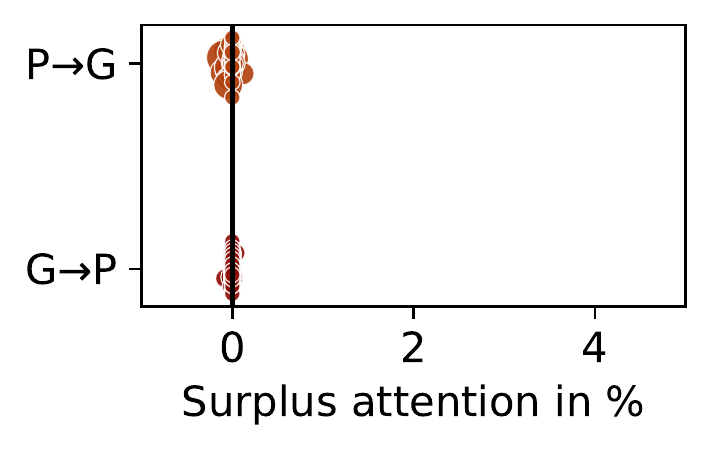}
        \caption{Surplus Attention MSCOCO}
    \end{subfigure}
    \begin{subfigure}{.99\linewidth}
    \centering
        \includegraphics[width=.23\linewidth]{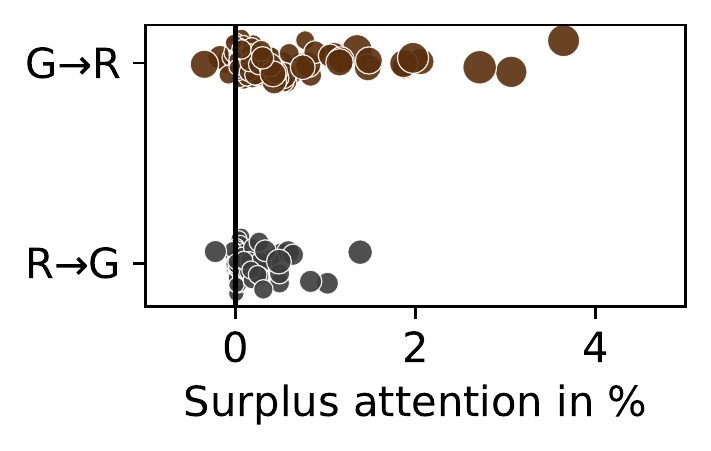}
        \includegraphics[width=.23\linewidth]{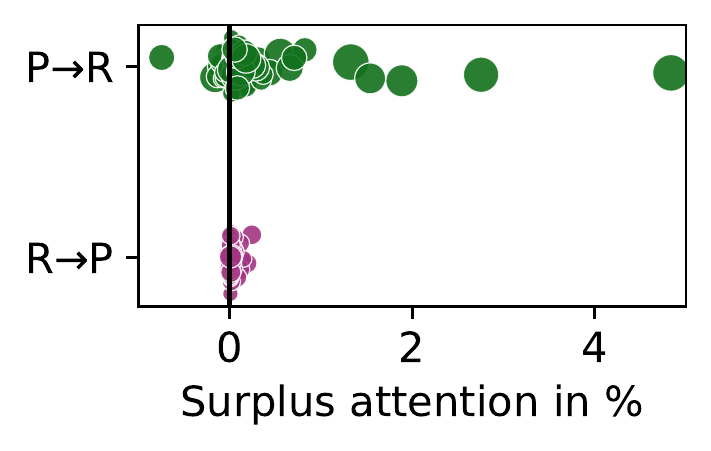}
        \includegraphics[width=.23\linewidth]{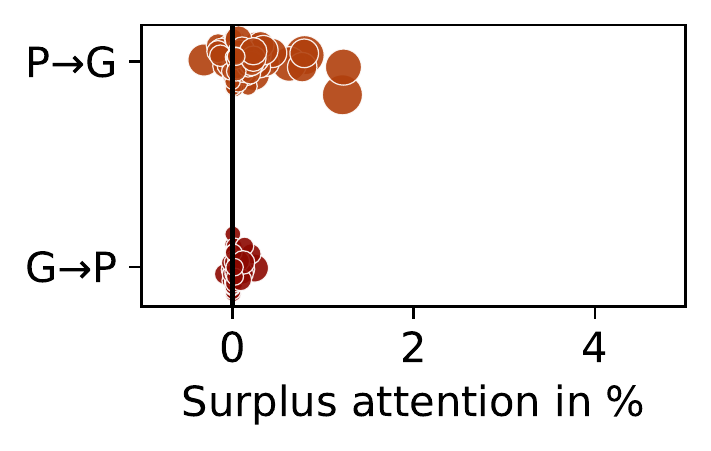}
        \caption{Surplus Attention GQA}
    \end{subfigure}
    \begin{subfigure}{.99\linewidth}
    \centering
        \includegraphics[width=.23\linewidth]{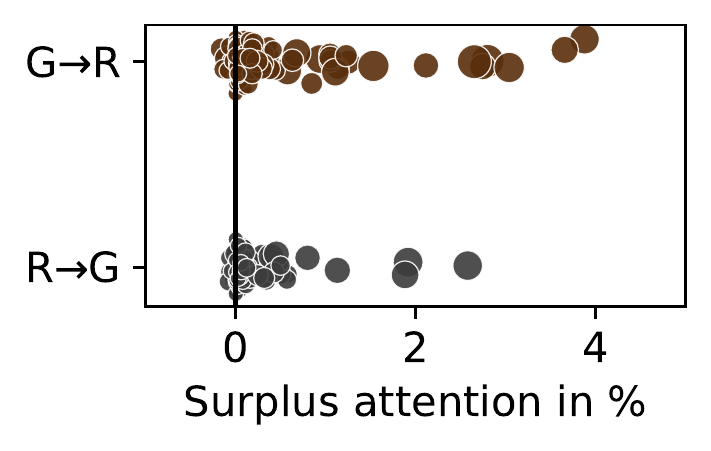}
        \includegraphics[width=.23\linewidth]{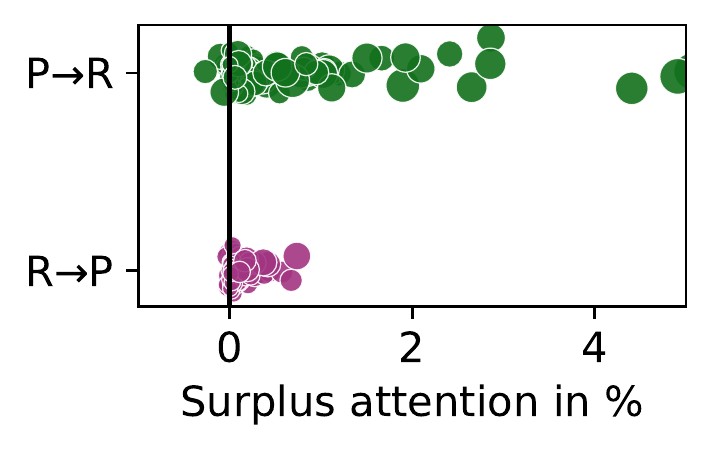}
        \includegraphics[width=.23\linewidth]{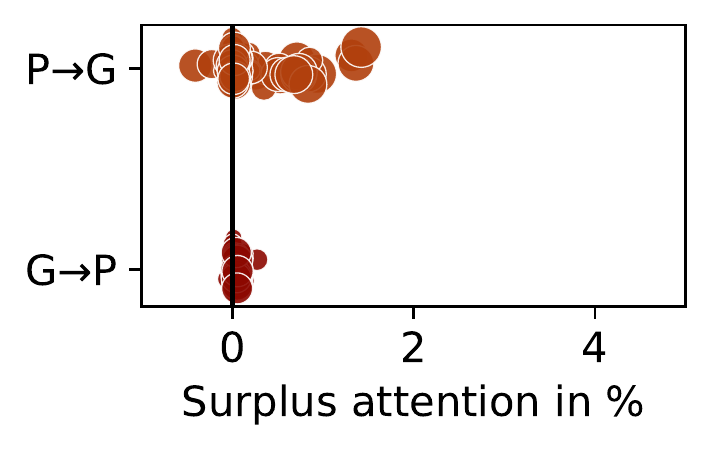}
        \caption{Surplus Attention VQA}
    \end{subfigure}
    \begin{subfigure}{.99\linewidth}
    \centering
        \includegraphics[width=.23\linewidth]{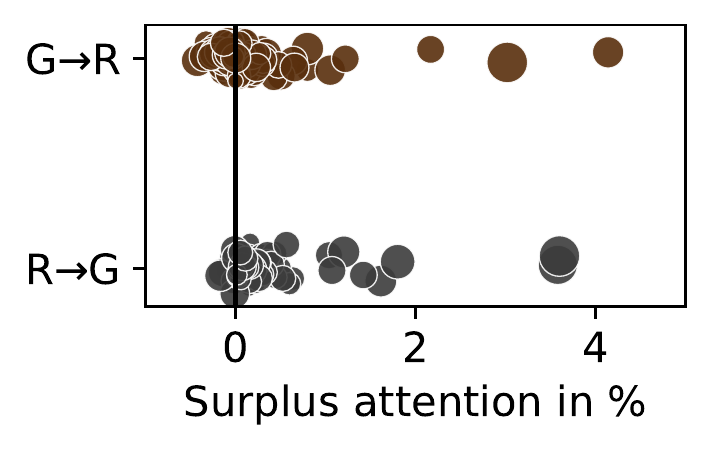}
        \includegraphics[width=.23\linewidth]{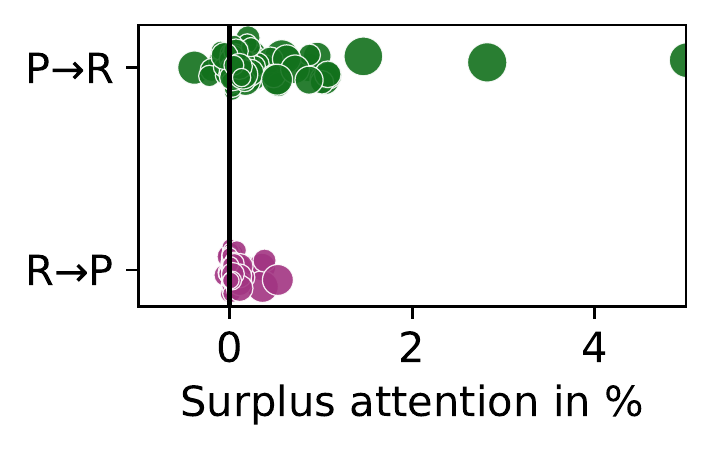}
        \includegraphics[width=.23\linewidth]{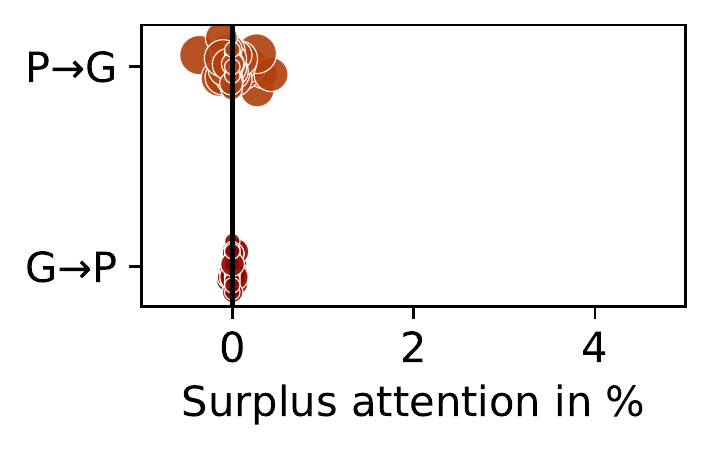}
        \caption{Surplus Attention SNLI-VE}
    \end{subfigure}
    \begin{subfigure}{.99\linewidth}
    \centering
        \includegraphics[width=.23\linewidth]{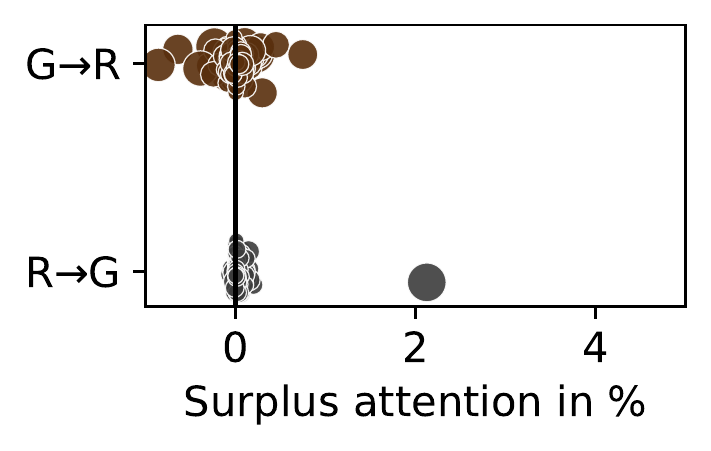}
        \includegraphics[width=.23\linewidth]{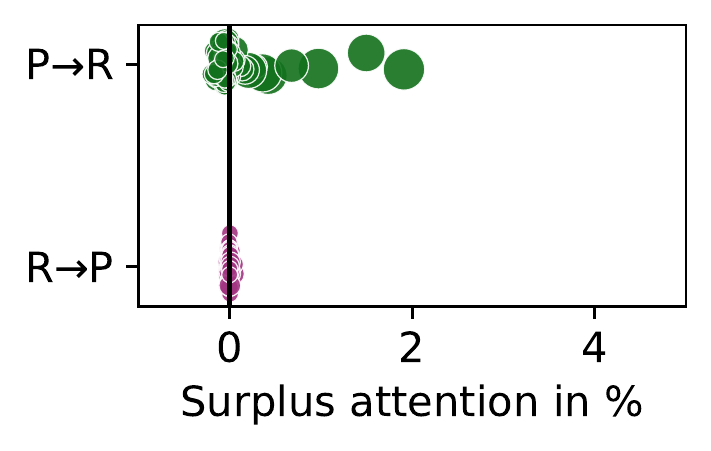}
        \includegraphics[width=.23\linewidth]{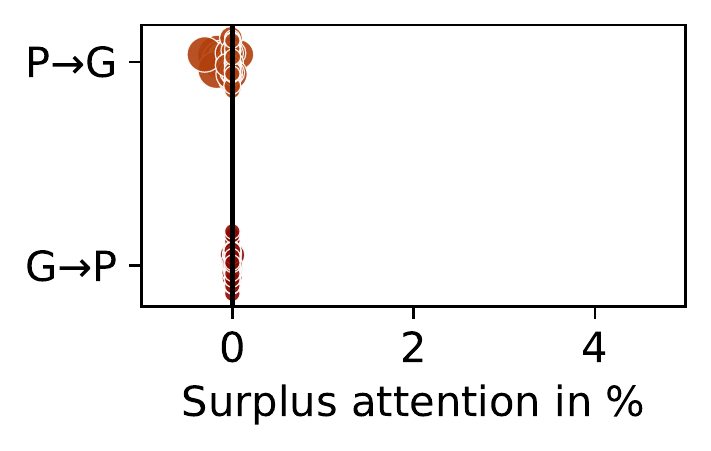}
        \caption{Surplus Attention Hateful Memes}
    \end{subfigure}
    \caption{(After VE-Dropout Training) Surplus attention of attention heads from one VE's tokens to another target VE's overlapping tokens compared to the other non-overlapping tokens of the target VE. \textbf{Dot size} represents the average total attention paid to the target VE by each head.  (Abbreviations: \textbf{R}egion, \textbf{G}rid, \textbf{P}atch).}
\label{fig:appendix:ana:crossimg:drop}
\end{figure*}

We present the full results for all VE combinations from the analysis of \S\ref{sec:ana} \textbf{after} VE-Dropout training.
Figure \ref{fig:appendix:ana:pruning} show the results for VE-Dropout at test time,
Figure \ref{fig:appendix:ana:clsattention:drop} the CLS attention, Figure \ref{fig:appendix:ana:cross:drop} the attention flow, Figure \ref{fig:appendix:ana:crossimg:drop} the surplus attention for overlapping tokens, and Figure \ref{fig:appendix:ana:grounding:drop} the visual grounding.

\end{document}